\pdfoutput=1
\pdfoutput=1
\pdfoutput=1
\pdfoutput=1
\pdfoutput=1

\documentclass[twocolumn,10pt]{IEEEtran}
\normalsize

\usepackage{enumerate}
\usepackage{color}

\usepackage{cite}

\usepackage{float}
\newfloat{figtab}{htb}{fgtb}
\makeatletter
\newcommand\figcaption{\def\@captype{figure}\caption}
\newcommand\tabcaption{\def\@captype{table}\caption}
\makeatother

\usepackage{graphicx}
\usepackage{booktabs}
\usepackage{algorithm}
\usepackage{bm}
\ifCLASSINFOpdf
\else
\fi

\usepackage{amsthm}
\usepackage{amssymb,amsfonts}
\usepackage{amsfonts,balance}

\usepackage[cmex10]{amsmath}

\newtheorem{theorem}{Theorem}

\usepackage{algorithmic}

\usepackage{array}
\usepackage{multirow,makecell}

\usepackage[caption=false,font=footnotesize]{subfig}

\usepackage{verbatim}

\begin{document}
%
\title{Federated Learning and Meta Learning:  Approaches, Applications, and Directions}

\author{Xiaonan~Liu,~\IEEEmembership{Member,~IEEE,}
        Yansha~Deng,~\IEEEmembership{Senior Member,~IEEE,}\\Arumugam~Nallanathan and Mehdi~Bennis,~\IEEEmembership{Fellow,~IEEE}\\
\thanks{X. Liu is with the Department of Engineering, King's College London, U.K. (e-mail:\{liuxiaonan19931107\}@gmail.com). (This work is performed at KCL.)}
\thanks{Y. Deng is with the Department of Engineering, King’s College London, London, WC2R 2LS, U.K. (e-mail:\{yansha.deng\}@kcl.ac.uk). (Corresponding author: Yansha Deng).}
\thanks{A. Nallanathan is with the School of Electronic Engineering and Computer Science, Queen Mary University of London (QMUL), U.K. (e-mail:\{a.nallanathan\}@qmul.ac.uk).}
\thanks{M. Bennis is with Faculty of Information Technology and Electrical Engineering, Centre for Wireless communications, University of Oulu, Finland. (e-mail:\{mehdi.bennis\}@oulu.fi).}
\thanks{This work was supported in part by Engineering and Physical Sciences Research Council (EPSRC), U.K., under Grant EP/W004348/1 , EP/W004100/1, and in part by UKRI under the UK government’s Horizon Europe funding guarantee (grant number 10061781), as part of the European Commission-funded collaborative project VERGE, under SNS JU program (grant number 101096034).}
}

\maketitle

\begin{abstract}
Over the past few years, significant advancements have been made in the field of machine learning (ML) to address resource management, interference management, autonomy, and decision-making in wireless networks. Traditional ML approaches rely on centralized methods, where data is collected at a central server for training. However, this approach poses a challenge in terms of preserving the data privacy of devices. To address this issue, federated learning (FL) has emerged as an effective solution that allows edge devices to collaboratively train ML models without compromising data privacy. In FL, local datasets are not shared, and the focus is on learning a global model for a specific task involving all devices. However, FL has limitations when it comes to adapting the model to devices with different data distributions. In such cases, meta learning is considered, as it enables the adaptation of learning models to different data distributions using only a few data samples. In this tutorial, we present a comprehensive review of FL, meta learning, and federated meta learning (FedMeta). Unlike other tutorial papers, our objective is to explore how FL, meta learning, and FedMeta methodologies can be designed, optimized, and evolved, and their applications over wireless networks. We also analyze the relationships among these learning algorithms and examine their advantages and disadvantages in real-world applications.
\end{abstract}

\begin{IEEEkeywords}
Centralized learning, distributed learning, federated learning, meta learning, federated meta learning, wireless networks.
\end{IEEEkeywords}

\IEEEpeerreviewmaketitle

\section{Introduction}
The rapid advancement of technology and the increasing proliferation of devices, including IoT sensors, mobile phones, and tablets, have resulted in an unprecedented growth in data generation \cite{Lueth,Cisco}. According to a report issued by the SG Analytics, 2.5 quintillion bytes of data were generated every day in 2020 \cite{datagrowth}. To extract meaningful information and enable dynamic decision-making for various tasks, machine learning (ML) algorithms are used to analyze these datasets \cite{Hzhou,Jiang1}, such as controlling self-driving cars \cite{Bojarski}, pattern recognition \cite{Bishop}, or prediction of user behavior \cite{Vasilakos}.

\begin{figure}[!h]
    \centering
    \includegraphics[width=3.5 in]{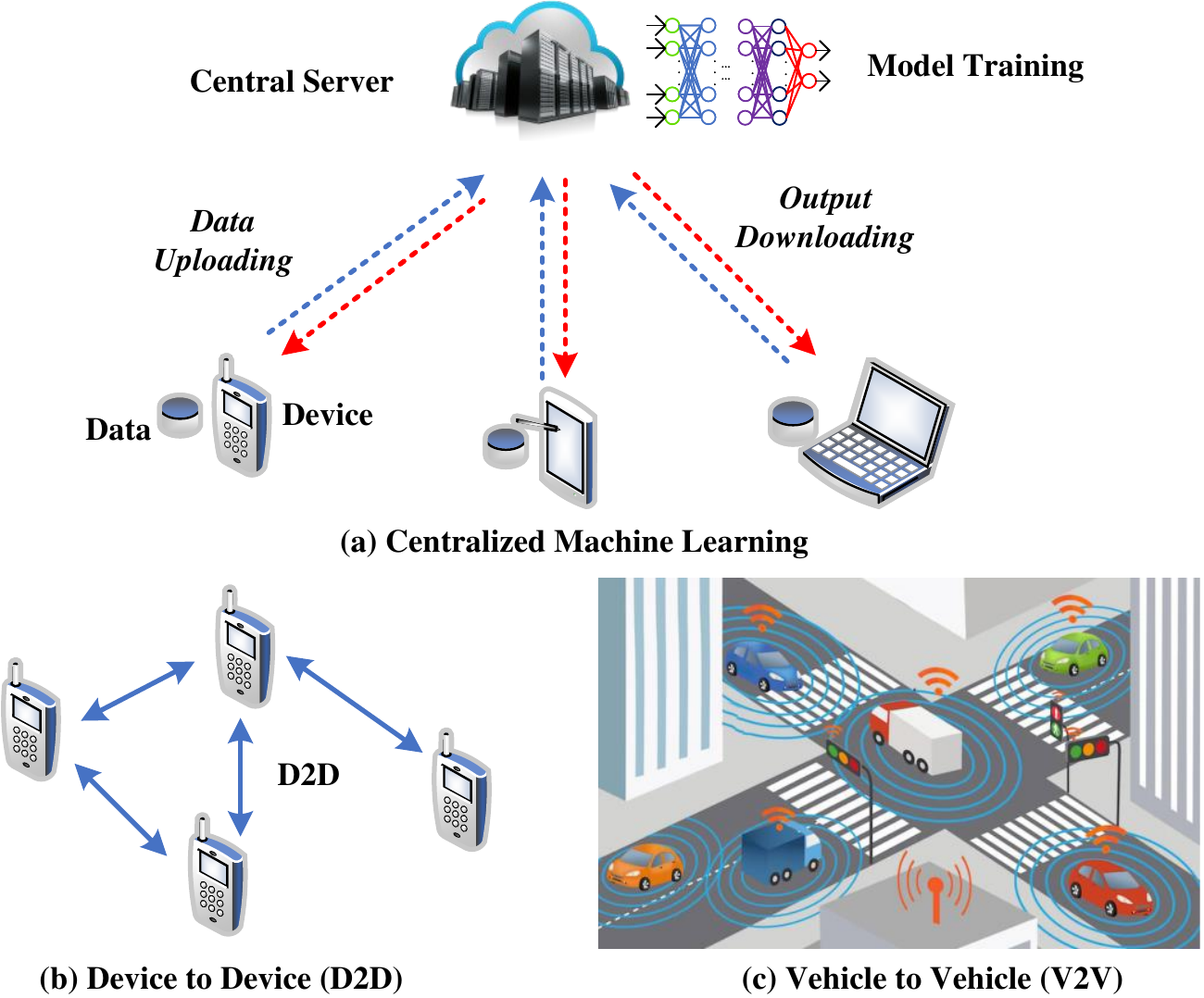}
    \caption{The canonical architecture of centralized machine learning, device to device communication, and vehicle to vehicle communication.}
    \label{basic_modules}
\end{figure}

According to \cite{Rimal}, in traditional cloud computing systems, various types of datasets, such as images, videos, audio files, and location information, acquired from the internet of thing (IoT) devices or mobile devices, are transmitted to a cloud-based server or data center \cite{pingli}, where centralized ML models are trained. As shown in Fig. 1, in centralized learning, the devices communicate with a central server to upload their data by wired/wireless connections. In particular, the devices transmit their local data to a central server, and the central server performs all the computational tasks to train the data. Then, the output of the centralized learning model is delivered to devices. Centralized learning is computationally-efficient for devices as they are not required to be equipped with high computation capability. However, the traditional centralized training method may not be suitable for the ever-growing complexity and heterogeneity of wireless networks, such as device to device (D2D) and vehicle to vehicle (V2V) communications, as shown in Fig. 1 (b) and (c), respectively, due to the following reasons:
\begin{itemize}
    \item The presence of limited bandwidth, dynamic wireless channels, and high interference can result in significant transmission latency, which can negatively affect real-time applications, such as wireless virtual reality (VR) \cite{Liu_vr1,Liu_vr2,Liu_vr3}, D2D, and self-driving car systems \cite{Ananthanarayanan}. These applications require immediate and responsive data processing, but the delays caused by limited bandwidth and the unstable wireless channels negatively affect their performance and reliability.
    \item The process of transmitting a large amount of data to the cloud leads to significant communication overhead, as well as increased storage and computational costs \cite{Tsoumakas,D_Liu}.
    \item The data collected from users, such as medical and financial information, can often be private and sensitive in nature. Transmitting this data to the cloud raises concerns on privacy and security, as it exposes users' personal information to potential risks. This situation becomes particularly problematic when it comes to its compliant with data privacy legislation, such as the European Commission’s General Data Protection Regulation (GDPR) \cite{custers} and the Consumer Privacy Bill of Rights in the U.S. \cite{Gaff}.
\end{itemize}

In order to overcome the aforementioned challenges, distributed learning has emerged as a solution to effectively and efficiently learn models from distributed data. Distributed learning refers to the use of multi-node machine learning algorithms and systems that are specifically designed to address the computational challenges associated with complex centralized algorithms operating on large-scale datasets  \cite{Galakatos,Muhammad_asad}. By employing distributed learning algorithms, multiple learning models can be trained based on distributed datasets. This approach offers several advantages over centralized approaches, particularly when the number of datasets is large. One notable advantage is the potential for reducing biases, as the distributed learning enables the utilization of diverse datasets and mitigates the impact of individual dataset characteristics \cite{Hansen,xindong,Mucke}.

Conventional machine learning algorithms, such as $k$-nearest neighbor \cite{Dellis}, support vector machine \cite{Mangasarian,Vaidya}, Bayesian networks \cite{Wright,Wright1}, and decision trees \cite{Fang_Yang,Patterso}, can be trained in a distributed manner by exchanging raw data, which can hardly protect the user data privacy \cite{Kulkarni}. In order to ensure the privacy of data and facilitate collaborative machine learning (ML) involving complex models distributed across IoT or mobile devices, a method called federated learning (FL) was initially introduced in \cite{B_McMahan} by McMahan et al. The standard steps of FL include: 1) each device trains a local model using its own dataset; 2) the devices send their local models to a central server for model aggregation; 3) the server updates the global model and transmits it back to the devices. These steps are repeated iteratively until convergence. In addition, FL offers several advantages over traditional centralized learning and distributed learning approaches that rely on the exchange of raw data. Some of these advantages include:
\begin{itemize}
    \item $\textbf{Data Privacy:}$ FL prioritizes user privacy by exchanging only model weights between the server and the devices, rather than the datasets themselves. This approach effectively protects user privacy during the collaborative learning process.
    \item $\textbf{Data Diversity:}$ FL enables the utilization of heterogeneous data by allowing the aggregation of models from different devices, leading to the development of an enhanced global model. This aggregation process enhances the overall performance and effectiveness of the model from various devices.
\end{itemize}

In addition to the aforementioned advantages, FL has demonstrated its effectiveness in various applications, including training predictive models for human trajectory/behavior by mobile devices \cite{Sozinov}, automatically learning users' behavior patterns through smart home IoT \cite{Seshan}, and enabling collaborative diagnosis in health artificial intelligence (AI) among multiple hospitals and government agencies \cite{Brisimi,Powell,Verma}. However, these applications usually suffer from data heterogeneity (i.e., data with different statistical characteristics) and system heterogeneity (i.e., systems with different types of computation units), which severely affect the convergence and accuracy of FL algorithms. It is worth noting that FL primarily focuses on learning a single ML task across multiple devices \cite{Arambakam,TRINDADE2022} and only develops a common output for all devices. Therefore, it does not adapt the model to each device. This is an important missing feature, especially given the heterogeneity of the underlying data distribution for various devices.
To address these limitations, integrating meta learning, also known as learning to learn, with FL becomes crucial, resulting in what is known as federated meta learning (FedMeta). FedMeta aims to solve multi-task learning problems. Meta learning aims to create AI models that are able to adapt to new tasks and improve their learning performance over time, without the need for extensive retraining. In other words, meta learning typically involves training a learning model on a variety of different tasks, with the goal of learning generalizable knowledge that can be transferred to new tasks. This is different from traditional ML, where a learning model is typically trained on a single task and then used for that task alone. In general, a meta learning algorithm is trained using the outputs, e.g., model predictions, and metadata of ML algorithms. After finishing training, its learning models are delivered for testing and used to make final predictions. In addition, meta learning includes tasks such as observing the learning performance of different ML models on learning tasks, learning from metadata, and faster learning process for new tasks.
While both federated learning (FL) and meta learning can be used in distributed learning systems and involve sharing a global model among multiple devices, they differ in three key aspects: (1) \textit{Data Distribution:} In FL, devices have their own datasets  but perform the same task. On the other hand, meta learning involves multiple tasks, each with its corresponding dataset. (2) \textit{Update Mechanism:} FL utilizes local updates deployed by devices to enhance learning performance, whereas meta learning employs inner-loop updates for each task to improve learning performance. (3) \textit{Aggregation and Parameter Updates:} FL applies model aggregation to improve the global performance of all devices. In contrast, meta learning employs an outer-loop to update global parameters for all tasks. Initialization-based meta-learning algorithms, in particular, are known for fast adaptation and good generalization to new tasks. These distinctions highlight the different approaches and objectives of FL and meta learning. While FL focuses on collaborative learning with devices having different data distributions but performing the same task, meta learning guarantees the adaptation of models to multiple tasks with their corresponding datasets, often emphasizing fast adaptation and generalization capabilities, especially by initialization-based algorithms \cite{Schweighofer}. The goal of FedMeta is to collaboratively meta-train a learning model using data from different tasks distributed among devices. The server maintains the initialized model and updates it by collecting testing loss from a mini batch of devices. The transmitted information in learning process consists of the model parameter initialization (from a server to devices) and testing loss (from devices to the server), and no data is required to be delivered to the server. Compared to FL, FedMeta has the following advantages: 
\begin{itemize}
    \item FedMeta brings a reduction in the required communication cost because of faster convergence, and an increase in learning accuracy. Additionally, it is highly adaptable and can be employed for arbitrary tasks across multiple devices. This flexibility allows for efficient and accurate learning in various scenarios.
    \item FedMeta enables model sharing and local model training without substantial expansion in model size. As a result, it does not consume a large amount of memory, and the resulting global model can be personalized for each device. This aspect allows for efficient utilization of resources and customization of the global model to satisfy the specific requirements of devices.
\end{itemize}

\subsection{Related Works}
This section provides a brief review of relevant surveys and tutorials on FL and meta-learning. Additionally, it highlights the novel contributions of this paper.

\subsubsection{FL}
In the past 5 years, numerous surveys and tutorials have been published on FL methodologies \cite{QYang,YJin,Priyanka,C_zhang,Shaoxiong,Shahid,yushi,jiajuqi,Mliu,Rongfeizeng,yufengzhang,9846956} and their applications over wireless \cite{JPark1,Niknam,ZZhao,GZhu,TLi,KYang,WYBLim,MChen,Imteaj,Dinh_Nguyen,JPark,chen_gunduz,ChenhaoXu} networks. To differentiate our tutorial from these existing surveys and tutorials, we have classified them into different categories based on their primary focus in Table I. We then compare and summarize the content of these surveys and tutorials, aligning them with the structure of our own tutorial, as presented in Tables II and III. Table I highlights that surveys and tutorials for FL methodologies primarily focus on either fundamental definitions, architectures, challenges, future directions, and applications of FL \cite{QYang,Priyanka,C_zhang}, or a specific subfield of FL, such as emerging trends of FL (FL in the intersection with other learning paradigms) \cite{Shaoxiong}, communication efficiency of FL (challenges and constraints caused by limited bandwidth and computation ability) \cite{Shahid}, fairness-aware FL (client selection, optimization, contribution evaluation, incentive distribution, and performance metrics) \cite{yushi}, federated reinforcement learning (FRL) (definitions, evolution, and advantages of horizontal FRL and vertical FRL) \cite{jiajuqi}, FL for natural language processing (NLP) (algorithm challenges, system challenges as well as privacy issues) \cite{Mliu}, incentive schemes for FL (stackelberg game, auction, contract theory, Shapley value, reinforcement learning, and blockchain) \cite{Rongfeizeng,yufengzhang}, unlabeled data mining in FL (potential research directions, application scenarios, and challenges) \cite{YJin}, and FL over next-generation Ethernet Passive Optical Networks \cite{9846956}. On the other hand, FL surveys and tutorials for wireless networks mainly focus on either fundamental theories, key techniques, challenges, future directions, and applications \cite{Niknam,ZZhao,GZhu,TLi,chen_gunduz}, or a specific subfield of FL applied in wireless networks, including FL in IoT (data sharing, offloading and caching, attack detection, localization, mobile crowdsensing, and privacy) \cite{KYang,Imteaj,Dinh_Nguyen}, communication-efficient FL under various challenges incurred by communication, computing, energy, and data privacy issues \cite{JPark1,JPark}, FL for mobile edge computing (MEC) (communication cost, resource allocation, data privacy, data security, and implementation) \cite{WYBLim}, collaborative FL (definitions, advantages, drawbacks, usage conditions, and performance metrics) \cite{MChen}, and asynchronous FL (device heterogeneity, data heterogeneity, privacy and security, and applications) \cite{ChenhaoXu}. To the best of the authors' knowledge, this is the first paper considering FL methodologies in different research areas, including model aggregation, gradient descent, communication efficiency, fairness, Bayesian learning, and clustering, and how FL algorithms evolve from the canonical one, namely, federated averaging (FedAvg), in detail. Also, we present a qualitative comparison considering advantages and disadvantages among different FL algorithms in the same research area.

\subsubsection{Meta Learning}
There have been several surveys and tutorials focusing on meta learning methodologies over the past 20 years \cite{Vilalta,Lemke,jaoquin,Yin,Hospedales,peng,yaoma}. However, to the best of the authors' knowledge, there are no existing surveys and tutorials in meta learning over wireless networks. To distinguish the scope of our tutorial from existing literature, we classify the available meta learning surveys and tutorials into different categories based on their focus, as presented in Table I. Furthermore, we compare and summarize the content of these existing surveys and tutorials in Table IV, aligning them with the structure and content of our own tutorial. Table I illutrates that the existing meta learning surveys and tutorials focused either on the general advancement, including definitions, models, challenges, research directions, and applications, of meta learning \cite{Vilalta,Lemke,Hospedales,peng}, or exploring the detailed applications of meta learning in a specific meta learning field, including algorithm selection (transfer learning, few-shot learning, and beyond supervised learning) for data mining \cite{jaoquin}, NLP (especially few-shot applications, including definitions, research directions, and some common datasets) \cite{Yin}, and multi-modal meta learning in terms of the methodologies (few-shot learning and zero-shot learning) and applications \cite{yaoma}. To the best of the authors' knowledge, no existing surveys or tutorials have covered the application of FedMeta in wireless networks, making our tutorial a unique contribution to this area of research.

\begin{table*}[]
    \centering
    \caption{An Overview of Selected Surveys and Tutorials on FL and Meta Learning}
    \label{table_survey}
    \resizebox{\textwidth}{32mm}{\begin{tabular}{|c|c|c|}
    \hline
    Subject & Ref. & Contributions\\
    \hline
    \multirow{6}{4em}{FL Methodologies} 
    &\cite{QYang,Priyanka,C_zhang} & Definitions,
    architectures, challenges, future directions, and applications of the FL framework\\
    &\cite{Shaoxiong}&Survey on emerging trends in FL, including FL in the intersection with other learning paradigms\\
    &\cite{Shahid}&Survey on communication efficiency of FL, including challenges and constraints caused by limited bandwidth and computation ability\\
    &\cite{yushi}&Survey on the fairness-aware FL, including client selection, optimization, contribution evaluation, incentive distribution, and performance metrics\\
    & \cite{jiajuqi} & Survey on federated reinforcement learning (FRL), including definitions, evolution, and advantages of horizontal FRL and vertical FRL\\
    & \cite{Mliu} & Survey on FL for natural language processing (NLP), including algorithm challenges, system challenges as well as privacy issues \\
    & \cite{Rongfeizeng,yufengzhang} & Surveys on incentive schemes for FL in terms of Stackelberg game, auction, contract theory, Shapley value, reinforcement learning, and blockchain \\
    & \cite{YJin} & Survey on unlabeled data in FL, including potential research directions, application scenarios, and challenges \\
    & \cite{9846956} & FL over ethernet passive optical networks, considering dynamic wavelength and bandwidth allocation for quality of service provisioning \\
    \hline
    \multirow{7}{4em}{FL in Wireless Networks}
    & \cite{Niknam,ZZhao,GZhu,TLi,chen_gunduz} & Fundamental theories, key techniques, challenges, future directions, and applications for FL over wireless networks \\
    & \cite{KYang,Imteaj,Dinh_Nguyen} & Surveys on FL in IoT, including data sharing, offloading and caching, attack detection, localization, mobile crowdsensing, and privacy \\
    & \cite{JPark1,JPark} & Surveys on communication-efficient FL under various challenges incurred by communication, computing, energy, and data privacy issues \\
    & \cite{WYBLim} & Survey on FL in MEC, including communication cost, resource allocation, data privacy, data security, and implementation \\
    & \cite{MChen} & Survey on collaborative FL, including the definitions, advantages, drawbacks, usage conditions, and performance metrics \\
    & \cite{ChenhaoXu} & Survey on asynchronous FL, including device heterogeneity, data heterogeneity, privacy and security, and applications \\
    \hline
    \multirow{5}{8em}{Meta Learning Methodologies} & \cite{Vilalta,Lemke,Hospedales,peng} & Introductory tutorial on meta learning, e.g., definitions, models, challenges, research directions, and applications \\ 
    & \cite{jaoquin} & Tutorial on meta learning on algorithm selection (transfer learning, few-shot learning, and beyond supervised learning) for data mining \\
    & \cite{Yin} & Survey on meta learning for NLP, especially few-shot applications, including definitions, research directions, and some common datasets \\
    & \cite{yaoma} & Tutorial on multimodality-based meta-learning in terms of the methodologies (few-shot learning and zero-shot learning) and applications \\
    
    \hline
    \end{tabular}}
\end{table*}

\begin{table*}[!t]
 \caption{Comparison of Surveys and Tutorials on FL Methodologies}
 \label{table_survey}
 \centering 
 \renewcommand{\arraystretch}{1}
  \scalebox{0.715}
  
 \resizebox{\textwidth}{14mm}{\begin{tabular}{c|c|c|c|c|c|c|c|c|c|c|c|c|c}
  \hline
  \multicolumn{2}{c||}{Reference}
    &\cite{QYang}
    &\cite{YJin}
    &\cite{Priyanka}
    &\cite{C_zhang}
    &\cite{Shaoxiong}
    &\cite{Shahid}
    &\cite{yushi}
    &\cite{jiajuqi}
    &\cite{Mliu}
    &\cite{Rongfeizeng}
    &\cite{yufengzhang}
    
    &This paper\\ \hline
    \multicolumn{2}{c||}{Year}
    &2019
    &2020
    &2021
    &2021
    &2021
    &2021
    &2021
    &2021
    &2021
    &2021
    &2021
    &-
  \\ \hline \hline
   \multirow{6}{*}{\makecell[c]{Section III:\\FL\\Methodologies}}
   &{\makecell{(III.A): Model Aggregation}}
   & \checkmark 
   &      \checkmark        
   &   \checkmark
   &        \checkmark   
   &  \checkmark
   & 
   &       
   & \checkmark
   &\checkmark
   &\checkmark
   &\checkmark
   &\checkmark  \\

   \cline{2-14}
   &{\makecell{(III.B): Gradient Descent}}
   & \checkmark 
   & \checkmark             
   &    \checkmark
   &     \checkmark
   & 
   &   
   & 
   &  \checkmark
   &\checkmark
   &\checkmark
   &
   &\checkmark            \\

   \cline{2-14}
   &{\makecell{(III.C): Communication Efficiency}}
   &
   &            
   &  
   &      
   &  
   &   \checkmark    
   &  
   &  
   &
   &
   &
   &\checkmark           \\

   \cline{2-14}
   &{\makecell{(III.D): Fairness}}
   &
   &            
   & 
   &  
   & \checkmark
   &
   & \checkmark
   &
   &
   &\checkmark
   &
   &\checkmark            
   \\ \cline{2-14}

   &{\makecell{(III.E): Bayesian Machinery}}
   &
   &            
   &  
   &            
   &  \checkmark
   &  
   &       
   & 
   &
   &\checkmark
   &
   &\checkmark  \\

   \cline{2-14}
   &{\makecell{(III.F): Clustering}}
   &
   &            
   &  
   &            
   &  \checkmark
   &         
   &  
   &  
   &
   &
   &
   &\checkmark           \\

   \cline{2-14}
   \hline
   \end{tabular}}
\end{table*}

\begin{table*}[!t]
 \caption{Comparison of Surveys and Tutorials on FL Over Wireless Networks}
 \label{table_survey}
 \centering 
 \renewcommand{\arraystretch}{1}
  \scalebox{0.715}
  
 \resizebox{\textwidth}{28mm}{\begin{tabular}{c|c|c|c|c|c|c|c|c|c|c|c|c|c|c|c}
  \hline
  \multicolumn{2}{c||}{Reference}
    &\cite{JPark1}
    &\cite{Niknam}
    &\cite{ZZhao}
    &\cite{GZhu}
    &\cite{TLi}
    &\cite{KYang}
    &\cite{WYBLim}
    &\cite{MChen}
    &\cite{Imteaj}
    &\cite{Dinh_Nguyen}
    &\cite{JPark}
    &\cite{chen_gunduz}
    &\cite{ChenhaoXu}
    &This paper\\ \hline
    \multicolumn{2}{c||}{Year}
    &2019
    &2020
    &2020
    &2020
    &2020
    &2020
    &2020
    &2020
    &2021
    &2021
    &2021
    &2021
    &2021
    &-
  \\ \hline \hline
  \multirow{14}{*}{\makecell[c]{Section\\ IV:\\FL\\in \\Wireless \\ Networks}}
   &{\makecell{(IV.A.1): Device \\Selection}}
   & 
   &              
   &  
   &          
   &  \checkmark
   & \checkmark
   &   \checkmark     
   &   
   &\checkmark 
   &
   &
   &\checkmark
   &
   &\checkmark  \\

   \cline{2-16}
   &{\makecell{(IV.A.2.a): Communication\\Efficiency}}
   & \checkmark
   &              
   &  \checkmark
   &          
   &  \checkmark
   & 
   &    \checkmark  
   & 
   &\checkmark
   &\checkmark
   &\checkmark
   &\checkmark
   &
   &\checkmark  \\

   \cline{2-16}
   &{\makecell{(IV.A.2.b): Computation\\Efficiency}}
   & 
   &              
   &  \checkmark
   &          
   &  
   & 
   &   \checkmark   
   & 
   &
   &
   &\checkmark
   &
   &
   &\checkmark  \\

   \cline{2-16}
   &{\makecell{(IV.A.2.c): Energy\\Efficiency}}
   & 
   &              
   &  
   &          
   &  
   & 
   &      
   & 
   &\checkmark
   &
   &\checkmark
   &
   &
   &\checkmark  \\

   \cline{2-16}
   &{\makecell{(IV.A.2): Packet\\Error}}
   & 
   &              
   &  
   &          
   &  \checkmark
   & 
   &      
   & 
   &\checkmark
   &
   &
   &\checkmark
   &
   &\checkmark  \\

   \cline{2-16}
   &{\makecell{(IV.A.3): Resource\\Allocation}}
   & 
   &   \checkmark           
   &  \checkmark
   &    \checkmark      
   &  
   & 
   &    \checkmark  
   & 
   &\checkmark
   &\checkmark
   &
   &\checkmark
   &
   &\checkmark  \\

   \cline{2-16}
   &{\makecell{(IV.A.4): Asynchronous}}
   & 
   &              
   &  
   &          
   &  \checkmark
   & 
   &    \checkmark  
   & \checkmark
   &\checkmark
   &
   &
   &
   &\checkmark
   &\checkmark  \\

   \cline{2-16}
   &{\makecell{(IV.B): Over the Air \\Computation}}
   & 
   &              
   &  
   &  \checkmark        
   &  
   & \checkmark
   &    
   & 
   &
   &
   &\checkmark
   &\checkmark
   &
   &\checkmark  \\

   \cline{2-16}
 
   \hline
   \end{tabular}}
\end{table*}

\begin{table*}[!t]
 \caption{Comparison of Surveys and Tutorials on Meta Learning}
 \label{table_survey}
 \centering 
 \renewcommand{\arraystretch}{1}
  \scalebox{0.715}
  
 \resizebox{\textwidth}{35mm}{\begin{tabular}{c|c|c|c|c|c|c|c|c|c}
  \hline
  \multicolumn{2}{c||}{Reference}
    &\cite{Vilalta}
    &\cite{Lemke}
    &\cite{jaoquin}
    &\cite{Yin}
    &\cite{Hospedales}
    &\cite{peng}
    &\cite{yaoma}
    &This paper\\ \hline
    \multicolumn{2}{c||}{Year}
    &2002
    &2015
    &2018
    &2020
    &2020
    &2020
    &2021
    &-
  \\ \hline \hline
   \multirow{16}{*}{\makecell[c]{Section V:\\Meta \\Learning\\Methodologies}}
   &{\makecell{(V.A.1): Siamese}}
   &
   &            
   & \checkmark
   &  \checkmark          
   &  \checkmark
   & \checkmark
   & \checkmark
   &\checkmark  \\

   \cline{2-10}
   &{\makecell{(V.A.2): Matching}}
   &
   &            
   &  \checkmark
   &   \checkmark         
   &  \checkmark
   &            
   &  \checkmark
   &\checkmark            \\

   \cline{2-10}
   &{\makecell{(V.A.3): Prototypical}}
   &
   &            
   & \checkmark
   &  \checkmark
   & \checkmark
   & \checkmark
   &\checkmark
   &\checkmark            
   \\ \cline{2-10}
   &{\makecell{(V.A.4): Relation}}
   &
   &            
   &  
   &  \checkmark          
   &  \checkmark
   &  \checkmark          
   &  \checkmark
   &\checkmark           \\

   \cline{2-10}

   &{\makecell{(V.B.1): Memory-augmented Neural Network}}
   &
   &            
   &  \checkmark
   &            
   &         
   &  \checkmark
   &  
   &\checkmark           \\

   \cline{2-10}
   &{\makecell{(V.B.2): Meta Network}}
   &
   &            
   &
   &            
   &         
   &
   &
   &\checkmark            
   \\ \cline{2-10}
   &{\makecell{(V.B.3): Recurrent Meta-learner}}
   &
   &            
   &  \checkmark
   &            
   &  
   &         
   & 
   &\checkmark  \\

   \cline{2-10}
   &{\makecell{(V.B.4): Simple Neural Attentive Meta-learner}}
   &
   &            
   &  \checkmark
   &            
   &  
   &   \checkmark
   &  
   &\checkmark            \\

   \cline{2-10}

   &{\makecell{(V.C.1): Model-Agnostic Meta-Learning (MAML)}}
   &
   &  
   &  \checkmark
   &  \checkmark
   &  \checkmark
   &  \checkmark
   & \checkmark
   &\checkmark  \\

   \cline{2-10}
   &{\makecell{(V.C.2): Meta-SGD}}
   &
   &            
   &  
   &            
   &  \checkmark
   &  \checkmark
   &  
   &\checkmark           \\

   \cline{2-10}
   &{\makecell{(V.C.3): Reptile}}
   &
   &            
   &  \checkmark
   &            
   &  
   &  \checkmark
   &  \checkmark
   &\checkmark            \\

   \cline{2-10}
   &{\makecell{(V.C.4): Bayesian MAML}}
   &
   &            
   &
   &            
   &\checkmark
   & \checkmark           
   &
   &\checkmark           \\

   \cline{2-10}
   &{\makecell{(V.C.5):  Laplace
Approximation for Meta Adaptation}}
   &
   &            
   &
   &            
   &
   &  \checkmark          
   &
   &\checkmark          \\

   \cline{2-10}
   
   &{\makecell{(V.C.6): Latent Embedding Optimization}}
   &
   &            
   &
   &            
   &
   &  \checkmark          
   &
   &\checkmark            
   \\ 

   \cline{2-10}
   &{\makecell{(V.C.7): MAML with Implicit Gradients}}
   &
   &            
   &
   &            
   &
   &            
   &
   &\checkmark            
   \\ 

   \cline{2-10}
   &{\makecell{(V.C.8): Online MAML}}
   &
   &            
   &
   &            
   &\checkmark
   &  \checkmark          
   &
   &\checkmark            
   \\ 
   \cline{2-10}
    \hline
   \multirow{3}{*}{\makecell[c]{Section VI:\\Meta Learning\\in Wireless Networks}}
   &{\makecell{(VI.A): Traffic Prediction}}
   &
   &
   &
   &
   & 
   &
   &
   &\checkmark
   \\ \cline{2-10}
   &{\makecell{(VI.B): Rate Maximization}}
   &
   &
   &
   &
   &
   &
   &
   &\checkmark
   \\ \cline{2-10}
   &{\makecell{(VI.C): MIMO Detectors}}
   &           
   &            
   & 
   &           
   &  
   & 
   &
   &\checkmark 
   \\ \cline{2-10}
   \hline
   \end{tabular}}
\end{table*}

\begin{figure*}[!h]
    \centering
    \includegraphics[width=5.7 in]{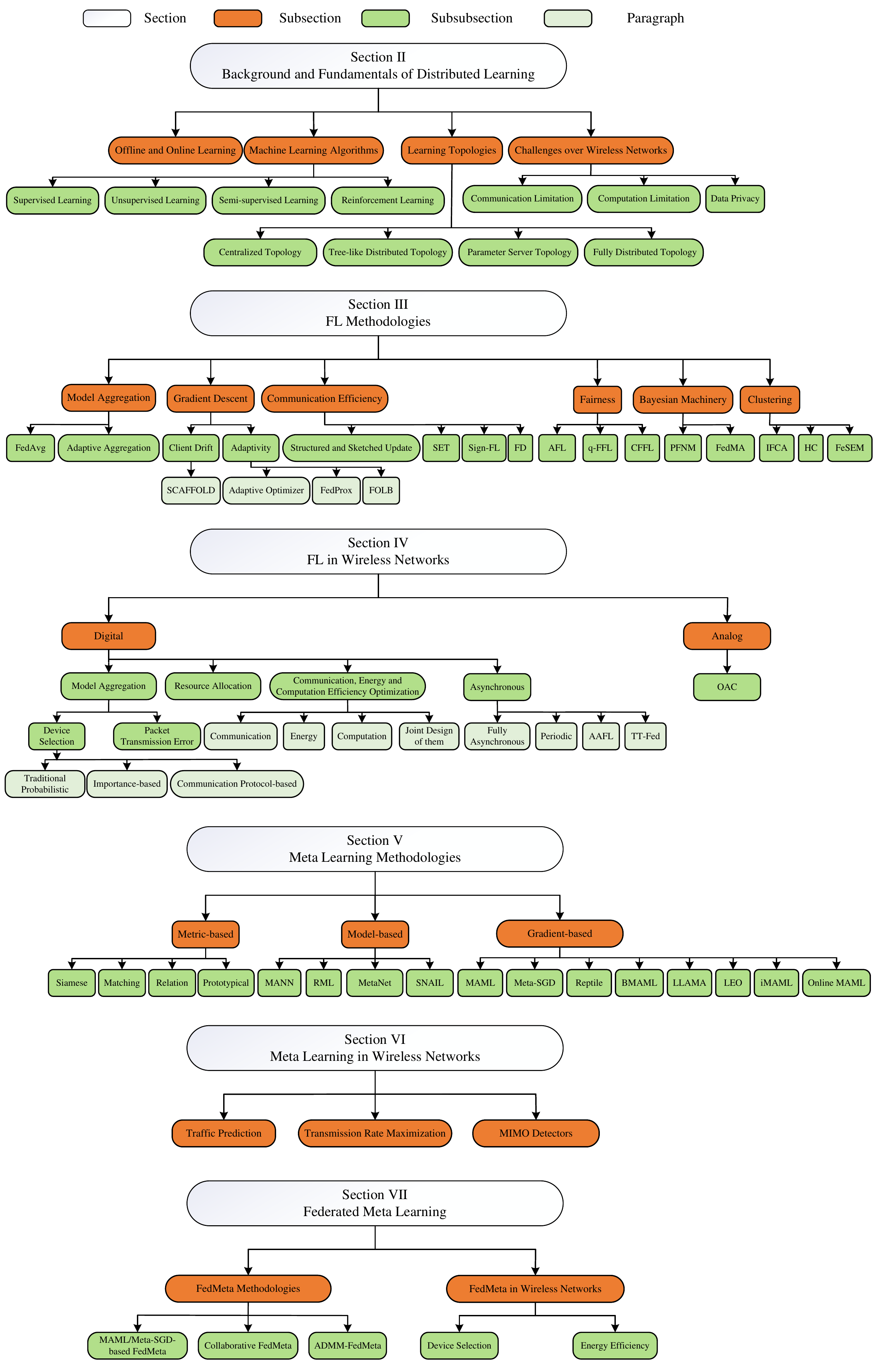}
    \caption{Organization and content of Sections II to VII.}
    \label{basic_modules}
\end{figure*}

Tables II, III, and IV provide a comparative analysis of existing surveys and tutorials in relation to our tutorial. It is observed that the existing literature only covers a limited number of subtopics related to our tutorial and offers only brief descriptions of the corresponding learning algorithms. In contrast, our tutorial goes beyond these limitations by providing a detailed introduction to the underlying design concepts of the relevant algorithms. Notably, our tutorial stands out by analyzing the relationship and evolution of these algorithms specifically and their applications over wireless networks. This crucial aspect has not been extensively investigated in other surveys or tutorials. By exploring the interplay and advancements of these algorithms and their applications over wireless networks, our tutorial contributes novel insights and addresses research gaps in the existing literature.

\subsection{Summary of Contributions}
Given the privacy-preserving characteristics of federated learning (FL) and the ability of meta learning to quickly adapt to different tasks, researchers from both academia and industry are now exploring the joint design of FL, meta learning, and federated meta learning (FedMeta) and wireless networks. This paper provides the first comprehensive tutorial that highlights the research areas, relationships, advancements, challenges, and opportunities associated with these three learning concepts and their applications over wireless network environments. The main contributions of this article can be summarized as follows:
\begin{itemize}
    \item Based on the foundational FedAvg algorithm, we outline six key research areas of FL methodologies, including model aggregation, gradient descent, communication efficiency, fairness, Bayesian learning, and clustering. We provide a detailed analysis of the relationships and evolutionary developments of the respective learning algorithms within these areas. Furthermore, we introduce two research areas that explore the interplay between FL and wireless factors over wireless networks, including digital and analog over-the-air computation schemes.
    \item We summarize three key research areas in meta-learning, including metric-based, model-based, and gradient-based meta-learning. Based on the gradient-based meta-learning paradigm, we focus on the fundamental scheme known as model-agnostic meta-learning (MAML). We discuss the evolution of MAML, providing a detailed analysis of its advancements and applications. Additionally, we explore the potential of meta-learning in solving wireless communication problems. By leveraging meta-learning techniques, we investigate how they can be utilized to tackle various challenges and optimize wireless communication systems.
    \item The fundamental principle of federated meta learning (FedMeta) and its evolution are comprehensively summarized. Furthermore, we introduce two specific research areas of FedMeta over wireless networks, including device selection and energy efficiency. We explore how FedMeta can be applied to address challenges related to device selection, such as optimizing the selection of devices for participation in the FL process. Additionally, we discuss how FedMeta can contribute to enhancing energy efficiency in wireless networks, thereby improving the overall energy consumption of the system.
    \item In addition to the previously discussed topics, we present several other important aspects related to the implementation of FL, meta-learning, and FedMeta. We explore different implementation platforms that facilitate the practical deployment of these learning methodologies. Furthermore, we present real-world applications where FL, meta-learning, and FedMeta have demonstrated their effectiveness and potential impact. In addition, we identify and highlight open research problems that present exciting opportunities for future research in the field of FL, meta-learning, and FedMeta. These research problems have the potential to drive innovation and pave the way for new directions and advancements in these learning concepts. By addressing these open challenges, researchers can contribute to the further development and practical applications of FL, meta-learning, and FedMeta.
\end{itemize}

The rest of this paper is organized as follows. In Section II, we introduce the background and fundamentals of distributed learning. In Sections III and IV, we present important research fields in FL methodologies and their applications over wireless networks, respectively. In Sections V and VI, we introduce research areas in meta learning methodologies and their applications over wireless networks, respectively. Section VII presents principle of FedMeta and its applications over wireless networks. Section VIII introduces open research problems and future directions in FL, meta learning, and FedMeta. A graphical illustration of the content of Sections II to VII is provided in Fig. 2. Finally, conclusions are drawn in Section IX.

\section{Background and Fundamentals of Distributed Learning}
In distributed learning, two main approaches are commonly used for learning across servers or devices: data parallelizing and model parallelizing. In the data parallelizing approach, the data is divided into multiple datasets, and each server or device applies the same machine learning (ML) algorithm to a different dataset. This approach allows for parallel processing of data, enabling faster training and improved scalability. On the other hand, the model parallelizing approach involves segmenting the ML model into different sub-models. Each sub-model is updated on different servers or devices, and therefore global model is therefore the aggregation of all sub-models. However, it is important to note that not all ML algorithms are compatible with the model parallel approach due to the specific requirements and dependencies of the model. It is also worth mentioning that data parallelizing and model parallelizing can be employed simultaneously, where both the data and the model are distributed across multiple servers or devices. This hybrid approach leverages the advantages of both techniques to achieve improved performance and efficiency in distributed learning scenarios \cite{Daiwei}. In this section, we will provide a comprehensive description of the background and fundamentals of distributed learning, including offline and online learning, learning topologies, challenges of distributed learning in wired/wireless networks, and application of FL in ML paradigms.

\subsection{Offline and Online Learning}
Based on whether the learning model is updated with newly arriving data \cite{S_C_Hoi}, distributed learning algorithms can be categorized into offline learning and online learning. 
Offline learning refers to the process of updating the learning model using the knowledge derived from previous observations. The objective of offline learning is to maximize accuracy or long-term reward for prediction or decision-making tasks, leveraging the information gathered from historical data. On the other hand, online learning involves continuously updating the learning model in response to newly arriving data and observations from the environment. The primary goal of online learning is to optimize prediction or decision-making performance in real-time, adapting the model to changing circumstances. Both offline and online learning methods are designed to improve the learning accuracy and long-term reward of prediction or decision-making tasks. While offline learning focuses on leveraging historical observations, online learning emphasizes adaptability to real-time data streams and the ability to quickly update the model to capture changing patterns and dynamics in the environment.

Offline learning consists of two phases, namely, the training phase and the testing phase. In the training phase, the learning model is first trained with the training dataset until an optimal set of hyperparameters achieving the highest accuracy or reward for the given task. In the testing phase, the trained learning model is employed for prediction or decision-making without any further updates. The model utilizes its learned knowledge to make predictions or decisions on new data. The offline training method suffers from high re-training costs when dealing with new training data/environments, and thus, has poor scalability for real-world applications, especially when the amount of data grows and the environment evolves rapidly \cite{Prudencio}. 

Online learning can be applied to a wide range of ML algorithms, where the learning models are trained to handle prediction or decision-making tasks by continuously learning from a sequence of data samples in a sequence manner, one by one, in each time slot. Online learning addresses the limitations of offline learning by enabling the learning models to be updated constantly and efficiently as new training data becomes available. This constant updating of the learning models in online learning makes them highly efficient and scalable for large-scale ML tasks in real-world applications. By incorporating new training data in a timely manner, online learning models can adapt to evolving environments and changing data distributions, enhancing their ability to handle real-time and dynamic scenarios \cite{Benczur}.

To efficiently solve prediction/decision-making problems, the servers and devices need to be connected and communicate with each other for information exchange, which can be represented via the topology of the distributed ML systems. In the following subsections, various types of topologies are introduced in detail.

\begin{figure*}[ht]
	\centering
	\subfloat[Centralized]{\includegraphics[width=2.3in]{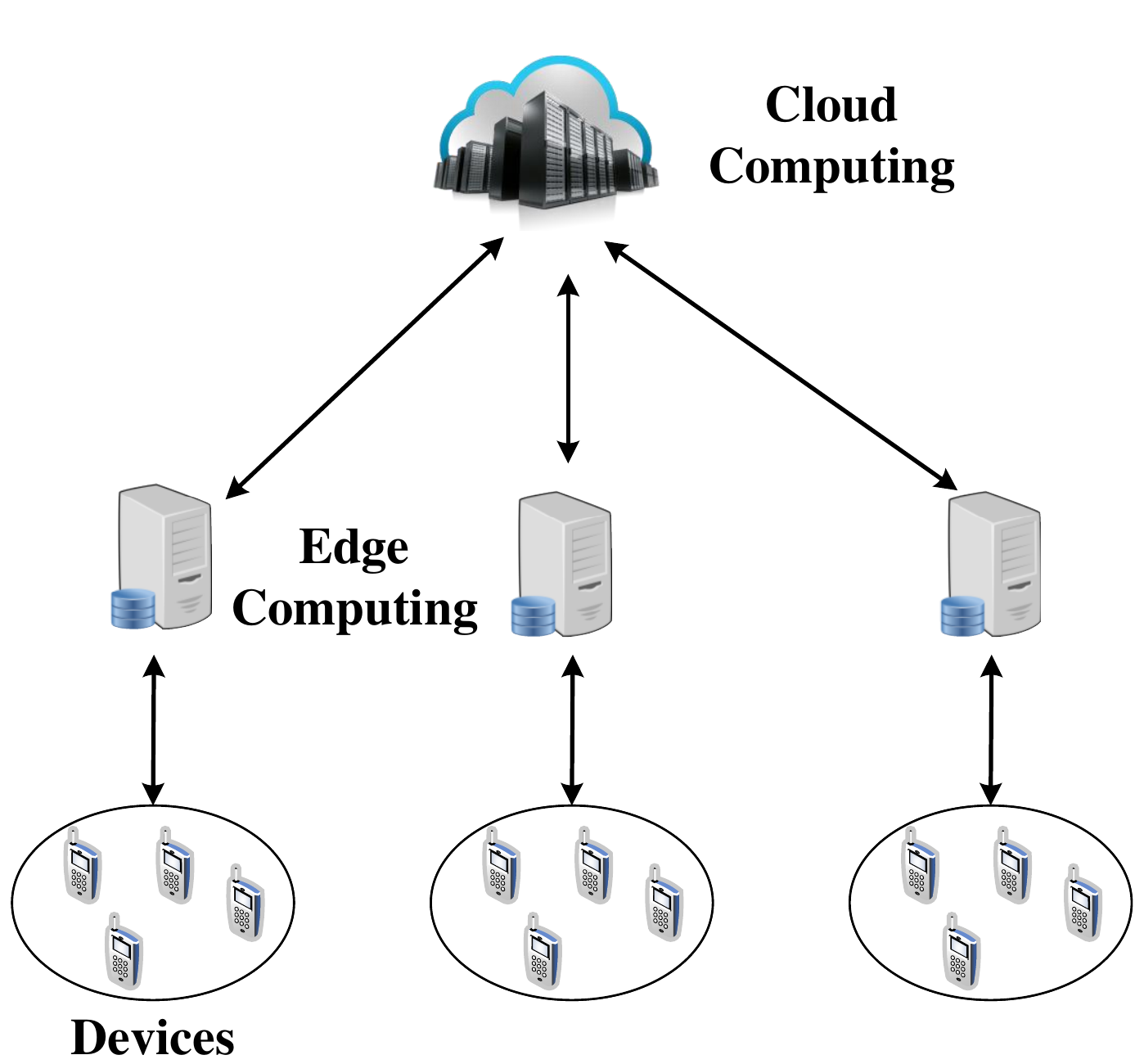}}\label{fig_first_case}
	\subfloat[Decentralized (Tree)]{\includegraphics[width=4.5in]{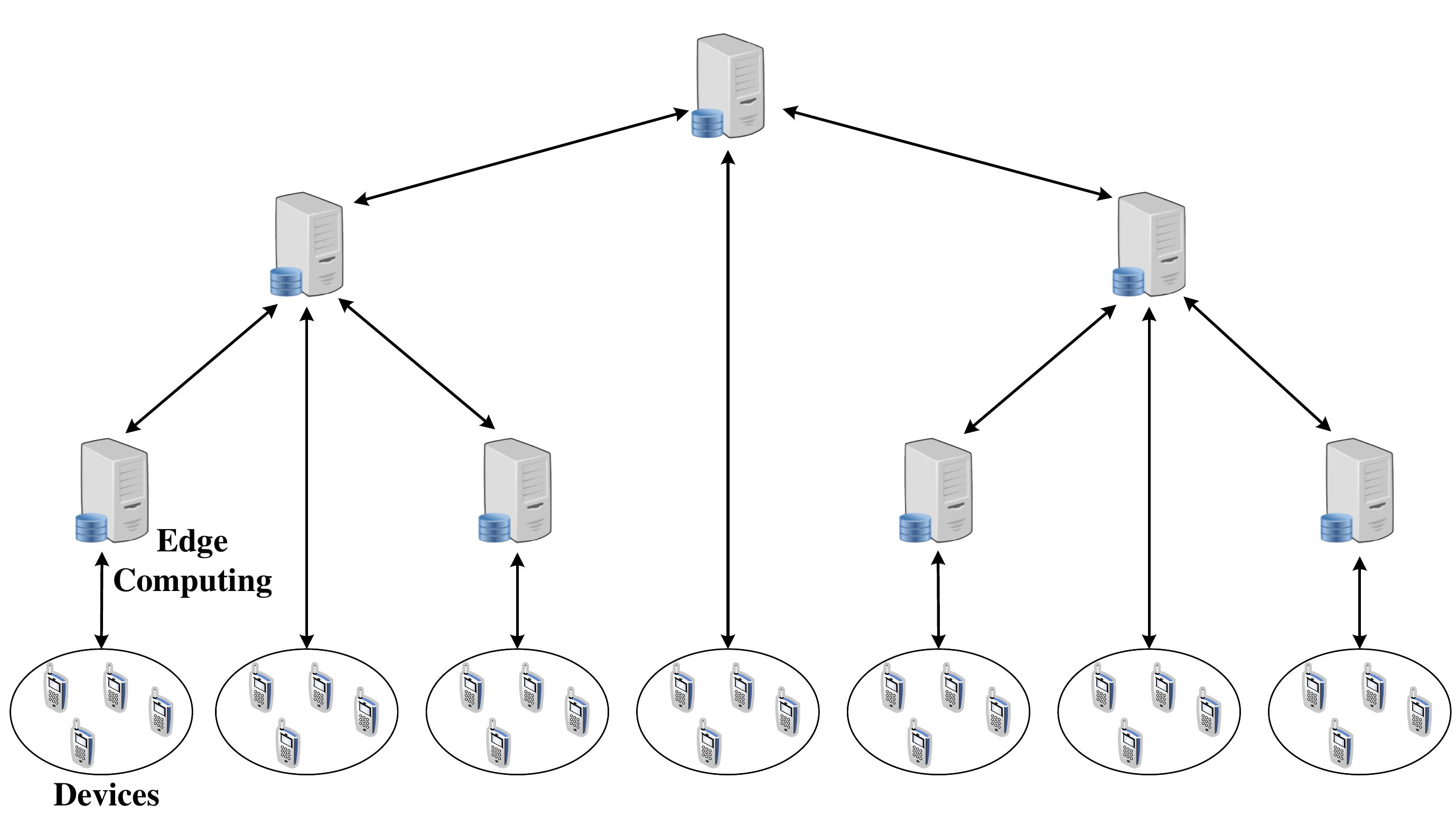}}\label{fig_second_case}
	\subfloat[Decentralized (Parameter Server)]{\includegraphics[width=3.0in]{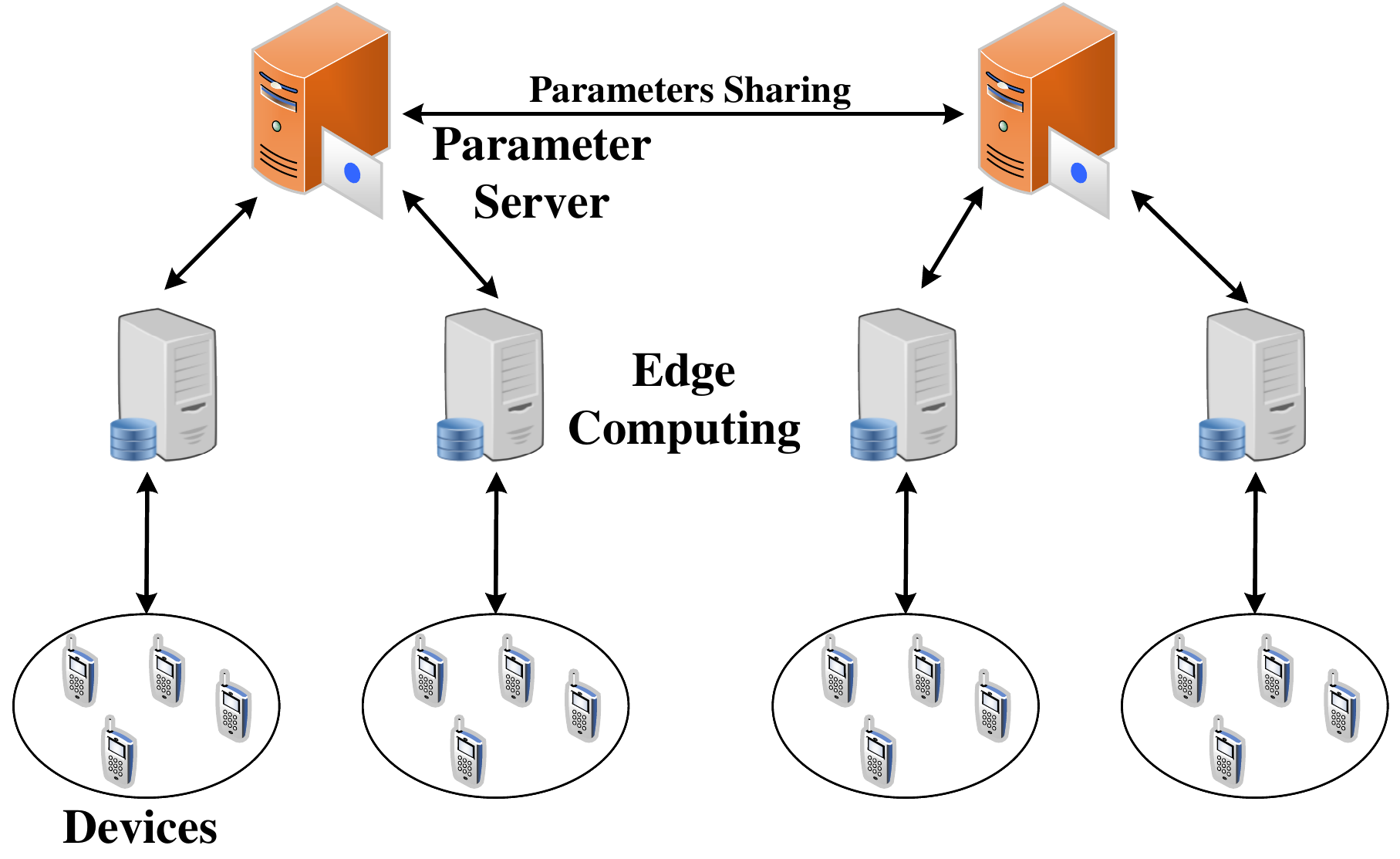}}\label{fig_first_case}
	\subfloat[Fully Distributed]{\includegraphics[width=3.0in]{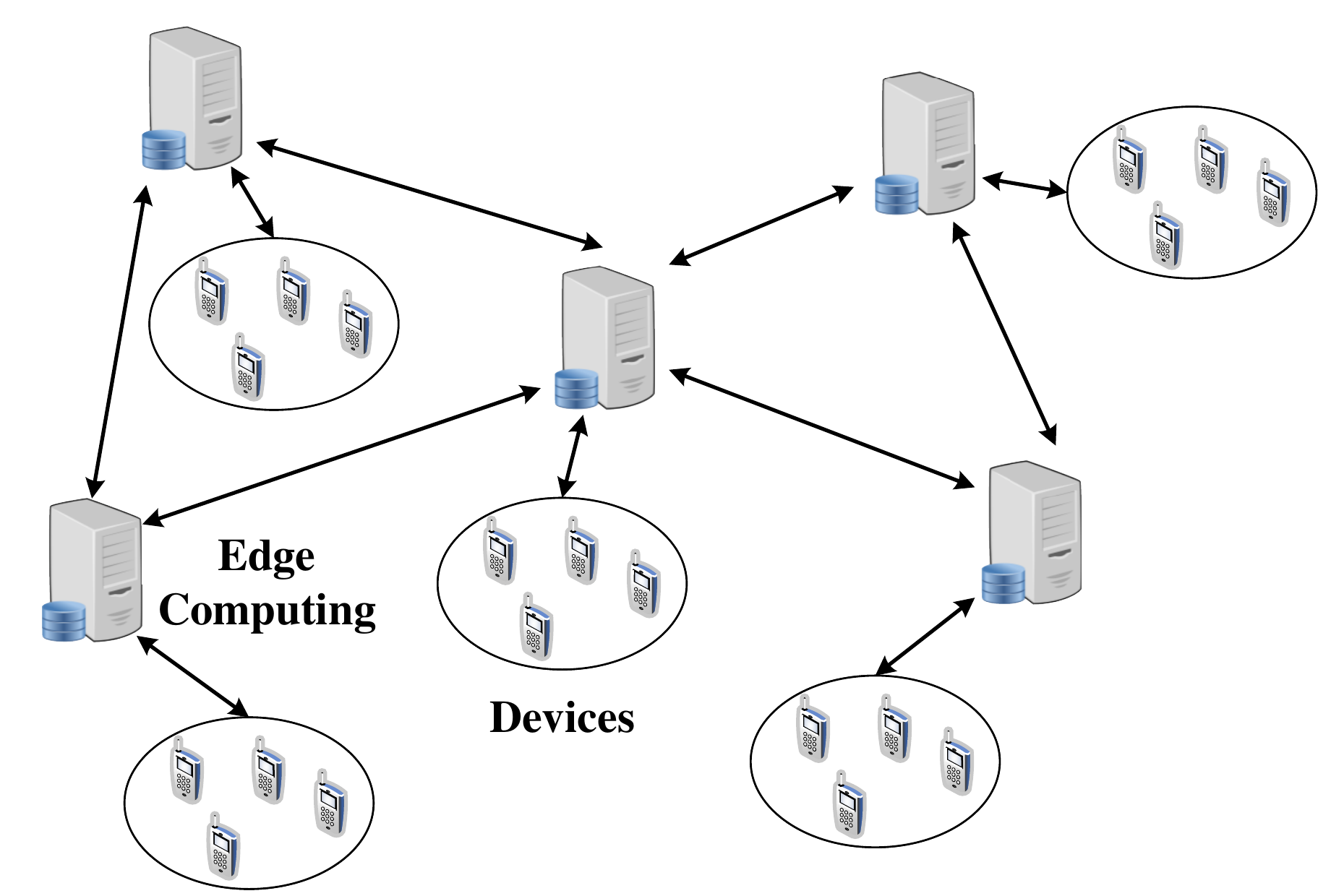}}\label{fig_second_case}
	\caption{Centralized and distributed machine learning topologies.}
	\label{basic_modules}
\end{figure*}

\subsection{Learning Topologies}
The selection of topology plays an important role in designing distributed ML systems. The topology determines how servers and devices are organized and connected within the system. Various factors, including the degree of distribution in transmitting learning models, influence the selection of a suitable topology. In this subsection, four general learning topologies in distributed ML are introduced and shown in Fig. 3, as proposed in \cite{Paul_Baran} and \cite{Joost_Verbraeken}.

\subsubsection{Centralized Topology} The centralized topology is characterized by a central server that is connected to multiple edge servers to obtain learning models, as shown in Fig. 3 (a). The central server first obtains learning models from all edge servers, and the the model aggregation is executed at the central server. Next, the central server transmits the aggregated learning model to all edge servers. Each edge server then broadcasts the learning model to the devices it is connected to, and each device updates the transmitted learning model based on its local dataset.

\subsubsection{Tree-like Distributed Topology} The tree-like distributed topology is characterized by a hierarchical arrangement of connected nodes, resembling the branches of a tree, as shown in Fig. 3 (b). There is only one connection between any two connected nodes, establishing a natural parent and child hierarchy. Therefore, parameters are shared between parent and child nodes. The tree-like distributed topology offers advantages in scalability and manageability. Communication is simplified since each edge server or device only needs to interact with its parent and child servers, reducing the overall complexity of information exchange and coordination \cite{Alekh_Agarwal}. In this topology, the edge servers within the tree structure accumulate the local gradients computed by their child servers. These accumulated gradients are then passed up to their parent servers to calculate a global gradient. This process enables collaborative gradient computation across the tree-like structure, allowing the learning models to collectively update and optimize based on the information exchanged between parent and child nodes.

\subsubsection{Parameter Server Topology} The parameter server topology consists of a distributed set of edge servers and a centralized set of parameter servers responsible for maintaining and sharing the models \cite{J_Wei,M_Li}, as shown in Fig. 3 (c). All learning models are stored in the parameter servers using a global shared memory. Edge servers have direct access to the models stored at the parameter servers and can obtain and update the models as needed in any time slot. Meanwhile, edge devices communicate and share parameters with edge servers. The advantage of this topology lies in the direct accessibility of models for all edge servers. However, one drawback of this topology is the communication overhead. As the parameter servers handle all communication, there will be a large amount of communication overhead between the edge servers and the parameter servers \cite{Jinliang}.

\subsubsection{Fully Distributed Topology} The fully distributed topology does not have any central servers, as shown in Fig. 3 (d). Instead, it consists of a set of independent edge servers, each responsible for updating its own learning model and directly exchanging its model with other edge servers. One of the key advantages of the fully distributed topology is its high scalability. Additionally, the fully distributed topology exhibits robustness against failures of individual servers, such as power outages or malfunctions \cite{Ian_Foster}. However, it's important to note that the fully distributed topology can incur significant communication overhead, especially when a large number of edge servers are present in the network.

To solve the high communication overhead problem in distributed topologies, a fast, and communication-efficient distributed framework, so-called group alternating direction method of multipliers (GADMM), was proposed in \cite{GADMM1,GADMM2,GADMM3}. In GADMM, at most half of the edge servers are competing for the limited communication resources in each time slot. Meanwhile, each edge server exchanges the trained model only with two neighboring edge servers, thereby training a global model with a lower amount of communication overhead in each exchange.

\subsection{Challenges of Distributed Learning}
In wired/wireless networks, it is obvious that the overall communication and computation overhead increases as the number of servers and devices grows. To develop effective distributed ML algorithms for wired/wireless networks, three primary factors should be considered, including the communication cost, computation cost, and data privacy. 

\subsubsection{Wireless Communication Limitation}
In wireless networks, a number of servers and devices may share the same spectrum resource due to the limited bandwidth \cite{Kibria}. Therefore, the communication among the devices and edge servers may suffer from high interference, poor channel conditions, and noise, which lead to low reliability, high transmission latency, and low learning accuracy \cite{Jihongpark}.

\subsubsection{Computation Limitation}
Training and operating ML algorithms usually require computation units with high processing capability, especially when the ML models are complex. However, the devices have limited computation and energy capabilities. To minimize the computation latency at the device side, the use of edge servers with high processing capability using graphics processing units (GPUs) have been recently proposed to move the computations from the device to the edge. 

\subsubsection{Data Privacy}
Transmitting datasets of edge devices to edge servers can cause data breach if the datasets have privacy-sensitive information, one potential solution is to only exchange the weights of ML models. Nevertheless, it is possible that the transmitted model parameters can be reversely traced, so that the privacy is still not preserved \cite{Fredrikson}.

In order to address the challenges related to communication, computation, and data privacy, FL has emerged as an effective approach to exploit the distributed devices to collaboratively train ML models. FL was first introduced by Google in 2016, where multiple devices jointly train an ML model without sharing their private data, under the supervision of a central server. This ensures the privacy of the training data of all devices. FL has two entities: a centralized server that owns the global model and a set of devices that store the local models and training datasets. Meanwhile, FL consists of four procedures \cite{Nuria}: 1) training the local model based on the local dataset at the local device; 2) transmitting the local models from the devices to the central server; 3) aggregating the local models to a global model at the central server; and 4) updating the received global model from the central server at devices. The original data of each device is kept locally and does not need to be exchanged or migrated between devices, which ensures the privacy of each device. As a result, devices can benefit from the advantages of shared models trained by other devices without data sharing.

\subsection{Application of FL in Machine Learning Paradigms}
ML algorithms have the ability to make predictions or decisions by learning from datasets or observed states in the environment. To train a learning model, feedback is required to iteratively improve the learning model. ML algorithms can be classified into different categories based on the type of feedback they receive, including supervised learning, unsupervised learning, semi-supervised learning, and reinforcement learning algorithms.

\subsubsection{Supervised Learning} Supervised learning, one of the fundamental ML approaches, involves training a function that can map inputs to outputs based on labeled data. Each training data sample consists of an input object, typically represented as a vector, along with its corresponding desired output value. The objective of the learning process is to minimize the error between the true data label and the predicted output. By analyzing the training data, the learning model can generalize and accurately predict labels for new datasets. Unfortunately, it is worth noting that as the volume of training data increases, more complex models may be required to achieve accurate predictions \cite{M_Mohri,Stuart}. FL is commonly used in tasks where labeled data is readily
available, which aligns well with supervised learning tasks. However, considering the challenge of limited labeled data in FL scenarios, the exploration of unsupervised and semi-supervised FL algorithms becomes essential.

\subsubsection{Unsupervised Learning} Unsupervised learning is an ML approach that involves learning from unlabeled data samples. It utilizes ML algorithms to analyze and cluster unlabeled datasets, which can discover hidden patterns or data groupings without the need for human intervention. One prominent example of an unsupervised learning technique is Principal Component Analysis (PCA) \cite{Tharwat}. PCA transforms high-dimensions data into lower-dimensions and it combines the original features into a new feature space by a projection direction that most information in the new feature space is retained from the original data \cite{G_Hinton}. A common application of unsupervised learning is clustering. In clustering, the learning model automatically groups the training data into groups with similar features. In \cite{FedUL}, a federation of unsupervised learning (FedUL) was proposed to verify the possibility of unsupervised FL, where the unlabeled data were transformed into surrogate labeled data for each device, a modified model was trained by supervised FL, and the desired model was recovered from the modified model. FedUL is a very general solution to unsupervised FL. It is compatible with many supervised FL methods, and the recovery of the wanted model can be theoretically guaranteed as if the data have been labeled. In \cite{NEURIPS2020_47a65822}, federated PCA in a memory-limited setting was proposed, which provided robustness against stragglers. In \cite{Yiuming}, federated PCA for vertically partitioned dataset method was considered, which reduced the dimensionality across the joint datasets over all devices and extracted the principal component feature information for downstream data analysis.

\subsubsection{Semi-supervised Learning}
Semi-supervised learning is a learning approach that leverages both labeled and unlabeled data to perform learning tasks, and is conceptually sitting between supervised and unsupervised learning. By utilizing a large amount of unlabeled data in combination with smaller sets of labeled data, semi-supervised learning enables the learning model to accurately classify unlabeled data, leading to a significant improvement in learning accuracy  \cite{Xiaojin}. In \cite{SemiFed}, semi-supervised FL (SemiFed) was proposed to unify two dominant approaches for semi-supervised learning, which are consistency regularization and pseudo-labeling. SemiFed first performs consistency
regularization, which encourages the network to produce similar output distributions when its inputs are perturbed. This regularization can be applied to all samples without labels. Following several rounds of training, the concept of pseudo-labeling is employed. Pseudo-labeling involves assigning artificial labels to unlabeled images and then training the model to predict these artificial labels when fed with unlabeled samples as input in the following training stages.

\subsubsection{Reinforcement Learning} Reinforcement learning (RL) is an ML approach in which agents make decisions based on the observations obtained from the environment, aiming to maximize the long-term reward \cite{Junyan}. Unlike supervised learning, RL does not rely on labeled input and output data. Instead, it focuses on striking a balance between exploration (of unknown territory) and exploitation (of current knowledge), and training  ML models to make a sequence of decisions  \cite{Kaelbling}. However, implementing RL in practical scenarios has several challenges. For example, it is impossible to explore the entire state-action spaces comprehensively. Although distributed RL algorithms can help solve the problem, they usually require data collection from each agent, which may compromise agent privacy and lead to information leakage. To address these privacy concerns, federated reinforcement learning (FRL) has been proposed \cite{jiajuqi}. FRL not only enables agents to learn optimal decisions in unknown environments but also ensures that the privately collected data during an agent’s exploration does not need to be shared with others.

In the following two sections, we introduce several main research areas in FL methodologies and their applications over wireless networks. In particular, we first present six FL research areas in FL methodologies. Then, we present two FL research areas in wireless networks.

\section{FL Methodologies}
In this section, we introduce six main FL research aspects of FL methodologies, including model aggregation, gradient descent, communication efficiency, fairness, Bayesian machinery, and clustering. The transmission between servers and devices is error-free without considering any wireless factors. For ML algorithms, given a labeled set of inputs and their corresponding outputs, the learning models are trained and tested, as shown in Fig. 4. The goal of ML algorithms is to minimize the loss function designed based on a specific learning problem. The loss function plays an important role in training ML algorithms and improvement of their performance \cite{Q_Wang}, and is expressed as
\begin{equation}
    \mathcal{L}(\bm{w}) = \frac{1}{N}\sum\limits_{i=1}^{N}l_i(\bm{w}),
\end{equation}
where $N$ is the number of data samples, and $l_i(\bm{w})$ is the loss of the $i$th input data based on learning weights $\bm{w}$. To find the minimum value of the loss function, gradient descent is introduced to calculate the derivative of the loss function via $\frac{\partial L(\bm{w})}{\partial \bm{w}}$. Then, the weights of the learning models are updated as
\begin{equation}
    \bm{w} = \bm{w} - \eta\frac{\partial \mathcal{L}(\bm{w})}{\partial \bm{w}},
\end{equation}
where $\eta$ is the learning rate. After a sufficient number of iterations, the loss function can achieve its minimum value, and the ML model converges. The gradient descent algorithm is presented in \textbf{Algorithm 1}.

\begin{algorithm}[t]
\begin{algorithmic}[1]
\caption{Gradient Descent Algorithm}
\STATE Initialize learning weights $\bm{w}$ and learning rate $\eta$.
\FOR{Iteration = 1,...,$I$}
    \FOR{$i$ = 1,...,$N$}
        \STATE Input the data $x_i$.
        \STATE Obtain the loss $l_i(\bm{w})$.
    \ENDFOR
    \STATE Calculate the loss function $\mathcal{L}(\bm{w})$ for the input data based on (1).
    \STATE Update the learning weights $\bm{w}$ based on (2).
\ENDFOR
\end{algorithmic}
\end{algorithm}

\begin{figure}[!h]
    \centering
    \includegraphics[width=3.5 in]{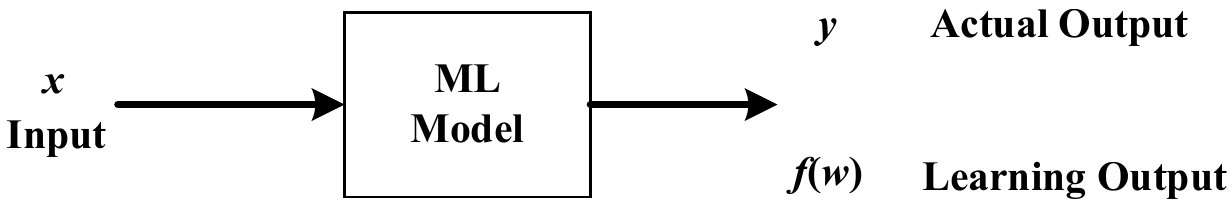}
    \caption{The architecture of the ML process.}
    \label{basic_modules}
\end{figure}

\begin{figure}[!h]
    \centering
    \includegraphics[width=3.5 in]{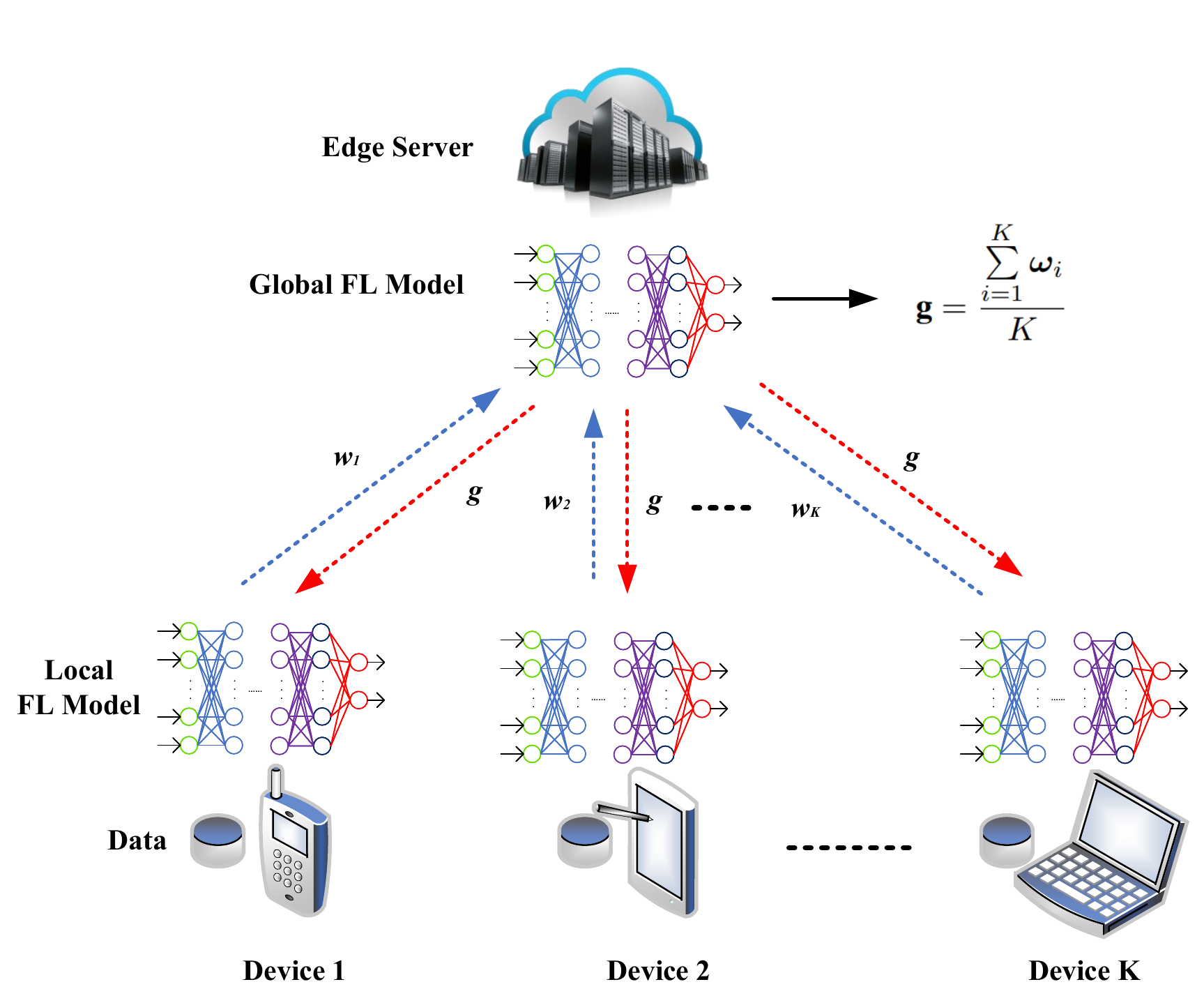}
    \caption{The canonical architecture of FedAvg.}
    \label{basic_modules}
\end{figure}

FL, as first proposed by McMahan in \cite{B_McMahan}, uses the loss function and gradient descent algorithm based on (1) and (2) to collaboratively learn a shared learning model while keeping all the training data at the devices. 

We assume that $\hat{K}$ devices have their own datasets $\{D_1,...,D_{\hat{K}}\}$, and each of them cannot access to the other devices' datasets. As shown in Fig. 5, FedAvg can learn a model by aggregating learning models from distributed devices. In each iteration, $K (K\leq\hat{K})$ devices are selected, and each device first performs the local gradient descent for its own dataset. Then, the local updated model weights $\bm{w}_k (k=1,2,...,K)$ are transmitted to the server. The server aggregates the local models $\{\bm{w}_1,...,\bm{w}_K\}$ to a global model $\bm{w}_G$, and transmits the global model to all devices to replace the local model \cite{H_B_McMahan}.

\subsection{Model Aggregation}
Model aggregation is the process of integrating models from multiple devices in order to create a new model \cite{Amedeo,Kittler,Ho}. There are several ways for model aggregation as explained in the following.

\textbf{Federated Averaging:} The most basic model aggregation method is the Federated Averaging (FedAvg) algorithm proposed in \cite{B_McMahan}, where the weights of local models are averaged at the central server to update the global model \cite{Sannara}. We define $p_k$ as the percentage of the number of data samples at the $k$th device over the total number of data samples at all devices, $n_k$ as the number of data samples of dataset $D_k$, and $n$ as the total number of data samples, which is calculated as $n = \sum_{k=1}^{K}n_k$. The training objective of FedAvg is given as follows \cite[Eq. (1)]{B_McMahan}
\begin{equation}
    \min_{\bm{w}}~\mathcal{L}(\bm{w}_{G}) = \sum\limits_{k=1}^{K}p_kf_{k}(\bm{w}_k),
\end{equation}
where
\begin{equation}
p_k = \frac{n_k}{n}~~\text{and}~~ f_{k}(\bm{w}_k) = \frac{1}{n_k}\sum_{i\in D_k}l_{i}(\bm{w}_k).
\end{equation}
In (4), $l_{i}(\bm{w}_k)$ is the loss of the FL model due to the $i$th data sample calculated by the local model weights $\bm{w}_k$ of the $k$th device, and $\bm{w}_G$ is the weight of the global model. 

There are two approaches to update the global model. The first approach is to compute the gradient of each device. Then, the central server aggregates these gradients from $K$ devices and updates the global model using
\begin{equation}
    \bm{w}_{G}^{t+1} = \bm{w}_{G}^{t} - \eta\sum_{k=1}^{K}p_k\textbf{g}_k^t,~~\text{with}~\textbf{g}_k^t = \nabla f_{k}(\bm{w}_k^t),
\end{equation}
where $\nabla f_{k}(\bm{w}_k^t)$ is the gradient computed at the $k$th device.
The second approach is to update the weights of the local model at each device using
\begin{equation}
    \bm{w}_k^{t+1} = \bm{w}_G^t - \eta \textbf{g}_k^t,
\end{equation}
where $\textbf{g}_k^t$ is the gradient calculated using (5). Then, the global model at the central server is updated as
\begin{equation}
    \bm{w}_{G}^{t+1} = \sum\limits_{k=1}^{K}p_k\bm{w}_k^{t+1}.
\end{equation}
In the second approach, each device first performs the gradient descent for its local model using the local datasets, and the central server averages these local models. In this way, each device can iterate the local update in (6) multiple times before uploading local models, which can accelerate the convergence speed. 

Although FedAvg has achieved great success and is one of the most well-known algorithms in FL, the statistical heterogeneity challenges in the data are still difficult to overcome. That is to say, the training data follow a non-independent and non-identical distribution (non-IID), which negatively affects the convergence behavior. To address this issue, adaptive aggregation is introduced. 

\textbf{Adaptive Aggregation:}
Different from the FedAvg, adaptive aggregation uses a different way to update the global model. To improve the accuracy and convergence performance of the global model, a temporally weighted aggregation method utilizing the previously trained local models was proposed in \cite{yangchen}. The authors in \cite{yangchen} assumed that the local models updated in the $(t-i)$th time slot $(i=1,...,t-1)$ are less important than those updated in the $t$th time slot. In practice, the training data at each device changes over time, and the local models that are more recently updated have a higher importance during model aggregation. In order to account for the freshness of the local models, the global model is updated using \cite[Eq. (1)]{yangchen}
\begin{equation}
    \bm{w}_{G}^{t+1} = \sum\limits_{k=1}^{K}p_k\left(\frac{e}{2}\right)^{-(t-\hat{t}^{k})}\bm{w}_k,
\end{equation}
where $p_k$ is given in (4), $e$ is the scalar constant used to represent the time effect, and $\hat{t}^{k}$ is the time slot in which the newest $\bm{w}_k$ is updated. By introducing parameters $\left(\frac{e}{2}\right)^{-(t-\hat{t}^{k})}$, the impact of the local models updated in previous time slots is reduced, and the global model updated in the current time slot is weighted to be more important for the new data, which results in higher learning accuracy. 

In \cite{Yeganeh}, an adaptive weighting approach, namely, Inverse Distance Aggregation (IDA), was proposed. The global model updating still follows (5) or (7). Compared with the FedAvg, the main difference of the IDA is the manner in which the weighting coefficient $p_k$ is calculated, which is based on the inverse distance of the local model weights to the global model weights. To realize this, the $l_1$-norm is used as a metric to measure the distance between the weights of the local model of the $k$th device $\bm{w}_{k}$ and that of the global model $\bm{w}_{G}$, and $p_k$ is calculated as
\begin{equation}
    p_k = \frac{\|\bm{w}_{G}^{t} - \bm{w}_{k}^{t}\|_{1}}{\sum\limits_{k=1}^{K}\|\bm{w}_{G}^{t} - \bm{w}_{k}^{t}\|_{1}}.
\end{equation}
Calculating $p_k$ via (9) allows us to give higher weight to devices whose distance between the weights of the local model and the weights of the global model are higher. It is important to note that the IDA approach is based on the assumption that devices with more data should have a greater contribution in updating the weights to the local model. 

A novel layer-wise adaptive aggregation scheme was proposed in \cite{Shaoxiongji} to iteratively update weights while attempting to reduce the distance between the global model and local models. Although the global model is still updated using either (5) or (7), $p_k$ in each layer in \cite{Shaoxiongji} is calculated to minimize the distance  between each layer of the local model and each layer of the global model. For the $l$th layer, the weighting coefficient $p_{k,l}$ is calculated as
\begin{equation}
    p_{k,l} = \text{softmax}(s_{k,l}^{t}) =  \frac{e^{s_{k,l}^{t}}}{\sum\limits_{i=1}^{K}e^{s_{i,l}^{t}}},
\end{equation}
where
\begin{equation}
    s_{k,l}^{t} = \|\bm{w}_{G,l}^{t} - \bm{w}_{k,l}^t\|_{d}.
\end{equation}
Here, $d$ refers to the $l_d$-norm, which is used to calculate the distance $s_{k,l}^{t}$ between the $l$th layer of the global model and the $l$th layer of the local model of the $k$th device. The softmax function in (10) is applied to guarantee $p_{k,l}$ in the range of $0$ to $1$. This is because the softmax function is a function that converts a vector of $\tilde{N}$ real values into a vector of $\tilde{N}$ real values that sum to 1 \cite{C_Nwankpa}. The advantage of the layer-wise adaptive FL is that it can minimize the distance between the global model and the local models. The summary of FL algorithms in model aggregation is summarized in Table V.

\begin{table*}
\centering
\caption{Summary of FL Algorithms in Model Aggregation}
\begin{tabular}[c]{c|c|c|c}
\hline
\hline Model Aggregation Algorithm & Advantage & Disadvantage & Condition \\
\hline Federated Averaging \cite{B_McMahan} & Guarantee device privacy & \makecell{Poor adaption to system \\and statistical heterogeneity} & Synchronous mode\\
\hline Temporally Weighted Aggregation \cite{yangchen} & \makecell{Low communication costs\\High learning accuracy} & Poor adaption to asynchronous mode & Synchronous mode \\
\hline Inverse
Distance Aggregation \cite{Yeganeh} & \makecell{Adapt to statistical heterogeneity\\High learning accuracy} & \makecell{Not robust to low quality\\ and poisonous data} & \makecell{Synchronous mode \\ Devices have enough data}  \\ 
\hline Layer-wise Adaptive Aggregation \cite{Shaoxiongji} & \makecell{Consider the relation between \\the server and device models\\Server model is well-generalized\\Low communication costs} & \makecell{Poor adaption to asynchronous mode\\Performance on non neural \\ language model is unknown} & \makecell{Execute on neural \\ language model} \\
\hline
\hline
\end{tabular}
\end{table*}

\subsection{Gradient Descent}
Standard federated optimization methods, such as FedAvg \cite{B_McMahan}, may show unfavorable convergence performance, especially in heterogeneous networks. This is mainly caused by two factors: 1) client drift, where the local models move away from the optimal global model, which can lead to unstable and slow convergence; and 2) lack of adaptivity, where the FedAvg may be unsuitable for datasets with heavy-tailed stochastic gradient noise distributions, this often happens in NLP research \cite{Jingzhao_Zhang}. Heavy-tailed distributions are probability distributions whose tails are not exponentially bounded, that is to say, they have heavier tails than the exponential distribution \cite{Eduard}. Several novel gradient descent methods have been proposed to solve the client drift  and lack of adaptivity problems, which are introduced in the following:

\subsubsection{Client Drift} To mitigate the problem of client drift, a new Stochastic Controlled Averaging algorithm (SCAFFOLD) was proposed in \cite{S_Karimireddy}, where control variates for the $k$th device $\textbf{c}_k$ and the variate for the server $\textbf{c}_G = \frac{1}{K}\sum_{i=1}^{K}\textbf{c}_i$ were used in the gradient descent to update the local and global models, respectively. Unlike the FedAvg, the gradient descent of the $k$th device of the SCAFFOLD algorithm is given by
\begin{equation}
    \bm{w}_k^{t+1} = \bm{w}_k^{t} - \eta(\textbf{g}_k^t + \textbf{c}_G^{t} - \textbf{c}_k^{t}),
\end{equation}
where $\textbf{c}_G^{t} - \textbf{c}_k^{t}$ guarantees the gradient descent moving towards the right direction, and $\textbf{c}_k^{t+1}$ is calculated using
\begin{equation}
    \textbf{c}_k^{t+1} = \textbf{c}_k^{t} - \textbf{c}_G^{t} + \frac{1}{\tilde{N}_k\eta}(\bm{w}_G^{t} - \bm{w}_k^{t}).
\end{equation}
In (13), SCAFFOLD uses the previous computed gradients to update the control variate, and $\tilde{N}_k$ is the number of updating iteration of the $k$th device with its local data in the $t$th time slot. Then, the global control variate $\textbf{c}_G$ is aggregated as
\begin{equation}
    \textbf{c}_G^{t+1} = \textbf{c}_G^t + \frac{1}{K}\sum_{i=1}^{K}(\textbf{c}_i^{t+1} - \textbf{c}_i^{t}).
\end{equation}
The correction term $(\textbf{c}_G - \textbf{c}_k)$ in (12) ensures that the local model updates move towards the optimal direction, so as to address the client drift issue of FedAvg.

\subsubsection{Adaptivity} An adaptive learning algorithm has adaptive learning parameters, such as learning rate, which can automatically adjust the statistics of the received data, available computational resources, or other information related to the environment in which it operates. Adaptive variants can help to learn algorithms to improve the convergence performance and learning accuracy \cite{Goodfellow}. To improve the convergence performance of FedAvg, three methods have been proposed, namely, Adaptive Optimizer, Federated Proximal (FedProx), and Fast-convergent FL (FOLB).

\paragraph{Adaptive Optimizer} The SGD in FedAvg may be unsuitable for settings with heavy-tailed stochastic gradient noise distributions. To address this issue, traditional adaptive optimization algorithms, such as Adagrad, Adam, and Yogi have been integrated into FL to update the global model in \cite{Sashank}. The global model is updated as
\begin{equation}
    \bm{w}_G^{t+1} = \bm{w}_G^{t} + \frac{\eta}{\sqrt{v^t} + \tau}\hat{\bm{w}}_t,
\end{equation}
where $v^t = \{v_{\text{Adagrad}}^{t},  v_{\text{FedAdam}}^{t}, v_{\text{FedYogi}}^{t}\}$ are exponential moving averages (EMAs) of the gradients of the FedAdagrad, FedAdam, and FedYogi optimization methods, $\eta$ is the learning rate, and $\tau$ controls the degree of adaptivity of the algorithm, where smaller values of $\tau$ denotes higher degrees of adaptivity. EMA gives a higher weight to the most recent data points. In (15), $\hat{\bm{w}}_t$ is the sum of the past gradient differences between the local and global model for given decay parameters $\beta_1\in[0,1)$ and $\beta_2\in[0,1)$, and is calculated as
\begin{equation}
    \hat{\bm{w}}_t = \beta_1\hat{\bm{w}}_{t-1} + (1-\beta_1)\left(\frac{1}{K}\sum\limits_{i=1}^{K}\triangle\bm{w}_i^t\right).
\end{equation}
In (16), the difference $\triangle\bm{w}_i^t$ between the local and global model of the $i$th device in the $t$th time slot is calculated as
\begin{equation}
    \triangle\bm{w}_i^t = \bm{w}_i^{t} - \bm{w}_G^{t}.
\end{equation}
FedAdam was proposed to address the problem of the rapid decay of the learning rate of the FedAdagrad optimization algorithm. However, FedAdam leads to a situation where the past gradients are forgotten in a fairly fast manner, which can be especially problematic in sparse settings, where gradients are rarely non-zero \cite{Sas}. To solve the problem caused by FedAdam, a simple adaptive method called FedYogi was proposed. 

For the FedAdagrad optimizer, $v_{\text{Adagrad}}^{t}$ is given by
\begin{equation}
    v_{\text{Adagrad}}^{t} = v_{\text{Adagrad}}^{t-1} + \|\hat{\bm{w}}_t\|^2.
\end{equation}
For the FedAdam optimizer, $v_{\text{FedAdam}}^{t}$ is written as
\begin{equation}
    v_{\text{FedAdam}}^{t} = \beta_2v_{\text{FedAdam}}^{t-1} + (1- \beta_2)\|\hat{\bm{w}}\|_t^2.
\end{equation}
For the FedYogi optimizer, $v_{\text{FedYogi}}^{t}$ is obtained as
\begin{equation}
    v_{\text{FedYogi}}^{t} = v_{\text{FedYogi}}^{t-1} - (1-\beta_2)\|\hat{\bm{w}}\|_t^2\text{sign}(v_{\text{FedYogi}}^{t-1} - \|\hat{\bm{w}}\|_t^2).
\end{equation}
These three adaptive optimizers have the same learning steps, including initialization, sampling subsets, and computing estimates, and they have been proved to achieve higher accuracy than FedAvg. However, these three adaptive optimizers have some differences. Unlike the FedAdagrad optimizer mainly well-suited for dealing with sparse data, both the FedAdam and FedYogi optimizers are suitable for sparse and non-sparse data. In addition, the FedAdam optimizer can rapidly increase the learning rate, while the FedYogi optimizer increases it in a controlled fashion, for which detailed proof is provided in \cite{Zaheer}.

\paragraph{FedProx} FL has two key challenges that need to be addressed: 1) Significant variability in terms of the system characteristics of each device, which is referred to as system heterogeneity. For example, the storage, computation, and communication capabilities of each device in federated networks may be different due to variability in hardware (CPU, memory), network connectivity (3G, 4G, 5G, THz, WiFi), and power (battery level) \cite{Barakat,Cavalcanti}. 2) Non-identically distributed data across networks, which is referred to as statistical heterogeneity \cite{Bonawitz}. Fortunately, to address these issues, FedProx has been proposed in \cite{Tianli} based on a federated optimization algorithm that can deal with heterogeneity both theoretically and empirically. Similar to FedAvg, FedProx randomly selects a subset of devices from $K$ devices and averages them to form a global model. Different from FedAvg, in the local model updating of FedProx, a proximal term $\frac{\mu}{2}\|\bm{w}_g^t - \bm{w}_k^t\|^2$ is added to effectively limit the impact of variable local updates, where the local training objective of the $k$th device is determined by
\begin{equation}
    \min_{\bm{w}_k^t}f_k(\bm{w}_G^t, \bm{w}_k^t) = \frac{1}{n_k}\sum_{i\in D_k}l_{i}(\bm{w}_k^t) + \frac{\mu}{2}\|\bm{w}_G^t - \bm{w}_k^t\|^2.
\end{equation}
In (21), $\mu$ is the regularization parameter. There are two advantages of the proximal term purpose: 1) it addresses the issue of statistical heterogeneity by restricting the local model updating to be closer to the global model without any need to set the number of local epochs manually, and 2) it allows for the aggregation of a large number of local models resulting from system heterogeneity. 

\paragraph{FOLB} Following the idea of the proximal term in FedProx, FOLB was proposed in \cite{Nguyen}. FOLB aims at maximizing the training loss reduction in each iteration. The main difference between FedProx and FOLB is the way they update the global model. Also, FOLB can achieve higher model accuracy, training stability, and higher convergence speed over FedAvg and FedProx. Different from FedAvg and FedProx, in FOLB, in the $t$th round, the server selects two multisets of devices $S_1^t$ and $S_2^t$ with $K$ randomly selected devices in each set, and transmits $\bm{w}_G^t$ to the $k$th device from set $S_1^t$ and the $\hat{k}$th device from set $S_2^t$. For the $k$th device in $S_1^t$, it computes the local update $\bm{w}_{k}^{t+1}$, and delivers both $\bm{w}_{k}^{t+1}$ and loss $f_{k}(\bm{w}_k^{t})$ to the server. For the $\hat{k}$th device in $S_2^t$, it only calculates and transmits its loss $f_{\hat{k}}(\bm{w}_k^{t})$ to the server. Then, rather than performing simple averaging, the server calculates the global model via
\begin{equation}
    \bm{w}_G^{t+1} = \bm{w}_G^{t} + \sum_{k\in S_1^t}\frac{<f_{k}(\bm{w}_G^t), \nabla_{S_{1}} \mathcal{L}(\bm{w}_G^t)>}{\sum\limits_{\hat{k}\in S_{2}^{t}}<f_{\hat{k}}(\bm{w}_{G}^t), \nabla_{S_{2}} \mathcal{L}(\bm{w}_G^t)>}\triangle \bm{w}_{k}^{t+1},
\end{equation}
where
\begin{equation}
    \nabla_{S_{i}} \mathcal{L}(\bm{w}_G^t) = \frac{1}{K}\sum_{k\in S_i^t}f_{k}(\bm{w}_G^t),~ i\in\{1, 2\}
\end{equation}
is the gradient of the global loss $\mathcal{L}(\bm{w}_G^t)$ obtained from the local loss of the devices in set $S_i^t$, and $\triangle \bm{w}_{k}^{t+1}$ is calculated as
\begin{equation}
    \triangle \bm{w}_{k}^{t+1} = \bm{w}_{k}^{t+1} - \bm{w}_{G}^{t}.
\end{equation}
In (22), the intuition is that the local update of the $k$th device is weighted by a measure of how correlated its gradient $f_{k}(\bm{w}_G^t)$ is with the global gradient $\nabla \mathcal{L}(\bm{w}_G^t)$. This correlation is assessed relative to $\nabla_{S_{1}} \mathcal{L}(\bm{w}_G^t)$, which is an unbiased estimate of $\nabla \mathcal{L}(\bm{w}_G^t)$ using gradient information obtained from $S_1^t$. The weights are normalized relative to a second unbiased estimate of the total correlation among $K$ devices, obtained from $S_2^t$. Using the inner product term $<f_{k}(\bm{w}_G^t), \nabla_{S_{1}} \mathcal{L}(\bm{w}_G^t)>$ can help connect the local model to the global model, and decrease the distance between them \cite{Xiang_Wu}. According to \cite{Nguyen}, the comparison of the testing accuracy of FOLB, FedProx, and FedAvg on different datasets is shown in Fig. 6. It is observed that FOLB is able to achieve a higher level of testing accuracy than the other two algorithms.

\begin{figure}[!h]
    \centering
    \includegraphics[width=3.5 in]{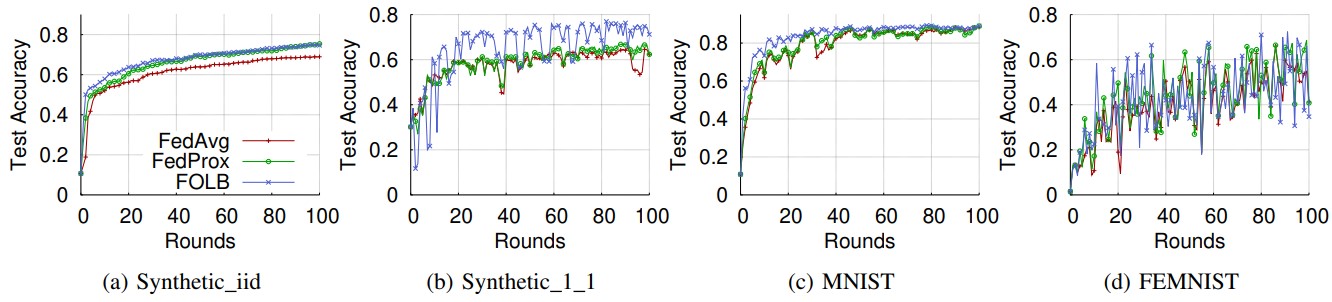}
    \caption{Comparison of the testing accuracy of FOLB, FedProx, and FedAvg on different datasets from \cite{Nguyen}.}
    \label{basic_modules}
\end{figure}

\subsection{Communication Efficiency}
Because of the asymmetric property of the internet connection and the large number of servers and devices, one of the major challenges in FL is the high communication overhead. There are mainly two ways to improve communication efficiency, which are decreasing and compressing the size of the learning model, respectively.

In \cite{J_Konecny}, the update of the local model of the $k$th device is calculated as (5), and the $k$th device transmits $ \bm{w}_{k}^{t+1}\in\mathbb{R}^{d_1\times d_2}$ to the server, where $\bm{w}_{k}^{t+1}$ has $d_1$ rows and $d_2$ columns. The authors in \cite{J_Konecny} proposed two ways to reduce the cost of transmitting local models to the server, which are structured updates and sketched updates. 

For structured updates, $\bm{w}_{k}^{t+1}$ is limited to having a pre-specified structure, either a low-rank or random-mask structure, where these two approaches are independent of each other. In the low-rank structure, $\bm{w}_{k}^{t+1}$ is the product of two matrices, written as
\begin{equation}
    \bm{w}_{k}^{t+1} = \textbf{A}_k^{t+1}\textbf{B}_k^{t+1},
\end{equation}
where $\textbf{A}_k^{t+1}\in\mathbb{R}^{d_1\times d}$, and $\textbf{B}_k^{t+1}\in\mathbb{R}^{d\times d_2}$. In (25), $\textbf{A}_k^{t+1}$ is a randomly generated matrix during the local updating, and only $\textbf{B}_k^{t+1}$ can be optimized. Thus, $\textbf{A}_k^{t+1}$ is presented in the form of a random seed and directly saved in the server, where the random seed means that it can generate the same random number in each time slot, and the $k$th device only needs to send the trained $\textbf{B}_k^{t+1}$ to the server, which can save a factor of $d_1/d$ during the uplink transmission. While in the random-mask structure, $\bm{w}_{k}^{t+1}$ is restricted to be a sparse matrix, where the sparse matrix consists of mostly zero values. Thus, the $k$th device only needs to send the non-zero values of $\bm{w}_{k}^{t+1}$. 

The sketched update is used to reduce communication costs. In this case, first the updated local model is computed without any constraints, and then the update is approximated or encoded in a compressed form before transmission to the server. Two approaches are used for the sketched update, which are subsampling and probabilistic quantizations. In subsampling quantization, rather than transmitting $ \bm{w}_{k}^{t+1}$, the $k$th device only needs to transmit $\hat{\bm{w}}_{k}^{t+1}$ to the server, which is created from a random subset of values from $\bm{w}_{k}^{t+1}$. The server then averages the subsampled update and calculates the global model as (5). In probabilistic quantization, the way of compressing the local update is through quantizing each scalar of the local weights into one bit. Let $\bm{w} = (w_1,...,w_{d_1\times d_2}) = \text{vec}(\bm{w}_t^k)$, and let $w_{\text{max}} = \max(w_i)$ and $w_{\text{min}} = \min(w_i)~(i = 1,2,...,d_1\times d_2)$, the compressed update of $\bm{w}$, denoted by $\tilde{w}$, is generated using
\begin{equation}
    \tilde{w}_i = \begin{cases}  {w_{\text{max}}}, &\text{with probability } \frac{w_i - w_{\text{min}}}{w_{\text{max}} - w_{\text{min}}}; \\
    w_{\text{min}}, & \text{with probability } \frac{w_{\text{max}} - w_i}{w_{\text{max}} - w_{\text{min}}}.
    \end{cases}
\end{equation}
According to (26), $\tilde{w}$ is an unbiased estimator of $\bm{w}$, and this method provides $32$ times compression explained in \cite{Suresh}.

A multi-objective evolutionary algorithm was designed in \cite{Hangyu} to minimize communication costs and improve learning accuracy simultaneously. To achieve these two objectives, modified sparse evolutionary training (SET) was proposed, which can decrease the number of connections between two layers in deep neural networks (DNNs) \cite{Mocanu}. In the modified SET algorithm, the connection probability between two layers is computed as
\begin{equation}
    p(\bar{\bm{w}}_{ij}^{k}) = \frac{\varepsilon(n_{k} + n_{k-1})}{n_{k}n_{k-1}},
\end{equation}
and the total number of connections between two layers is calculated as
\begin{equation}
    n_{\bar{\bm{w}}_{ij}^{k}} = n_{k}n_{k-1}p(\bar{\bm{w}}_{ij}^{k}),
\end{equation}
where $n_{k-1}$ and $n_{k}$ are the number of neurons of the $(k-1)$th and the $k$th layer, respectively, $\bar{\bm{w}}_{ij}^{k}$ is the sparse weight matrix between two layers, and $\varepsilon\in(0,1)$ is the parameter of SET to determine the connection sparsity. By setting $\varepsilon$, a fraction of the weights with small updates will be removed in each training epoch. In this way, the number of weights of the learning model is decreased, and thus the size of the learning model is reduced. The local and global updates still follow (6) and (7), respectively. 

However, if the size of the FL model is large, decreasing or compressing the FL model via the methods in \cite{J_Konecny} and \cite{Hangyu} still has low communication efficiency. To address this issue, quantization-based SGD has been widely adopted in FL, where the number of quantization bits and also the quantization function are optimized to minimize the total number of bits to be transmitted. Recently, sign-SGD FL was proposed in \cite{signFL} to guarantee high robustness and communication efficiency. Rather than delivering gradient $\textbf{g}_k^t$ calculated by (5), each device quantizes the gradient with a stochastic 1-bit compressor $q(.)\in\{-1, 1\}$ and sends $q(\textbf{g}_k^t)$ to the central server. Then, the central server calculates 
\begin{equation}\label{eq: sign grad}
    \tilde{\textbf{g}}^t = \text{sign}(\frac{1}{K}\sum_{k=1}^{K}q(\textbf{g}_k^t)),
\end{equation}
and delivers $\tilde{\textbf{g}^t}$ to the devices. The local model of each device is updated as
\begin{equation}
    \bm{w}_k^{t+1} = \bm{w}_k^t - \eta \tilde{\textbf{g}}^t.
\end{equation}
In sign-FL, the size of the quantized gradient for model aggregation is just 1 bit, which significantly increases communication efficiency. However, it may lead to low learning accuracy and convergence rate. This is because fewer model weights are delivered to the central server for model aggregation, compared with the compression methods proposed in \cite{J_Konecny} and \cite{Hangyu}.

Unlike compression methods in \cite{J_Konecny}, \cite{Hangyu}, and \cite{signFL}, federated distillation (FD) was designed in \cite{Eunjeong}. FD leverages ensemble distillation techniques and exchanges model outputs between the central server and participating devices. Therefore, in FD,  each device only exchanges the output of the local model, and the communication payload size is not determined by the model size but by the output dimension, resulting in advantageous communication properties and achieving orders of magnitude reduction of the communication overhead compared with FedAvg, especially when large models are trained \cite{Hinton}. In FD, each device only exchanges the output of the model, the dimension of which is much smaller than that of the local model. In FD, each device treats itself as a student, and treats the averaged model output from all other devices as its teacher’s output. The model output of each device is a set of logit values normalized through a softmax function, and is denoted as a logit vector whose size is determined by the number of labels of all data samples. The teacher-student output difference is measured periodically by cross entropy and becomes the loss regularizer of the student, namely, the distillation regularizer.

To guarantee communication efficiency, each device stores mean logit vectors, and periodically uploads these local-average logit vectors to a server. For each label, the uploaded local-average logit vectors from all devices are averaged, resulting in a global-average logit vector per label, which will be further downloaded to each device. When each device computes the distillation regularizer, its teacher's output is selected as the global-average logit vector associated with the same label as the current training sample's label.

In the $t$th time slot, the global-average logit vector $\hat{\textbf{F}}_{k,l}^{t}$ is calculated as
\begin{equation}
    \hat{\textbf{F}}_{k,l}^{t} = \frac{\sum_{i=1,i\neq k}^K\bar{\textbf{F}}_{k,l}^{t}}{K-1},
\end{equation}
where $\bar{\textbf{F}}_{k,l}^{t}$ is the local-average logit vector of the $l$th label of the $k$th device, and is updated as
\begin{equation}
    \bar{\textbf{F}}_{k,l}^{t} = \textbf{F}_{k,l}^{t} / N_l.
\end{equation}
In (32), $N_l$ is the number of samples, whose learning output is the $l$th label, and ${\textbf{F}}_{k,l}^{t}$ is a logit vector of the $l$th label calculated as
\begin{equation}
    {\textbf{F}}_{k,l}^{t} = \sum_{x\in \mathcal{S}_k} F_l(\bm{w}_k, x),
\end{equation}
where $F_l(\bm{w}_k, x)$ is the logit vector of the $l$th label given input $x$ and local model weights $\bm{w}_k$. In (33), $\bm{w}_k$ is still updated using (6), $\mathcal{S}_k$ is the set containing all data samples of the $k$th device, when the number of data samples of the $k$th device is larger than $N_l$.
In the server, the local-average logit vector of the $l$th label $\bar{\textbf{F}}_{l}^{t}$ of all devices is updated using
\begin{equation}
    \bar{\textbf{F}}_{l}^{t} =  \sum_{i=1}^{K}\bar{\textbf{F}}_{i,l}^{t}.
\end{equation}
In the $(t+1)$th time slot, $\hat{\textbf{F}}_{k,l}^{t+1}$ is updated as
\begin{equation}
    \hat{\textbf{F}}_{k,l}^{t+1} = \frac{\bar{\textbf{F}}_{l}^{t} - \bar{\textbf{F}}_{k,l}^{t}}{K-1}.
\end{equation}
Then, $\hat{\textbf{F}}_{k,l}^{t+1}$ is transmitted to the $k$th device. As only the logit vector is sent, the transmission size is much smaller than the learning model, allowing on-device ML to adopt large-sized local models.

\subsection{Fairness}
Most of the current FL works assume all devices contribute equally to the global model in each communication round, rather than prioritizing them based on their contributions. However, in practice, not all devices contribute equally due to various reasons, such as the different quality and quantity of the data owned by each device. Therefore, the local model from some devices may result in better global model updates, whereas those of others may impair the performance of the global model. To address this issue, fairness needs to be considered in the FL \cite{fairness}.

One possible learning scenario for FL in large-scale applications is that it is trained based on data originating from a large number of devices in large-scale applications, and the FL model may become biased towards certain devices. To address this issue, agnostic FL (AFL) was proposed in \cite{Mohri}, where the global model was optimized for any target distribution formed by a mixture of device distributions, with the aim to minimize the loss function of all devices and guarantee fairness among devices. Different from FedAvg, the training objective of AFL is given by
\begin{equation}
    \min_{\bm{w}_G}\max_{\bm{\lambda}_{G}}~\mathcal{L}(\bm{w}_G,\bm{\lambda}_{G}) = \sum\limits_{k=1}^{K}\bm{\lambda}_{k}f_{k}(\bm{w}_k),
\end{equation}
where $\bm{\lambda}_k$ is the mixture weight of the $k$th device. To solve the problem, each device needs to optimize $\bm{w}_k$ and $\bm{\lambda}_k$ simultaneously. Using SGD, in the $(t+1)$th time slot, $\bm{w}_k^{t}$ and $\bm{\lambda}_k^{t}$ are calculated as
\begin{equation}
    \bm{w}_k^{t+1} = \bm{w}_k^{t} - \gamma_{\bm{w}_k}\delta_{\bm{w}_k}f_{k}(\bm{w}_k^{t}),
\end{equation}
and
\begin{equation}
    \bm{\lambda}_k^{t+1} = \bm{\lambda}_k^{t} + \gamma_{\bm{\lambda}_k}\delta_{\bm{\lambda}_k}f_{k}(\bm{w}_k^{t}),
\end{equation}
respectively, where $\delta_{\bm{w}_k}f_{k}(\bm{w}_k^{t})$ and $\delta_{\bm{\lambda}_k}f_{k}(\bm{w}_k^{t})$ are the unbiased estimates of the gradient, and $\gamma_{\bm{w}_k}$ and $\gamma_{\bm{\lambda}_k}$ are the respective learning rates. Then, the global model update can still be written as in (7). Using AFL,  accuracy and fairness in applications with unknown mixture of device distributions can be guaranteed, thus, it can be used in large-scale networks.

A $q$-Fair FL ($q$-FFL) was proposed in \cite{Tianli1} to encourage a fairer accuracy distribution across all devices, where $q$ $(q>0)$ controls the tradeoff between fairness and accuracy. If $q=0$, fairness in the FL is not encouraged. A larger $q$ means imposing more uniformity in the training accuracy distribution and potentially inducing fairness. The objective of $q$-FFL is given by
\begin{equation}
    \min_{\bm{w}_G}~\mathcal{L}_{q}(\bm{w}_G) = \sum\limits_{k=1}^{K}\frac{p_k}{q+1}f_k^{(q+1)}(\bm{w}_k).
\end{equation}
To solve $q$-FFL in (39), it is important to first determine how to set $q$. In practice, $q$ can be tuned based on the desired amount of fairness. Also, it is possible that a family of objectives with different $q$ values has to be trained so that the algorithm can explore the trade-off between accuracy and fairness for different applications. However, one concern with addressing such a family of objectives is that it requires step-size tuning for every value of $q$ and can cause the search space of $q$ to explode. To solve the problem, the authors in \cite{Tianli1} considered estimating the local Lipschitz constant, which could prevent the function value from skipping the optimal value, and dynamically adjust the step-size of the gradient-based optimization method for the $q$-FFL objective, avoiding manual tuning for each $q$ \cite{Xu_zhou}.

The local model updates of the $k$th device are calculated as in (6). The global model updates are given by the sum of the first-order derivatives of $f_k^{q}(\bm{w}_k)$ divided by the sum of second-order derivatives of $f_k^{q}(\bm{w}_k)$. Thus, the $k$th device computes
\begin{equation}
    \triangle \bm{w}_k^t = L(\bm{w}_G^t - \bm{w}_k^{t+1}),
\end{equation}
\begin{equation}
    \triangle_k^t = f_k^q(\bm{w}_k^t)\triangle \bm{w}_k^t,
\end{equation}
\begin{equation}
    h_k^t = qf_k^{(q-1)}(\bm{w}_k^t)\|\triangle \bm{w}_k^t\|_2 + Lf_k^q(\bm{w}_k^t),
\end{equation}
where $L$ is the Lipschitz constant, $\triangle_k^t$ is the first-order derivative of $f_k^{q}(\bm{w}_k)$, and $h_k^t$ is the second-order derivative of $f_k^{q}(\bm{w}_k)$. Then, the update of the global model in $q$-FFL is given by
\begin{equation}
    \bm{w}_G^{t+1} = \bm{w}_G^{t} - \frac{\sum_{k=1}^{K}\triangle_{k}^t}{\sum_{k=1}^{K}h_{k}^t},
\end{equation}
where $\triangle_{k}^t$ and $h_{k}^t$ are given in (41) and (42), respectively.

Unlike using the same version of the global model in \cite{Mohri} and \cite{Tianli1}, collaborative Fair FL (CFFL) was proposed in \cite{Lingjuan_Lyu} to utilize reputation to update the local model of each device to converge to different models, which can achieve collaborative fairness by adjusting the performance of the models allocated to each
participant based on their contributions.
The reputation is applied to quantify the contribution of each device, and the reputation of the $k$th device is represented as
\begin{equation}
c_k = \sinh(\alpha\ast \frac{{vacc}_k}{\sum_{i=1}^{K}{vacc}_i}),
\end{equation}
where ${vacc}_k$ is given as
\begin{equation}
    {vacc}_k = \bm{w}_k + \eta \textbf{g}_k.
\end{equation}
In (44), $\sinh(\alpha)$ serves as a punishment function, and $\alpha$ denotes the punishment factor, that is used to distinguish the reputations of different devices based on how informative their uploaded gradients are. The larger the variation of the gradient of the local model of the $k$th device, the higher ${vacc}_k$, and the higher contribution of the $k$th device.  The local and global models in the CFFL are updated using (6) and (7), respectively. However, the global model allocated to the $k$th device needs to be calculated according to its contribution using
\begin{equation}
    \bm{w}_k^{t+1} =  \frac{c_k}{\sum\limits_{i=1}^{K}c_i}\bm{w}_G^{t+1},
\end{equation}
where $c_k$ is given in (44).

Due to the fact that CFFL enables devices to converge to different final models, the most contributive device receives the most accurate model. CFFL not only can achieve comparable accuracy to FedAvg, but also can guarantee higher fairness than  FedAvg. 

\subsection{Bayesian Learning}
FedAvg requires access to locally stored data for learning. However, it is possible that the local model cannot be trained by the local data. Such situations may be caused by catastrophic data loss or by regulations such as the general data protection regulation \cite{GDPR}, which places severe restrictions on the storage and access of personal data. Thus, the transmitted local model cannot be updated by the local data in the current time slot. To solve this problem, Bayesian machinery can be deployed to estimate the local model weights via probabilistic neural matching based on the pre-trained local models. Then, the local models estimated by the Bayesian machinery and the updated local models trained by local data are delivered to the server for model aggregation.

The authors in \cite{Yurochkin} proposed a probabilistic federated neural matching (PFNM) algorithm, which used a Beta Bernoulli Process (BBP) \cite{romain} to model the multi-layer perceptron (MLP) weights. Through the permutation invariance of a fully-connected neural network, the proposed PFNM algorithm first matches the weights of the transmitted local model from each device to the weights of the global model. Then, it aggregates these local models by maximizing the posterior estimation of the global weights. As a result, the PFNM algorithm can achieve higher accuracy and communication efficiency than FedAvg. However, the PFNM algorithm can only be effective for simple architectures of the neural network, such as fully-connected feedforward neural networks. 

To deal with this issue, Federated Matched Averaging (FedMA) was proposed in \cite{Hongyi_Wang}, which constructed the shared global model in a layer-wise manner by matching and averaging hidden layers, including channels for convolutional neural networks (CNNs), hidden units for recurrent neural networks (RNNs), and weights for fully connected layers. The training objective of the FedMA algorithm is represented as
\begin{align}
    \min_{\pi_{li}^{k}}\sum_{i=1}^{L}&\sum_{k,l}\min_{\bm{w}_{G,i}}\pi_{li}^{k}c(\bm{w}_{k,l},\bm{w}_{G,i}),\\
    \text{s.t.} &\sum_i\pi_{li}^{k} = 1,\nonumber \\
    &\sum_l\pi_{li}^{k}=1, \nonumber
\end{align}
where $L$ is the number of hidden layers, $\bm{w}_{k,l}$ denotes the weights of the $l$th layer learned based on the dataset of the $k$th device, $\bm{w}_{G,i}$ denotes the weights of the $i$th layer of the global model, $c(.,.)$ is an appropriate similarity function between a pair of weights, and $\pi_{li}^{k}$ is the permutation, which determines the contribution of the weights of the $l$th layer of the local model of the $k$th device to the neurons of the $i$th layer of the global model. From (47), we observe that the objective is to minimize the weight distance between the local model and the global model, and $c(.,.)$ is the squared Euclidean distance. Through optimizing the permutation $\pi_{li}^{k}$, the total weight distance between the layers of the local model of all $K$ devices and the layer of the global model can be minimized.

\begin{figure}[!h]
    \centering
    \includegraphics[width=3.5 in]{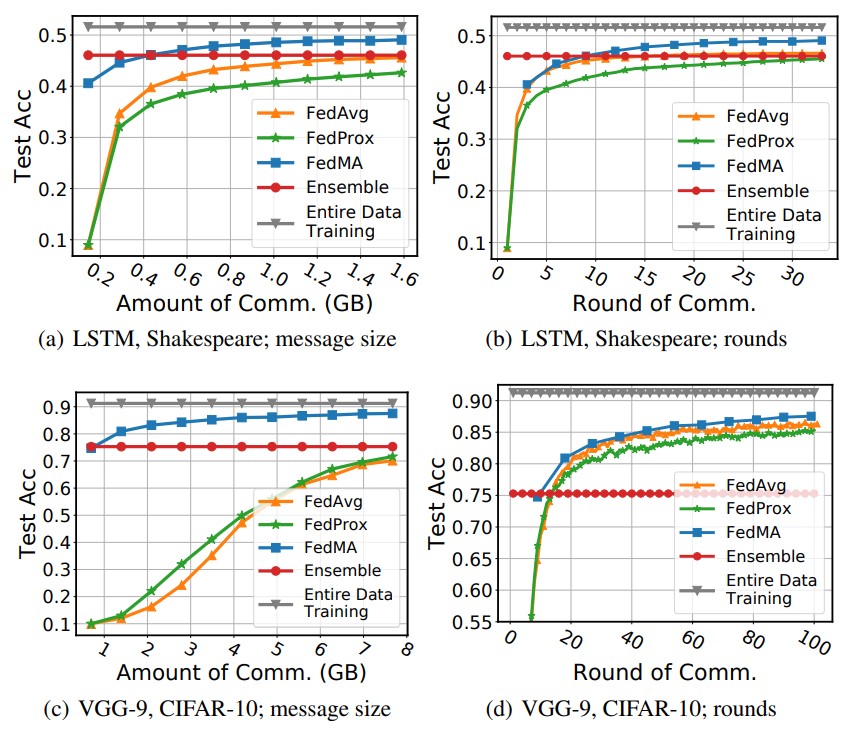}
    \caption{Comparison of the convergence rate and testing accuracy of FedMA, FedProx, and FedAvg on different datasets from \cite{Hongyi_Wang}.}
    \label{basic_modules}
\end{figure}

\begin{figure*}[!h]
    \centering
    \includegraphics[width=6.5 in]{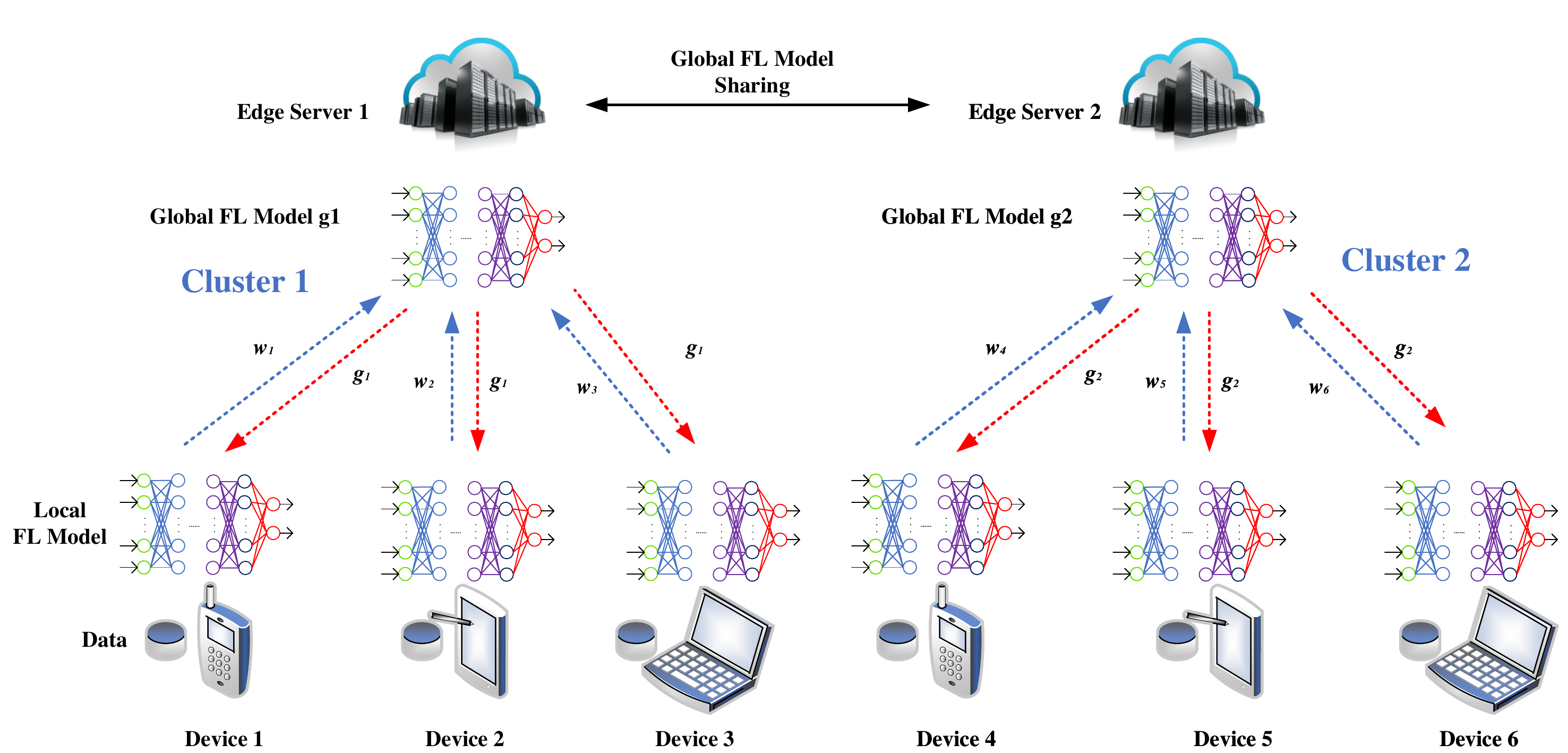}
    \caption{Multi-client local model aggregation.}
    \label{basic_modules}
\end{figure*}

In the FedMA algorithm, first, the server gathers only the weights of the first layers from devices and performs one-layer matching to obtain the weights of the first layer of the global model. Then, the server broadcasts these weights to the devices, and each device updates the first layer of its local model to train all consecutive layers with its own dataset, keeping the matched layer frozen. The procedure continues until all layers have finished matching. Thus, FedMA requires the number of communication rounds to be equal to the number of layers in the neural network. We assume that there are $N$ layers in the neural network. Through computing a posterior estimate (MAP) of the Bayesian non-parametric model based on the BBP, for the $n$th layer, if $n<N$, the global model in FedMA is updated as
\begin{equation}
    \bm{w}_{G,n} = \frac{1}{K}\sum\limits_{k=1}^{K}\bm{w}_{k,n}\{\prod\}_{k=1}^{K},
\end{equation}
where
\begin{equation}
    \{\prod\}_{k=1}^{K} = \text{BBP-MAP}(\{\bm{w}_{k,n}\}_{k=1}^{K}).
\end{equation}
In (49), $\{\prod\}_{k=1}^{K}$ is a permutation matrix. The permutation matrix is an orthogonal matrix and is usually used to match the layers of neural networks in the weight space. If $n=N$, the global model is updated as
\begin{equation}
    \bm{w}_{G,N} = \sum\limits_{k=1}^{K}p_{k}\bm{w}_{k,N},
\end{equation}
where $p_{k}$ is calculated as (4). For the $k$th device, the local model is updated as
\begin{equation}
    \bm{w}_{k,n+1} = \{\prod\}_k\bm{w}_{k,n+1}.
\end{equation}

According to \cite{Hongyi_Wang}, the comparison of the convergence rate and testing accuracy of FedMA, FedProx, and FedAvg on different datasets is shown in Fig. 7. Compared to FedProx and FedAvg, FedMA not only improves communication efficiency, as the number of communication rounds is equal to the number of layers of the neural network, but also guarantees learning accuracy.

\subsection{Clustering}
For the aforementioned FL techniques, the central server updates only a single global model. In contrast, for clustered FL, the central server updates multiple global models, where the number of global models is equal to that of the clusters. Clustered FL partitions devices into different groups as in \cite{Ghosh}, where $K$ devices were partitioned into $M$ disjoint clusters. This method captures settings where different groups of devices have their own learning tasks. It is assumed that all devices do not have any knowledge of 
the other device's cluster identity. To minimize the loss function while estimating the cluster identities, an Iterative Federated Clustering Algorithm (IFCA) was proposed in \cite{Ghosh}. In the $t$th time slot, the central server transmits $M$ updated global models to these $M$ clusters. Then, a random subset of devices is selected to update their local models with their corresponding data samples and delivers the local models to the central server. As the central server does not know the cluster identities of the selected devices, it estimates the identity of the $k$th device using
\begin{equation}
    \hat{j} = \arg\min_{j\in\mathcal{M}}f_k(\bm{w}_{G,j}^t),
\end{equation}
where $\mathcal{M}$ is the set including all clusters, $f_k(\bm{w}_{G,j}^t)$ is the loss of the $k$th device in the $j$th cluster, and $\bm{w}_{G,j}^t$ is the global model of the $j$th cluster. From (52), we observe that the $k$th device belongs to the $\hat{j}$th cluster that achieves the minimum loss. Given the estimated clusters, the global model of each cluster is updated using (7), and the local model of each device is still updated using (6). 

However, in \cite{Ghosh}, all selected devices need to communicate with the central server, thus, there is a large overhead when a large number of devices transmit. To address this issue, a hierarchical clustering (HC) algorithm for local model updating was proposed in \cite{Briggs}, where the devices were iteratively merged into clusters with high simmilarity by calculating L1 (Manhattan), L2 (Euclidean), and cosine distance metrics. In the $t$th time slot, the $d$-norm distances between all clusters are calculated to judge their similarity. The distance $\hat{d}_{i,j}$ between the $i$th cluster and the $j$th cluster in the global model is calculated as
\begin{equation}
    \hat{d}_{i,j} = \|\bm{w}_{G,i}^{t} - \bm{w}_{G,j}^{t}\|_d.
\end{equation}
If $\hat{d}_{i,j}$ is smaller than the threshold $\hat{d}_{\text{th}}$, the $i$th and $j$th clusters can be merged together. This procedure continues until all clusters with similarity are merged into a single cluster. Then, these merged clusters select a portion of their devices to aggregate the local models in the server via (7), and the local models in the devices are still updated using (6). By selecting a portion of the devices in the merged clusters to aggregate the local models, the communication overhead is reduced. 

Although authors in \cite{Ghosh} and \cite{Briggs} considered clustered FL, both works adopted a single central server to capture the global models of all devices by aggregating their local models. In \cite{Ming_Xie}, a multi-center aggregation mechanism for multiple global models in clustered FL was proposed, where devices belong to a specific cluster, and the cluster updates its own global model with its corresponding updated local models, as shown in Fig. 8. Each device calculates the distance between its local model and the
global model of the server in each cluster. The device selects the cluster with the minimum
distance. The learning objective of multi-center clustered FL is to minimize the total weighted distance between the global model and the local models, and the multi-center weight distance-based loss (MD-Loss) is represented as
\begin{equation}
   \mathcal{L} = \frac{1}{K}\sum_{i=1}^{m}\sum_{k=1}^{K}r_{k,i}~\text{Dist}(\bm{w}_k,\bm{w}_{G,i}),
\end{equation}
where
\begin{equation}
    \text{Dist}(\bm{w}_k,\bm{w}_{G,i}) = \|\bm{w}_k - \bm{w}_{G,i}\|_2,
\end{equation}
$r_{k,i} = \{0, 1\}$ is the cluster assignment, $r_{k,i} = 1$ indicates that the $k$th device belongs to the $i$th cluster, vice versa. In (54), $K$ and $m$ are the number of devices and clusters, respectively. In (55), $\bm{w}_{G,i}$ is the global model of the $i$th cluster.  To solve (54), a federated stochastic expectation maximization (FeSEM) algorithm is deployed in the following three steps:

First, the cluster assignment $r_{k,i}$ is updated using
\begin{equation}
    r_{k,i} = \begin{cases}  {1}, &\text{if } i = \arg\min_j \text{Dist}(\bm{w}_k,\bm{w}_{G,j}); \\
    0, & \text{otherwise}.
    \end{cases}
\end{equation}
From (56), we observe that the $k$th device belongs to the $i$th cluster that can achieve the minimum weight distance.

Second, the global model of the $k$th cluster is aggregated as
\begin{equation}
    \bm{w}_{G,i} = \frac{1}{\sum\limits_{k=1}^{m}r_{k,i}}\sum\limits_{k=1}^{K}r_{k,i}\bm{w}_k.
\end{equation}

Finally, the local model of the $k$th device is still updated using (6). The multi-center aggregation mechanism can better capture the heterogeneity of data distributions across devices, and simultaneously facilitates the optimal matching between devices and servers.

\begin{figure*}[!h]
    \centering
    \includegraphics[width=5.5 in]{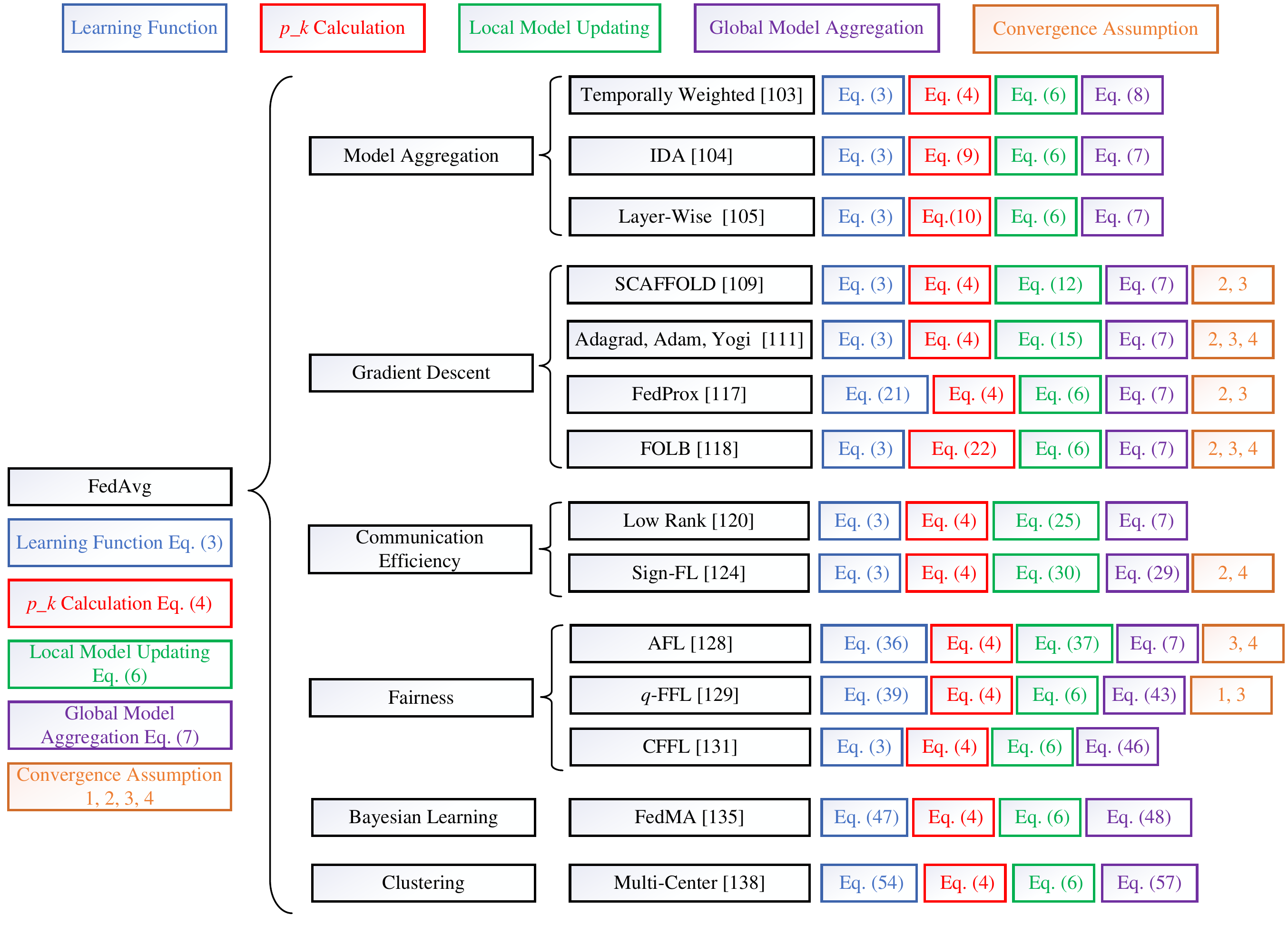}
    \caption{Evolution of FedAvg in FL methodologies.}
    \label{basic_modules}
\end{figure*}

FL with a single server or multiple servers still has high vulnerability when there exists a failure or an attack on the server. Therefore, unlike authors in \cite{Ghosh,Briggs,Ming_Xie} considered single or multiple servers for model aggregation, authors in \cite{Kavalionak} proposed a general fully-decentralized FL framework where the device was denoted as a ``server" that update its local model by aggregating models from other devices. The model updating follows (7). Furthermore, authors in \cite{Lalitha} derived convergence analysis of fully decentralized FL under the condition that devices update their local models by aggregating models from their one-hop neighbors.

\subsection{Convergence Analysis of FL Methodologies}
The derivation of convergence analysis of FL is usually based on the following fundamental assumptions.

$\textbf{Assumption 1.}$ (Lipschitz). Loss functions $f_1,...,f_N$ are all $\rho-$ lipschitz for any $\bm{w}$ and $\bm{w}^{'}$:
\begin{equation}
    \|f_{n}(\bm{w}) - f_{n}(\bm{w}^{'})\| \leq \rho\|\bm{w} - \bm{w}^{'}\|,
\end{equation}
where $\rho$ is a positive constant.

$\textbf{Assumption 2.}$ (Smoothness). Loss functions $f_1,...,f_N$ are all $L-$ smooth for any $\bm{w}$ and $\bm{w}^{'}$:
\begin{equation}
    \|\nabla f_{n}(\bm{w}) - \nabla f_{n}(\bm{w}^{'})\| \leq L\|\bm{w} - \bm{w}^{'}\|,
\end{equation}
where $L$ is a positive constant.

$\textbf{Assumption 3.}$ (Bounded Gradient and model). The second moments of stochastic gradients and weights are bounded for any $\bm{w}$, and they are guaranteed by the $l_2$-regularization \cite{Bartlett,Salimans}, which are denoted as
\begin{equation}
    \mathbb{E}\|\nabla f_{n}(\bm{w})\|^2\leq G,
\end{equation}
and
\begin{equation}
    \mathbb{E}\|\bm{w}\|^2 \leq A.
\end{equation}
In (60) and (61), both $G$ and $A$ are positive constants.

$\textbf{Assumption 4.}$ (Gradient divergence). The gradient divergence for any $n$ and $\bm{w}$ is denoted as
\begin{equation}
    \mathbb{E}\|\nabla f_{n}(\bm{w}) - \nabla f(\bm{w})\|^2\leq \hat{\sigma}^2.
\end{equation}
In (62), $\hat{\sigma} > 0$ is a constant.

For FL methodologies, researchers usually derive the number of communication rounds or the upper bound of  $\mathbb{E}\|\nabla f(\bm{w})\|^2$ in convergence analysis. We will give several examples as follows.

In SCAFFOLD, based on assumptions 2 and 3, the number of communication rounds to achieve convergence in non-convex FL settings is derived as
\begin{equation}
    R \! = \! \mathcal{O}\!\!\left(\!\frac{L\hat{\sigma}^2(f(\bm{w}^{0}) - f(\bm{w}^{*}))}{NS\epsilon} \!\!+\!\! \left(\frac{K}{S}\right)^{\frac{2}{3}}\!\!\frac{B(f(\bm{w}^{0}) - f(\bm{w}^{*}))}{\epsilon}\!\right)\!\!,
\end{equation}
where $N$ is the number of local updates, $K$ is the total number of devices, $S$ is the selected number of devices for model aggregation, and $\epsilon$ is the required learning accuracy to reach.

In FedProx, based on assumptions 1 and 3, after $T = \mathcal{O}(\frac{f(\bm{w}^{0}) - f(\bm{w}^{*})}{\rho G})$ iterations, it is obtained that
\begin{equation}
    \frac{1}{T}\sum_{t=0}^{T-1}\mathbb{E}[\|\nabla f(\bm{w}^{t})\|^2]\leq G.
\end{equation}

\textit{Remark:} Although most research works consider the same assumptions for convergence analysis, there is no uniform equation in convergence analysis for all FL algorithms. Therefore, the convergence rate for different FL settings should be derived according to specific situations in detailed FL works.

\subsection{Summary and Lessons Learned}
In this section, we have reviewed six main research areas of FL methodologies. We summarize the approaches along with references. From this review, we gather the following lessons learned:
\begin{itemize}
    \item The differences between FedAvg and other FL algorithms mentioned are presented in Fig. 9. Based on Fig. 9, FedAvg includes four components, which are the learning function, $p_k$ calculation, local model updating, and global model aggregation, corresponding to equations (3), (4), (6), and (7), respectively. Through jointly or separately optimizing these four equations, the learning efficiency, accuracy, and fairness can be improved. In addition, theoretical convergence analysis is usually derived based on four fundamental assumptions.
    \item For the research works we have discussed in this section, the authors assumed that the proposed FL algorithms are implemented in a synchronous mode. The heterogeneity among devices, such as computation capabilities, is usually not considered. Meanwhile, authors assume that all devices successfully transmit and receive local models and updated global models. In real-world applications, the approach may not be feasible for devices with weak processing power or unstable network connection, which can lead to straggler effect.
    \item The simulations in the research works we have discussed are usually performed on MNIST, FashionMNIST, and CIFAR. The data in these datasets are usually labeled and the data distribution of them is clear. Therefore, whether the proposed FL algorithms are effective in unlabeled datasets or other datasets with unknown data distribution still needs to be investigated.
    \item For the studies that we have discussed in this section, local models are trained with non-poisoned data. Data poisoning refers to the malicious device who either actively or passively uses some poisoned data for model training, and model poisoning means that an attacker tampers the weights of the local model, which further affects the parameters of the global model. Although FL by itself has a certain level of resilience against attacks, the frequent communications between server and devices may spread the risk over the
    networks and thus breaks down the overall learning systems. Therefore, how to design a FL algorithm that is robust to poisoned data and design resilient networks for FL to avoid spreading attacks needs to be investigated.
\end{itemize}

\section{FL in Wireless Networks}
Recently, there is a growing interest in optimizing wireless networks with data-driven ML-based methods. In this section, based on the fundamental FL algorithms in Section III, we present the research areas of designing FL for wireless networks.  Considering FL in wireless networks, all weights of local or global models are delivered via wireless links instead of wired ones. Thus, the model aggregation, learning accuracy, and learning efficiency of FL can be influenced by wireless factors, such as the set of devices that participate in FL, computational capacity, transmission power and wireless channel, and spectrum resource allocation. The impact of these factors on FL performance is introduced as follows:
\begin{itemize}
    \item With the increasing number of devices participating in model aggregation, the generalization of the global model increases. However, more devices lead to high interference and a low transmission rate.
    \item Computational capacity of each device affects the learning latency. High computational capability leads to high learning efficiency and low learning latency. 
    \item Transmission power and wireless channel determine transmission rate and reliability. Low transmission power and dynamic wireless channel result in low transmission rates and high transmission errors.
    \item Spectrum resource allocated to each device affects transmission rate. When more spectrum is allocated to the device, its transmission rate increases.
\end{itemize}
To exchange a large number of model
weights over time-varying channels, there are two types of solutions, which are ``digital'' and ``analog'' approaches that convert all global or local models into bits and modulated signals, respectively. In this section, we introduce these key research areas of FL over wireless networks. For the digital approach, we consider model aggregation, communication, energy, and computation efficiency optimization, resource
allocation, and asynchronous FL. For the analog approach, we consider over-the-air computation.

\subsection{Digital Approach of FL over Wireless Networks}
The digital approach needs to guarantee a high transmission rate, low transmission error, and high communication, energy, and computation efficiency of FL over wireless networks. In this subsection, we mainly introduce 1) Model Aggregation, 2) Communication, Energy, and Computation Efficiency Optimization, 3) Resource Allocation, and 4) Asynchronous FL.

\subsubsection{Model Aggregation} The model uploading and downloading of FL can be affected by dynamic wireless channels. Due to the limited bandwidth of the wireless network, not all devices can transmit local models to the central server. Also, poor channel state results in high transmission error. To select optimal devices to upload the local models and minimize transmission error, device selection and packet transmission error minimization schemes need to be considered.

\textbf{Device Selection:} Three main device selection schemes have been studied, which are probabilistic updating, importance-based updating, and novel communication protocol-based updating.

\paragraph{Probabilistic Updating} A traditional probabilistic scheduling policy was developed in \cite{Howard_yang} to characterize the convergence performance of FL in wireless networks. In particular, the effectiveness of three different scheduling policies, i.e., random scheduling (RS), round robin (RR), and proportional fair (PF) were considered to select a portion of devices for local model aggregation under limited bandwidth constraints. In RS, the access point (AP) randomly selects $K$ associated devices in each time slot for local model updating, and each device is allocated with a sub-channel to deliver the local model. In RR, the AP arranges all devices into $G$ groups and assigns each group to access the radio channels and update their weights in each time slot. While in PF, the AP selects $K$ out of $ \hat{K}(K\leq\hat{K})$ associated devices in each time slot according to the following policy
\begin{equation}
    \textbf{m}^{*} = \mathop{\arg\max}_{\textbf{m}\in{\{1,2,...,\hat{K}\}}}\left\{\frac{\tilde{\rho}_{1,t}}{\bar{\rho}_{1,t}},...,\frac{\tilde{\rho}_{K,t}}{\bar{\rho}_{K,t}}\right\},
\end{equation}
where $\textbf{m} = \{m_1,...,m_K\}$ is a length-$K$ vector and $\textbf{m}^{*} = \{m_1^{*},...,m_K^{*}\}$ represents the indices of the selected $K$ devices, $\tilde{\rho}_{k,t}$ and $\bar{\rho}_{k,t}$ are the instantaneous and time average signal-to-noise ratio (SNR) of the $k$th device in the $t$th time slot, respectively. Thus, the device with a higher SNR is selected \cite{Choi}. The updating of the local and global models still follows (6) and (7), respectively. 

Based on the probabilistic analysis of the scheduling policies in \cite{Howard_yang}, it shows that PF outperforms RS and RR in terms of convergence rate under a high SNR threshold. This is because a high SNR threshold reduces the chance of successful transmission from an arbitrary device, while PF improves the convergence rate by selecting devices with better channel qualities in order to increase their transmission success probabilities. However, RR is preferable when the SNR threshold is low, this is because low SNR threshold results in a high success probability. 

\paragraph{Importance-based Updating} To exploit the importance of devices, a novel probabilistic scheduling framework was developed to apply unbiased update aggregation for the federated edge learning (FEEL) in \cite{Jinke}, where the importance of a local model update was measured by its gradient divergence. In the $t$th time slot, $K$ devices are selected for model aggregation according to a scheduling distribution $\mathcal{P}^{t} = (p_{1}^{t}, p_{2}^{t},..., p_K^{t})$, where $p_{k}^{t}$ is the probability that the $k$th device is selected, and can also indicate the level of importance that the $k$th device can contribute to the global model convergence. The local model is still updated using (6), and the global model is updated using
\begin{equation}
    \bm{w}_G^{t} = \frac{1}{n}\sum\limits_{k=1}^{K}\frac{n_k^{t}}{p_{k}^{t}}\bm{w}_{k}^{t},
\end{equation}
where $n_k$ is the number of data at the $k$th device, and $n = \sum_{i=1}^{K}n_i$. The selected local model $\bm{w}_{k}^{t}$ needs to be scaled by a coefficient $\frac{n_k^{t}}{n}$ at the edge server. This is because this coefficient well quantifies the unbalanced property in global data distribution and thus makes the global model unbiased. Based on the probabilistic scheduling framework, the importance indicator of each local model is defined as 
\begin{equation}
    I_{k}^{t} = \left\|\frac{n_k^{t}}{np_{k}^{t}}\bm{w}_{k}^{t} - \bm{w}_G^{t}\right\|_2.
\end{equation}
The model divergence reflects the deviation between the local model and the global model, and the smaller the model divergence, the more it can contribute to the global model convergence. In other words, the smaller the $p_{k}^{t}$, the less contribution of the $k$th device to the global model. The optimization problem is to achieve a trade-off between gradient divergence and latency, where $\rho\in[0,1]$ is defined as the weight coefficient that balances the gradient divergence and latency. Then, the objective function is represented as
\begin{align}
    \min_{(p_1^t,...,p_K^t)}&\sum\limits_{k=1}^{K}p_k^t\left[\rho I_k^t + (1-\rho)T_k^t\right],\\
    \text{s.t.}~&\sum\limits_{k=1}^{K}p_k^t = 1,\\
    &p_k^t \geq 0,
\end{align}
where $I_k^t$ is given by (67), and the optimal probability $p_k^{t*}$ is given by
\begin{equation}
    p_k^{t*} = \frac{n_k}{n}\|\bm{w}_k^{t}\|\sqrt{\frac{\rho}{(1-\rho)T_{k}^{t} + \lambda^{t}}},
\end{equation}
where $\lambda^{t}$ is the lagrangian multiplier that satisfies (69), and $T_{k}^{t}$ is the uplink transmission latency. From (71), the optimal scheduling decision is mainly determined by data unbalanced indicator $\frac{n_k}{n}$, the norm of local model $\|\bm{w}_k^{t}\|$, and the uplink latency $T_{k}^{t}$. Through the importance-aware device scheduling strategy, it can achieve less than half of the convergence time and up to $2\%$ higher final accuracy.

\paragraph{Novel Communication Protocol-based Updating} Apart from deploying probabilistic scheduling to select the optimal devices for model aggregation, a communication protocol designed for FL over wireless networks, called federated learning with client selection (FedCS), was proposed in \cite{Nishio}. First, the server requests $\lceil K\times C\rceil$ random devices to participate in the current training task, where $K$ is the total number of all devices, $C\in(0,1]$ is the fraction of devices that participating in training in each time slot, and $\lceil .\rceil$ is the ceiling function. The server selects as many devices as possible to transmit their local models for model aggregation within a certain deadline after receiving the resource information, such as wireless channel states, computational capacities, and size of data resources, from devices. The optimization problem is to maximize the number of devices for model aggregation as
\begin{align}
    \max~~&S,\\
    \text{s.t.}~&T_{\text{CS}} + T_{\text{Agg}} + T_{\text{up}} + T_{\text{down}} \leq T_{\text{th}},
\end{align}
where $S$ is the number of the selected devices for model aggregation, and $T_{\text{CS}}$, $T_{\text{Agg}}$, and $T_{\text{down}}$ are the time required for device selection, model aggregation, and downlink transmission, respectively. In (73), $T_{\text{up}}$ is the time required for local model updating and uplink transmission.

To solve the optimization problem in (72), a heuristic algorithm based on the greedy algorithm was proposed. Although the proposed FedCS algorithm can achieve a higher accuracy than that of FedAvg, the computation complexity is extremely high with a large number of devices. Also, the local and global updating still follow (6) and (7), respectively.

However, authors in \cite{Howard_yang,Jinke,Nishio} considered perfect wireless channels, and in the uplink transmission, there will be transmission errors caused by unstable wireless channels. To solve the problem, the impact of the packet transmission error on FL is considered.

\textbf{Packet Transmission Error:} The authors in \cite{mingzhe} considered packet transmission errors, which could affect the local model aggregation of FL. In \cite{mingzhe}, a closed-form solution for the convergence rate of FL was derived as a function of packet error rates. Based on this solution, the BS optimizes the resource allocation and the device optimizes its transmission power to decrease packet error rates. The optimization problem in \cite{mingzhe} is to minimize the training loss of FL over wireless networks. The expression of the model aggregation with packet transmission error is denoted as
\begin{equation}
    \bm{w}_g(\textbf{a}^{t},\textbf{P}^{t},\textbf{R}^{t}) = \frac{\sum_{k=1}^{K}\bar{N}_ka_k^{t}\bm{w}_k^{t}C(\bm{w}_k^{t})}{\sum_{k = 1}^{K}\bar{N}_ka_k^{t}C(\bm{w}_k^{t})}.
\end{equation}
In (74), $\sum_{k = 1}^{K}\bar{N}_ka_k^{t}C(\bm{w}_k^{t})$ is the total number of training data samples, which is determined by $a_k^{t}$ and $C(\bm{w}_k^{t})$, $N_k$ is the number of training data samples of the $k$th device,  $a_{k}^{t}\in\{0,1\}$ is the device association index of the $k$th device. If $a_{k}^{t} = 1$, the $k$th device is selected to update the local model to the BS, and vice versa. Also, in (74), $\textbf{P}^{t} = [P_1^{t},...,P_K^{t}]$ is the transmit power matrix, $C(\bm{w}_k^{t})$ is the packet transmission index of the $k$th device, which is presented as
\begin{equation}
    C(\bm{w}_k^{t}) = \begin{cases}  {1}, &\text{with probability } 1 - q_k(r_k^{t}, P_k^{t}), \\
    0, & \text{with probability } q_k(r_k^{t}, P_k^{t}),
    \end{cases}
\end{equation}
and $q_k(r_k^{t}, P_k^{t})$ is the packet error rate of the local model of the $k$th device. If $C(\bm{w}_k^{t})=0$, the local model of the $k$th device contains data error, and the BS will not use it to update the global model. In (75), $r_{k,n}^{t}$ is the resource block (RB) allocation index, and $r_{k,n}^{t} = 1$ means that the $n$th RB is allocated to the $k$th device in the $t$th time slot. Meanwhile, in (75), $q_k(r_k^{t}, P_k^{t})$ is expressed as
\begin{equation}
    q_k(r_k^{t}, P_k^{t}) = \sum_{n=1}^{R}r_{k,n}^{t}q_{k,n}^{t},
\end{equation}
where $q_{k,n}^{t}$ is the packet error rate over the $n$th RB with $m$ being a waterfall threshold and is defined as
\begin{equation}
    q_{k,n}^{t} = \textit{E}_{h_k}\left(1 - \text{exp}\left(-\frac{m(\bar{I}_n + BN_0)}{P_k^{t}h_k}\right)\right).
\end{equation}
In (77), $h_k$ is the channel gain between the BS and the $k$th device, $N_0$ is the noise power spectral density, and $\bar{I}_n$ is the interference caused by the other devices. The local model updating is still written as (7). Through optimizing the power and RB allocation via the Hungarian algorithm \cite{Jonker}, both the packet transmission error and the training loss can be minimized.

\subsubsection{Communication, Energy and Computation Efficiency Optimization}
For FL over wireless networks, one of the challenges is to maximize the communication, energy, and computation efficiency, which can be influenced by the bit rate, energy, and computation capability. In this subsection, the research on the optimization of communication, energy, and computation efficiency is introduced.

\paragraph{Communication Efficiency} Authors in \cite{Abad} introduced momentum gradient and sparse communication to increase the communication efficiency of FL over wireless networks. To optimize the transmission rate of each device, the optimal sub-carrier is allocated to each device. The modified momentum method is used to accelerate the performance of SGD. Then, based on the sparse communication, the global model is updated using
\begin{equation}
    \bm{w}_G^{t+1} = \sum_{k=1}^{K}p_k\text{sparse}(\bm{w}_k^t),
\end{equation}
where $p_k$ is calculated by (4), and sparse function $sparse()$ in (78) converts $\bm{w}_k^t$ to sparse form by squeezing out any zero elements \cite{Dally}. Using sparse communication, the server and devices only transmit a fraction of the weights that considerably reduce the communication latency. With the help of momentum and sparse communication, the convergence speed and latency of FL over wireless networks can be guaranteed, however, the accuracy decreases.

It is important to know that training and transmitting weights during FL may consume a large amount of energy. To deal with this issue, the energy consumption minimization problem was studied.

\paragraph{Energy Efficiency} To minimize the total energy consumption for local computation and wireless transmission, an iterative algorithm was proposed in \cite{Zhaohui}. The total energy consumption of all devices at each step is calculated as
\begin{equation}
    E = \sum\limits_{k=1}^{K}(E_k^C + E_k^T),
\end{equation}
where $E_k^C$ and $E_k^T$ are the local computation energy and wireless transmission energy of the $k$th device, respectively. To minimize the energy, closed-form solutions for the time allocation $t_k$, bandwidth allocation $b_k$, power control $p_k$, computation frequency $f_k$, and learning accuracy $\delta$ are derived. At each iteration, to optimize $(t_k, b_k, p_k, f_k, \delta)$, the authors first optimized $(t_k, \delta)$ under fixed $(b_k, p_k, f_k)$. Then, $(b_k, p_k, f_k)$ are updated based on the obtained $(t_k, \delta)$. Thus, the optimal solution of $(b_k, p_k, f_k)$ or $(t_k, \delta)$ can be obtained at each time slot. Also, the local and global model updatings still follow (6) and (7), respectively.

The works in \cite{Abad} and \cite{Zhaohui} mainly focused on accelerating the training tasks from the communication and energy perspective, i.e., minimizing the communication overhead and energy consumption. However, computation efficiency is also one of the major characteristics of FL over wireless networks, which may greatly affect learning performance. 

\paragraph{Computation Efficiency} There are mainly two ways to increase computation efficiency for FL in wireless networks, including deploying high computation units and efficient gradient descent methods.

\textbf{High Computation Unit:} In recent years, GPU has been proposed to accelerate the training latency and efficiency of FL. Authors in \cite{Jinke1} considered the training acceleration problem from the CPU to GPU under communication and computation resource constraints. Using the Karush-Kuhn-Tucker (KKT) conditions, the closed-form solutions of joint batch size selection and communication resource allocation were derived, and the relationship between training latency and training batch size was analyzed. The local and global updatings in \cite{Jinke1} still follow (6) and (7), respectively. Although using GPU for FL training improved the learning efficiency, authors in \cite{Jinke1} relied on an impractical assumption that each device was equipped with a GPU. The mobile device with GPU for training can cost a large amount of energy, especially for battery-limited devices.

\textbf{Gradient Descent:}
To improve the training performance of FL over wireless networks, several novel gradient descent methods have been studied. For the gradient descent methods in FL over wireless networks, the authors in \cite{Tao} and \cite{Abad} mainly deployed SGD and momentum gradient descent to update the local model, respectively. 

(1) SGD: Different from the gradient descent in (6), that calculated from the entire dataset, SGD randomly selects one data sample from the whole dataset at each time slot to reduce computation complexity \cite{Martin}. For the SGD in \cite{Tao}, the local model of the $k$th device in the $t$th time slot is updated as
\begin{equation}
    \bm{w}_k^{t+1} = \bm{w}_k^t - \eta \textbf{g}_{k,i}^t,
\end{equation}
where $i$ is the $i$th data sample of the $k$th device. 

(2) Momentum Gradient Descent: Momentum is a method that helps to accelerate the gradient descent in the relevant direction and dampens oscillations. This is achieved by adding a momentum term $\sigma$ of the update vector of the past time slots to the current update vector \cite{Sebastian}. For the gradient descent with momentum in \cite{Abad}, the local model is updated using
\begin{equation}
    \bm{w}_k^t = \bm{w}_k^{t-1} + \textbf{u}_k^t,
\end{equation}
where
\begin{equation}
    \textbf{u}_k^t = \sigma \textbf{u}_k^{t-1} + \textbf{g}_k^t.
\end{equation}
Also, the model aggregation of both gradient descent methods follows (7). Using SGD and momentum, a faster convergence rate can be achieved. Note that the gradient descent methods of FL methodologies mentioned in Section III can still be deployed in  FL over wireless networks.

Authors in \cite{Tao,Zhaohui,Jinke1,Abad} optimized the communication, energy, and computation efficiency, separately. In fact, the communication, energy, and computation efficiencies are correlated with each other, which requires joint design and optimization among them. 

\paragraph{Joint Design of Communication, Energy, and Computation Efficiency} In FL over wireless networks, there are two trade-offs, between computation and communication latencies, and between learning latency and device energy consumption. Authors in \cite{Tran} decomposed the problem into two sub-problems, which were the learning latency versus device energy consumption problem solved by the Pareto efficiency model, and the computation versus communication latencies problem, solved via finding the optimal learning accuracy with KKT conditions. The Pareto efficiency model minimizes the learning latency and does not increase the energy costs of each device \cite{Hashimzade}. Through iteratively obtaining the closed-form solutions of these two sub-problems, the authors characterized how the computation and communication latencies of mobile devices affect trade-offs between energy consumption, learning time, and learning accuracy. However, authors in \cite{Tran} relied on an impractical assumption that the channel state information (CSI) remained unchanged during the whole FL process. In \cite{Wadu}, the authors considered imperfect CSI. Under imperfect CSI, channels between the server and devices over each resource block (RB) are predicted using their past observations. Then, based on Lyapunov optimization, joint device scheduling and RB allocation policy were proposed to minimize the loss function in FL over wireless networks. The local and global model updatings in \cite{Tran,Wadu} still follow (6) and (7), respectively.

In summary, authors in \cite{Howard_yang,Jinke,Nishio,mingzhe,Tao,Zhaohui,Jinke1,Abad,Tran,Wadu} independently considered the model aggregation and communication, energy, and computation efficiency optimization. However, in practical scenarios, they are correlated with each other, and in the next section, we introduce the resource allocation of FL over wireless networks.

\subsubsection{Resource Allocation}
For the research on resource allocation of FL over wireless networks, it jointly designs the spectrum resource allocation and device selection. Due to the limited number of RBs in the uplink transmission, only a fraction of devices can transmit local models to the base station (BS). To solve this problem, a probabilistic device selection scheme was proposed in \cite{mingzhe1} to select the devices whose local models have significant effects on the global model. The update of the global model in the $t$th time slot is still updated as (7), while $p_k$ in (7) is calculated as
\begin{equation}
    p_k = \frac{a_{k}^{t}\bar{N}_k}{\sum_{i=1}^{K}a_{i}^{t}\bar{N}_i},
\end{equation}
where $a_{k}^{t}\in\{0,1\}$ is the device association index of the $k$th device, and $\bar{N}_k$ is the number of training data samples of the $k$th device. If $a_{k}^{t} = 1$, the $k$th device is selected to update the local model to the BS, vice versa. To select the optimal set of devices to upload the local models and minimize the uplink and downlink transmission latency, the optimization problem of a joint RB allocation and device selection scheme was proposed and written as
\begin{align}
    &\min_{\textbf{A},\textbf{R}}\sum\limits_{t=1}^{\hat{C}}T^{t}(\textbf{a}^{t},\textbf{R}^{t})I^{t}\\
    \text{s.t.}~&a_{k}^{t},r_{k,n}^{t},I^{t} \in \{0,1\},\\
    &\sum\limits_{t=1}^{K}r_{k,n}^{t}\leq 1,\\
    &\sum\limits_{n=1}^{R}r_{k,n}^{t} = a_{k}^{t},
\end{align}
where $T^{t}(\textbf{a}^{t},\textbf{R}^{t})$ is the transmission latency, $\textbf{A} = [\textbf{a}_1,...,\textbf{a}_{\hat{C}}]$ is the device selection matrix of all iterations, $\textbf{R} = [\textbf{R}_1,...,\textbf{R}_{\hat{C}}]$ is the RB allocation matrix for all devices of all iterations, $R$ is the number of RBs, and $\hat{C}$ is a constant, which is large enough for the proposed FL to converge. In (85), $I^{t} = 0$ means that the proposed FL converges, vice versa. In (86), $r_{k,n}^{t} = 1$ means that the $n$th RB is allocated to the $k$th device in the $t$th time slot, and (86) implies that at most one RB is allocated to the $k$th device. In (87), it means that all RBs should be allocated to the devices associated with the BS. To increase the convergence speed of FL, deep neural networks (DNNs) are used to predict the local models of devices that cannot transmit their local model weights. To enable the BS to predict the local model, each device should have a chance of connecting to the BS to provide local model weights for training DNNs. Thus, a probabilistic device association scheme was proposed as
\begin{equation}
    \bar{P}_{k}^{t} = \begin{cases}  \frac{\|\textbf{e}_{k}^{t}\|}{\sum\limits_{k=1,k\neq k^{*}}^{K}\|\textbf{e}_{k}^{t}\|}, &\text{if } k \neq k^{*}, \\
    1, & \text{if } k = k^{*},
    \end{cases}
\end{equation}
where $\bar{P}_{k}^{t}$ is the probability of the $k$th device connecting to the BS in the $t$th time slot, $\textbf{e}_{k}^{t}=\bm{w}_{G}^{t} - \bm{w}_{k}^{t+1}$ is the variation between the global model and the local model of the $i$th device, and $\|\textbf{e}_{k}^{t}\|$ is the norm of $\textbf{e}_{k}^{t}$. In (88), the association probability between the BS and the $k$th device increases as $\|\textbf{e}_{k}^{t}\|$ increases. Thus, the probability that the BS deploys the local model of the $k$th device to aggregate the global model increases. Through using the device association scheme in (88), the BS has a higher probability of selecting devices whose local models significantly affect the global model. In addition, the $k^{*}$th device is always connected to the BS to provide a local model for the prediction of other devices' local models. With the predicted local models, the global model is updated as
\begin{equation}
\bm{w}_G^{t+1} = \frac{\sum\limits_{k=1}^{K}\bar{N}_k a_{k}^{t}\bm{w}_{k}^{t} + \sum\limits_{k=1}^{K}\bar{N}_k(1-a_{k}^{t})\hat{\bm{w}}_{k}^{t}\mathbb{I}_{\{E_{k}^{t}\leq\gamma\}}}{\sum\limits_{k=1}^{K}\bar{N}_k a_{k}^{t} + \sum\limits_{k=1}^{K}\bar{N}_k(1-a_{k}^{t})\mathbb{I}_{\{E_{k}^{t}\leq\gamma\}}},
\end{equation}
where $\hat{\bm{w}}_{k}^{t}$ is the predicted local  model of the $k$th device, $\sum_{k=1}^{K}\bar{N}_k a_{k}^{t}\bm{w}_{k}^{t}$ is the sum of local models of the devices connected to the BS, $\sum_{k=1}^{K}\bar{N}_k(1-a_{k}^{t})\hat{w}_{k}^{t}\mathbb{I}_{\{E_{i}^{k}\leq\gamma\}}$ is the sum of the predicted local models of the devices are not connected to the BS, $E_{k}^{t}=\|\hat{\bm{w}}_{k}^{t}-\bm{w}_{k}^{t}\|^2$ is the prediction error, and $\gamma$ is the error threshold. If $E_{k}^{t}\leq\gamma$, the BS uses the predicted local model $\hat{\bm{w}}_{k}^{t}$ to update the global model, otherwise, not. From (89), we can obtain that the BS uses the predicted local models together with the transmitted local models to update the global model to decrease the FL training loss and improve the convergence speed.

\begin{figure}[!h]
    \centering
    \includegraphics[width=3.5 in]{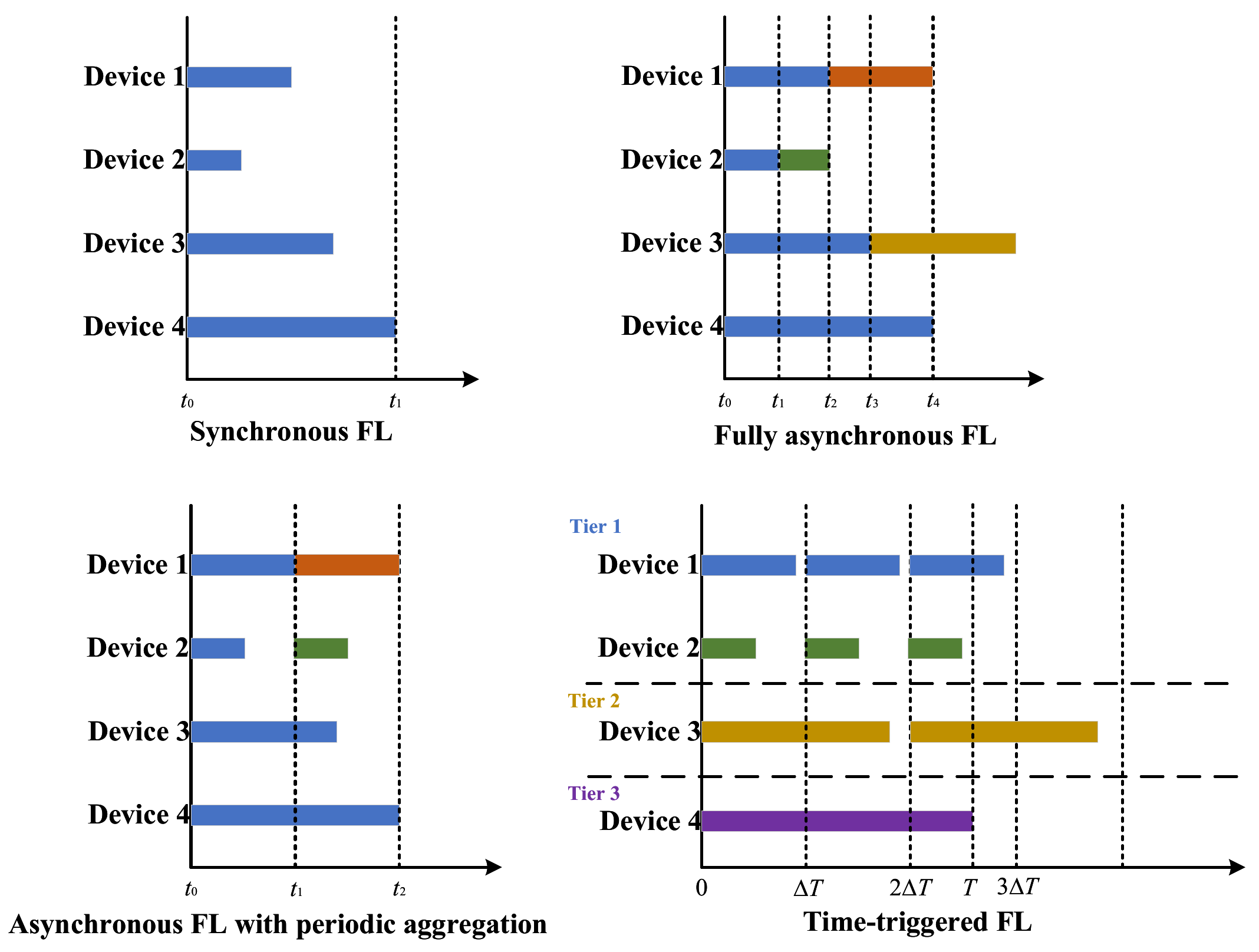}
    \caption{Illustration of synchronous FL, fully asynchronous FL, asynchronous FL with periodic aggregation, and time-triggered FL.}
    \label{basic_modules}
\end{figure}

\subsubsection{Asynchronous FL} Authors in \cite{Howard_yang,Jinke,Nishio,mingzhe,Tao,Zhaohui,Jinke1,Abad,Tran,Wadu,mingzhe1} mainly considered synchronous FL over wireless networks. One common problem of FL systems is the straggler issue. This problem originates from the fact that the time duration of each training round is strictly limited by the slowest participating device \cite{J_chen}. Two asynchronous FL policies are introduced in \cite{C_Xie} and \cite{CH_Hu}, which are fully asynchronous FL and asynchronous FL with periodic aggregation, as shown in Fig. 10. For the fully asynchronous FL in \cite{CH_Hu}, in each time slot, the server receives a locally trained model $\bm{w}_{\text{new}}$ from an arbitrary device and updates the global model $\bm{w}_{G}$ by weight averaging, which is denoted as
\begin{equation}
    \bm{w}_{G}^{t} = (1-\alpha)\bm{w}_{G}^{t-1} + \alpha\bm{w}_{\text{new}},
\end{equation}
where $\alpha\in(0,1)$ is a mixing hyperparameter to determine the contribution of the local model to the global model. Intuitively, a larger bias of the local model results in a higher error when updating the global model. The local model updating is still written as (6). Based on \cite{CH_Hu}, the fully asynchronous FL was used in \cite{Yu_chen} for edge devices with non-IID data, so that the server does not need to wait for the devices with high communication delays.

However, fully asynchronous FL with sequential updating has the problem of high communication costs caused by frequent local model updating and transmission. To address this issue, an asynchronous FL with periodic aggregation and an adaptive asynchronous FL (AAFL) were proposed in \cite{CH_Hu} and \cite{Jianchun}, respectively.

For the asynchronous FL with periodic aggregation in \cite{CH_Hu}, the edge server periodically collects local models to update the global model from devices that have completed local training. While other devices continue their local training without being interrupted or dropped. Particularly, after each device updates its local model by (6), it transmits a signal to the server indicating its completion of local model training. After each time duration $T$, the server schedules a subset of ready-to-update devices to upload their local models. The received local models are aggregated at the server by (7), and then the updated global model is distributed to these devices, and continue their local training based on the newly received global model. 

The AAFL algorithm in \cite{Jianchun} is an experience-driven algorithm based on deep reinforcement learning (DRL), which can adaptively determine the optimal fraction value $\alpha$ in each time slot. Given the completion time of the learning task, local model parameters, loss function, the difference between the current loss value and target loss value, bandwidth consumption, and remaining resource budget in each time slot, the DRL agent selects the value of $\alpha$ for model aggregation. Integrating AAFL with DRL reduces the training time and improves learning accuracy compared to fully asynchronous FL.

Based on the proposed synchronous and asynchronous FL, authors in \cite{Xzhou} proposed a time-triggered FL (TT-Fed) over wireless networks, which was a generalized form of classic synchronous and asynchronous FL and achieved a good balance between training and communication efficiencies. The global model aggregation in TT-Fed is triggered in each fixed global model aggregation round duration $\triangle T$. Assuming that $T$ is the time required for the slowest device to complete one single local updating round. Thus, all devices are partitioned into $M = \lceil \frac{T}{\triangle T}\rceil$ tiers ($\lceil .\rceil$ is the ceiling function), where the first tier is the fastest tier and the $M$th tier is the slowest tier. As shown in Fig. 9, assuming that 4 devices are partitioned into 3 tiers according to the global model aggregation round duration partitioning. Device 1 and device 2 in the first tier need a single global model aggregation  $\triangle T$ to complete their local updating, while device 4 in the third tier needs three $\triangle T$. Thus, the server has new updates from different tiers in each global model aggregation round. By using TT-Fed, it is possible for the global model to be broadcast to users in different tiers for communication overhead reduction.

Authors in \cite{Howard_yang,Jinke,Nishio,mingzhe,Tao,Zhaohui,Jinke1,Abad,mingzhe1,Jianchun,CH_Hu,C_Xie,Xzhou} considered OFDMA and TDMA to transmit the local or global models. However, when a large number of devices uploading high-dimensional local models, the classic orthogonal-access schemes, such as OFDMA and TDMA, are not able to scale well with an increasing number of devices. To deal with this issue, over-the-air computation (OAC) has been proposed, it is a disruptive technology for fast data aggregation in wireless networks through exploiting the waveform superposition property of multi-access channels. In particular, the transmitted signal in the uplink transmission is superimposed over-the-air and their weighted sums, so-called the aggregated signal, are processed at the edge \cite{W_liu,Letaief1}.

\begin{figure*}[!h]
    \centering
    \includegraphics[width=5.5 in]{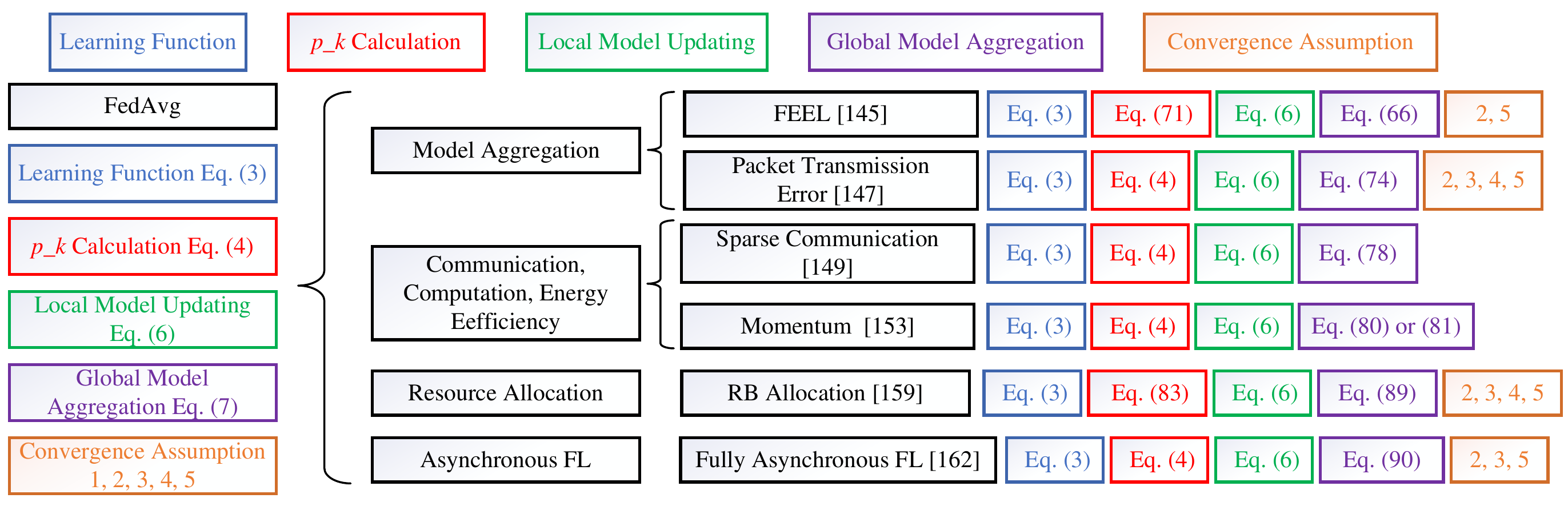}
    \caption{Joint design of FL and wireless networks.}
    \label{basic_modules}
\end{figure*}

\subsection{Analog Approach of FL over Wireless Networks}
In the digital approach, one challenge is to overcome the communication bottleneck, which is caused by many devices uploading high-dimensional models to a central server. A promising approach is to design new multiple access schemes, and a recently emerged approach, the so-called OAC, can provide the required scalability for FL over wireless networks.

OAC is a promising approach for fast wireless data aggregation via computing a nomographic function of distributed data from multiple devices in the uplink transmission, and it can accomplish the computation of target function by concurrent transmission, thereby significantly improving the communication efficiency compared to orthogonal transmission \cite{Goldenbaum}. The uplink transmission includes local model weights or gradients, while the downlink information usually includes updated model weights or aggregated gradients. 

The OAC-based approach for fast global model aggregation in the uplink transmission was proposed in \cite{kaiyang} to explore the superposition property of a wireless multi-access channel via the joint device selection and beamforming design. In the $t$th time slot, the $k$th device transmits the signal $\textbf{s}_k^t$ to the BS, and the received signal at the BS is expressed as
\begin{equation}
    \textbf{y} = \sum\limits_{k=1}^{K}\textbf{h}_kb_k\textbf{s}_k + \textbf{n},
\end{equation}
where $\textbf{h}_k$ is the channel vector between the $k$th device and the BS, $b_k$ is the transmitter scalar, and $\textbf{n}$ is the noise vector. Through designing the receiver beamforming vector $\textbf{m}$, the estimated global model at the BS is calculated as
\begin{equation}
    \hat{\bm{w}_G} = \frac{1}{\sqrt{\eta}}\textbf{m}^{H}\textbf{y},
\end{equation}
where $\eta$ is a normalizing factor. For the global model updating in the OAC, the difference between the estimated global model $\hat{\bm{w}}_G$ and the target function $\bm{w}_G$ should be minimized, and a mean-square-error (MSE) is used to quantify the performance, which is defined as
\begin{equation}
    \text{MSE}(\hat{\bm{w}}_g, \bm{w}_g) = \textit{E}(\|\hat{\bm{w}}_g - \bm{w}_g\|_2).
\end{equation}
Motivated by \cite{L_Chen}, given the receiver beamforming vector $\textbf{m}$, the MSE is minimized by using the zero-forcing transmitter
\begin{equation}
    b_k = \sqrt{\eta}p_k\frac{(\textbf{m}^{H}\textbf{h}_k)^{H}}{\|\textbf{m}^{H}\textbf{h}_k\|_2}.
\end{equation}

Based on \cite{kaiyang}, authors in \cite{guangxu} considered broadband analog aggregation (BAA) for OAC model aggregation to further maximize the number of scheduled devices under update-distortion constraints, which could reduce the communication latency. In BAA, the weights of local models are first modulated into symbols. Then, the symbol sequence is divided into blocks, and each block is transmitted in a single OFDM symbol over one frequency sub-channel. Sub-channels are inverted by power control, so that weights transmitted by different devices are received with identical amplitudes, achieving amplitude alignment at the receiver as required for BAA.

Different from BAA, only transmitting the sign of gradients in OAC also enables a large number of devices to participate in model aggregation and achieve convergence, which has been exploited in \cite{guanxu,C_ZHONG,N_ZHANG,Y_Xue}. In the downlink transmission, broadcasting updated model weights guarantees that the devices compute the gradients based on the same model weights. On the other hand, broadcasting aggregated gradients in multi-cell OAC promotes the personalization of model weights for devices located at the cell edge \cite{A_Sahin}.

According to \cite{guanxu,C_ZHONG,N_ZHANG,Y_Xue,A_Sahin}, various techniques have been developed to investigate the performance of OAC, which can find its application in distributed sensing and autonomous control \cite{Z_Wang}. In distributed sensing, OAC is able to achieve efficient distributed sensing by conducting simultaneous transmission among all sensors, and the desired
function can be directly computed over the air. In autonomous control, a group of agents desire to perform some actions
to accomplish an overall cooperative task by interacting with
each other. As a result, each agent needs to iteratively collect information from others for updating its own state until convergence, which generally consists of two phases within
each iteration, i.e., the communication phase for information exchanges and the computation phase for state updates. By
integrating these two phases, efficient network-wide consensus
can be achieved by OAC in a distributed manner. However, authors in \cite{guanxu,C_ZHONG,N_ZHANG,Y_Xue,A_Sahin} generally assumed some ideal communication conditions and ignore practical implementation issues.

Since OAC focuses on uplink
transmission, time-division duplexing (TDD) is commonly considered in the existing research works to achieve local CSI estimation at each device based
on the pilot signal broadcast by the edge server. In TDD systems, the uplink and downlink transmissions are carried out in the same frequency channel but in different time slots. Therefore, the edge server first broadcasts pilot signals to devices for estimating the local CSI, and then obtains the global CSI based on the feedback from devices by assuming the channel reciprocity. However, the communication overhead is linearly scaling with the number of devices for CSI feedback, which may lead high transmission latency in ultra-dense networks. To address this issue, effective channel feedback approaches should be developed to realize OAC while avoiding massive overhead for CSI gathering. In \cite{9838988}, random orthogonalization was proposed for FL in massive MIMO systems, which significantly reduced the channel estimation overhead while achieving OAC model aggregation without requiring transmitter side CSI.

Furthermore, in practical systems, the transceivers can only obtain imperfect CSI because of multiple reasons, such as inaccurate channel estimation and finite-rate feed back. Therefore, simply implementing the system design based on the assumption of perfect CSI may lead to poor learning performance in practical wireless networks with imperfect CSI. To address this issue, robust design of OAC is required to achieve a reliable functional computation under imperfect CSI. In \cite{yulin}, a maximum-likelihood estimation design for misaligned AirComp with residual channel-gain variation and
symbol-timing asynchrony among devices was proposed. To further address the error propagation and noise enhancement problems at the maximum likelihood estimator, a whitened matched filtering and sampling (WMFS) scheme was deployed.

In addition, most of the current design for OAC relies on the instantaneous CSI at the edge server or devices. However, the cumbersome CSI acquisition introduces extra transmission latency and communication overhead, which motivates the CSI-free design, so-called blind design for OAC. To mitigate the possible destructive signal superposition due to the phase difference when CSI is not available, authors in \cite{9641940}, \cite{9771881} considered the one-bit quantization to realize blind OAC, where the receiver obtains the sign of aggregated signals based on majority vote by detecting the energy accumulated on different OFDM subcarriers \cite{9641940} and superposed pulse-position modulation symbols \cite{9771881}, which avoided the reliance on the CSI and relaxed the requirement of synchronization. In addition, the balanced number system was considered in \cite{A_Sahin_new} to enable continuous-valued computations in the digital OAC system without the need of CSI acquisition.

It should be noted that FL with OAC inherits the common problems in FL literature such as convergence under different data distributions, device heterogeneity, stragglers, data privacy, and various security issues. Therefore, these application-specific challenges need to be re-evaluated for a given OAC scheme.

\subsection{Convergence Analysis of FL over Wireless Networks}
The derivation of convergence analysis of FL is still based on the fundamental assumptions 1 - 4 in Section III. Different from the convergence analysis of FL methodologies, researchers usually derive the upper bound of $\mathbb{E}\{f(\bm{w}) - f(\bm{w}^{*})\}$ or $\mathbb{E}\{\frac{1}{T}\sum_{t=0}^{T-1}\|\bm{g}^{t}\|\}$ in FL over wireless networks. We will present several examples as follows.

In \cite{mingzhe}, except from assumptions 1 - 4, authors added another assumption in the following.

$\textbf{Assumption 5.}$ We assume that $f_{n}(\bm{w})$ is strongly convex with positive parameter $\mu$, such that
\begin{equation}
    f(\bm{w}^{t+1}) \!\geq\! f(\bm{w}^{t}) \!+\! (\bm{w}^{t+1} \!\!- \!\bm{w}^{t})^{T}\nabla f(\bm{w}^{t}) \!+\! \frac{\mu}{2}\|\bm{w}^{t+1}\!\! - \!\bm{w}^{t}\|^2\!\!.
\end{equation}
The upper bound of $\mathbb{E}\{f(\bm{w}) - f(\bm{w}^{*})\}$ is derived as
\begin{align}
    &\mathbb{E}\{f(\bm{w}^{t+1}) - f(\bm{w}^{*})\} \leq B^t\mathbb{E}\{f(\bm{w}^{0}) - f(\bm{w}^{*})\}\nonumber\\
    & + \frac{2G}{LD}\sum_{k=1}^{K}D_k(1 - a_k + a_kq_k(r_k,P_k))\frac{1-B^t}{1 - B},
\end{align}
where $D$ is the total number of training data, $a_k$, $q_k$, $r_k$, and $P_k$ have already been defined in (75) - (78), and $B$ is denoted as
\begin{equation}
    B = 1 - \frac{\mu}{L} + \frac{4\mu G}{LD}\sum_{k=1}^{K}D_k(1 - a_k + a_kq_k(r_k.P_k)).
\end{equation}
It is observed that there is a gap $\frac{2G}{LD}\sum_{k=1}^{K}D_k(1 - a_k + a_kq_k(r_k,P_k))\frac{1-B^t}{1 - B}$ between $\mathbb{E}\{f(\bm{w}^{t+1})\}$ and $\mathbb{E}\{ f(\bm{w}^{*})\}$, which is caused by the packet errors and device selection policy. The gap decreases when the packet error rate decreases or the number of devices participating the model aggregation increases. Also, the value of $B$ decreases with decreasing packet error rate, which means that the convergence rate of FL improves.

In \cite{guanxu}, based on assumptions 2 - 4, the convergence rate of OAC in FL over AWGN channels is given by
\begin{align}
    \mathbb{E}\{\frac{1}{T}\sum_{t=0}^{T-1}\|\bm{g}^{t}\|\}\leq&\frac{a_{\text{AWGN}}}{\sqrt{T}}(\sqrt{L}(f(\bm{w}^{0}) - f(\bm{w}^{*}) + \frac{\gamma}{2}) +\nonumber\\ &\frac{2\gamma}{\sqrt{K}}\hat{\sigma} + b_{\text{AWGN}}),
\end{align}
where the scaling factor $a_{\text{AWGN}}$ and the bias term $b_{\text{AWGN}}$ are denoted as
\begin{equation}
    a_{\text{AWGN}} = \frac{1}{1 - \frac{1}{K\sqrt{\hat{\rho}}}},~b_{\text{AWGN}}\frac{2\gamma\hat{\sigma}}{K\sqrt{\hat{\rho}}}.
\end{equation}
In (98) and (99), $\gamma > 0$ is a positive constant, and $\hat{\rho}$ denotes the receive SNR. (98) means that the existence of channel noise slows down the convergence rate by adding a scaling factor $a_{\text{AWGN}}$ and a positive bias term $b_{\text{AWGN}}$ to the upper bound on the time-averaged gradient norm. Therefore, more communication rounds are required for convergence. However, the negative effect of channel noise vanishes at a scaling rate of $\frac{1}{K}$ with increasing number of devices participating in the model aggregation.

\subsection{Summary and Lessons Learned}
In this section, we have reviewed two main research areas of FL over wireless networks. We summarize the approaches along with references. From this review, we gather the following lessons learned:

\begin{itemize}
    \item Wireless factors, such as transmission power, wireless channel, and spectrum resource allocation, affect the convergence and learning accuracy of FL over wireless networks. The evolution of FedAvg in wireless networks is shown in Fig. 11. According to Fig. 11, through jointly or separately optimizing equations (3), (4), (6), and (7), FL can be more appropriate for wireless networks. In addition, theoretical convergence analysis is usually derived based on five fundamental assumptions.
    \item Synchronous FL systems are susceptible to the straggler effect. As a result, asynchronous FL has been proposed to solve the problem. Asynchronous FL allows devices to participate in model training even a training round is in progress. This is more reflective of practical FL scenarios and is an important contribution towards guaranteeing the scalability of FL. However, synchronous FL is still the most common approach used because of its convergence guarantees. Therefore, the convergence analysis in asynchronous FL settings needs to be investigated.
    \item For the research works we have discussed in this section, most works assume that the wireless transmission between servers and devices is successful and error-free. However, in wireless networks, because of network congestion, interference, and bit error, the model transmission between servers and devices may fail or contain errors, which further affects learning performance. Therefore, robust FL algorithms over wireless networks should be investigated to address the issue.
    \item In this section, we observe that FL usually focuses on model training for a single ML task across multiple devices, namely, the well-trained model weights cannot generalize to multiple tasks. To address this issue, meta learning is discussed in the following sections.
\end{itemize}

\section{Meta Learning Methodologies}
Meta learning is most commonly understood as learning to learn, which refers to a learning algorithm that can generalize across different tasks. Thus, it has an advantage over traditional data-driven ML algorithms that can only work on a single task with the well-trained weights $\bm{w}$ \cite{jaoquin,Hospedales,peng}. Therefore, meta learning is able to help FL adapt to multiple tasks. Specifically, in meta learning, we evaluate the performance of weights $\bm{w}$ over a distribution of tasks $p(\mathcal{T})$. We loosely assume a task consisting of a dataset and loss function $\mathcal{T}=\{\mathcal{D}, \mathcal{L}\}$. The objective of meta learning is to minimize the expectation of loss function over all tasks, which is given by
\begin{equation}
    \min_{\bm{w}}\mathbb{E}_{\mathcal{T}\sim p(\mathcal{T})}\mathcal{L}(\mathcal{D};\bm{w}),
\end{equation}
where $\mathcal{L}(\mathcal{D};\bm{w})$ measures the performance of a model trained using weights $\bm{w}$ on dataset $\mathcal{D}$. To solve the problem in (100), a set of $M$ source tasks are sampled from $p(\mathcal{T})$ and used in the meta-training stage as $\mathcal{D}_{source}=\{(\mathcal{D}_{source}^{train},\mathcal{D}_{source}^{test})^{i}\}_{i=1}^{M}$, where each task has both training and testing data. Also, the source training and testing datasets are usually called support and query sets, respectively. The meta-training step of ``learning how to learn'' is written as
\begin{equation}
    \bm{w}^{*} = \arg\max_{\bm{w}}\log p(\bm{w}|\mathcal{D}_{source}).
\end{equation}
Then, a set of $Q$ target tasks used in the meta-testing stage is denoted as $\mathcal{D}_{target}=\{(\mathcal{D}_{target}^{train},\mathcal{D}_{target}^{test})^{i}\}_{i=1}^{Q}$, where each task has both training and testing datasets. In the meta-testing stage, we use the learned weights $\bm{w}$ to train the weights of each new target task $i$, which is denoted as
\begin{equation}
    \bm{\theta}^{*(i)} = \arg\max_{\bm{\theta}}\log p(\bm{\theta}|\bm{w}^{*}, \mathcal{D}_{target}^{train(i)}).
\end{equation}
According to (102), we can obtain that learning on the training set of a target task $i$ benefits from meta-knowledge $\bm{w}^{*}$, and evaluate the accuracy of meta-learner by the performance of $\bm{\theta}^{*(i)}$ on the testing dataset of each target task $\mathcal{D}_{target}^{test(i)}$. Meta-learning algorithms can be categorized into three main directions: (1) metric-based, (2) model-based, and (3) gradient-based optimization methods, which are introduced in detail in the following subsections. The introduced meta learning methodologies are fundamental meta learning algorithms from the CS community without considering any wireless factors.

\begin{figure}[!h]
    \centering
    \includegraphics[width=3.5 in]{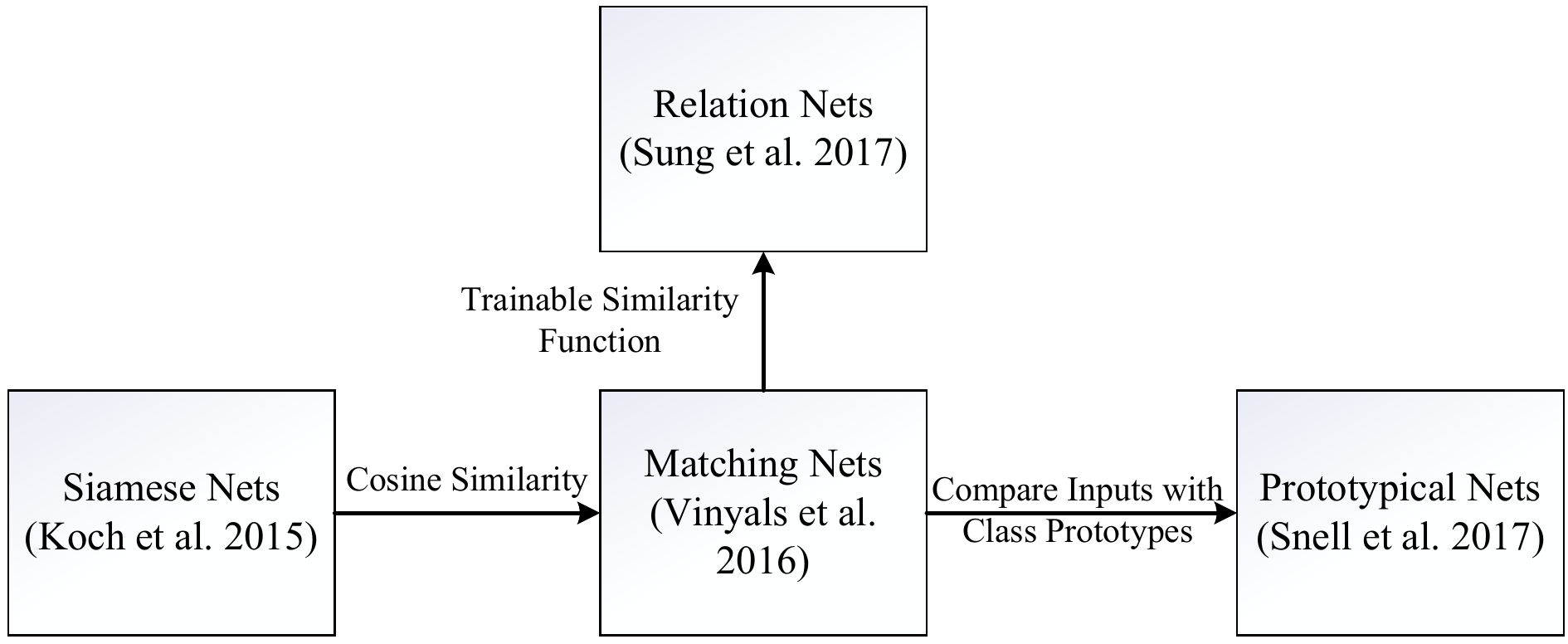}
    \caption{Development of metric-based meta learning.}
    \label{basic_modules}
\end{figure}

\subsection{Metric-based Meta Learning}
Metric-based methods learn the meta knowledge $\bm{w}$ through a feature space that is used for various new tasks. The feature space is integrated with the weights $\bm{\theta}$ of the neural networks. Then, new tasks are learned by comparing new inputs with example inputs in the meta-learned feature space. The higher the similarity between the new input and the example input, the more likely that the new input has the same label as the example input. Thus, metric-based meta-learning aims to learn a similarity kernel that takes two inputs, and outputs their similarity score. Larger similarity scores present larger similarities. In this subsection, we introduce four key metric-based meta-learning methods, including siamese networks, matching networks, prototypical networks, and relation networks. The relationship among these four methods is presented in Fig. 12.

\subsubsection{Siamese Networks}
Authors in \cite{Koch} used a siamese network to compare the distance between data samples. A siamese network consists of two neural networks that share the same weights $\bm{\theta}$. It takes two inputs $\bm{x}_1$ and $\bm{x}_2$, and computes two hidden states $f_{\bm{\theta}}(\bm{x}_1)$ and $f_{\bm{\theta}}(\bm{x}_2)$. Then, these two hidden states are input into a distance layer to calculate a distance vector, which is given by
\begin{equation}
    \bm{d} = |f_{\bm{\theta}}(\bm{x}_1) - f_{\bm{\theta}}(\bm{x}_2)|.
\end{equation}
According to the distance vector $\bm{d}$, we can obtain whether two inputs $\bm{x}_1$ and $\bm{x}_2$ belong to the same class. The siamese network is a simple approach in metric-based meta-learning, and can only be deployed to supervised learning scenarios.

\subsubsection{Matching Networks}
Based on the distance comparison idea in siamese networks, authors in \cite{Vinyals} proposed a matching network to learn the similarity between support sets and new inputs from query sets. The matching networks use a weighted combination of all example labels in the support set and an attention kernel to compute the similarity of inputs $\bm{x}_i$ and new input $\bm{x}$. The attention kernel uses the cosine distance \cite{Cosine} to calculate the similarity of the input representations, rather than using the distance vector in (103) to quantify the similarity of two inputs. The matching network is still a simple approach in metric-based meta-learning, and is not applicable outside of the supervised learning scenarios. Furthermore, it suffers from performance degradation when label distributions are biased.

\subsubsection{Prototypical Networks}
Similar to matching networks, prototypical networks proposed in \cite{Snell} also used samples in the support set. However, rather than calculating the similarity between samples in the support set and new inputs, prototypical networks map inputs to a dimensional vector space such that inputs of a given output class are close together. Since the number of class prototypes is smaller than that of samples in the support set, the amount of comparisons decreases, which further reduces computational costs. However, prototypical networks can only be used in supervised learning scenarios.

\subsubsection{Relation Networks}
Different from the pre-defined similarity metric in siamese and matching networks, relation networks (RN) proposed in \cite{Sung} used a trainable similarity metric. RN consists of two modules, which are an embedding module responsible for embedding inputs, and a relation module computing similarity scores between new inputs $\bm{x}$ and example inputs $\bm{x}_i$ from support sets. Then, a classification decision is made by selecting the class of the example input which outputs the highest similarity score. RN uses the Mean-Squared Error (MSE) as a similarity score, and the MSE is then propagated backward through the entire network to update the weights in embedding and relation modules. Because of the trainable similarity metric, the accuracy performance of RN is better than that of siamese and matching networks with a fixed similarity metric.

\begin{figure}[!h]
    \centering
    \includegraphics[width=3.5 in]{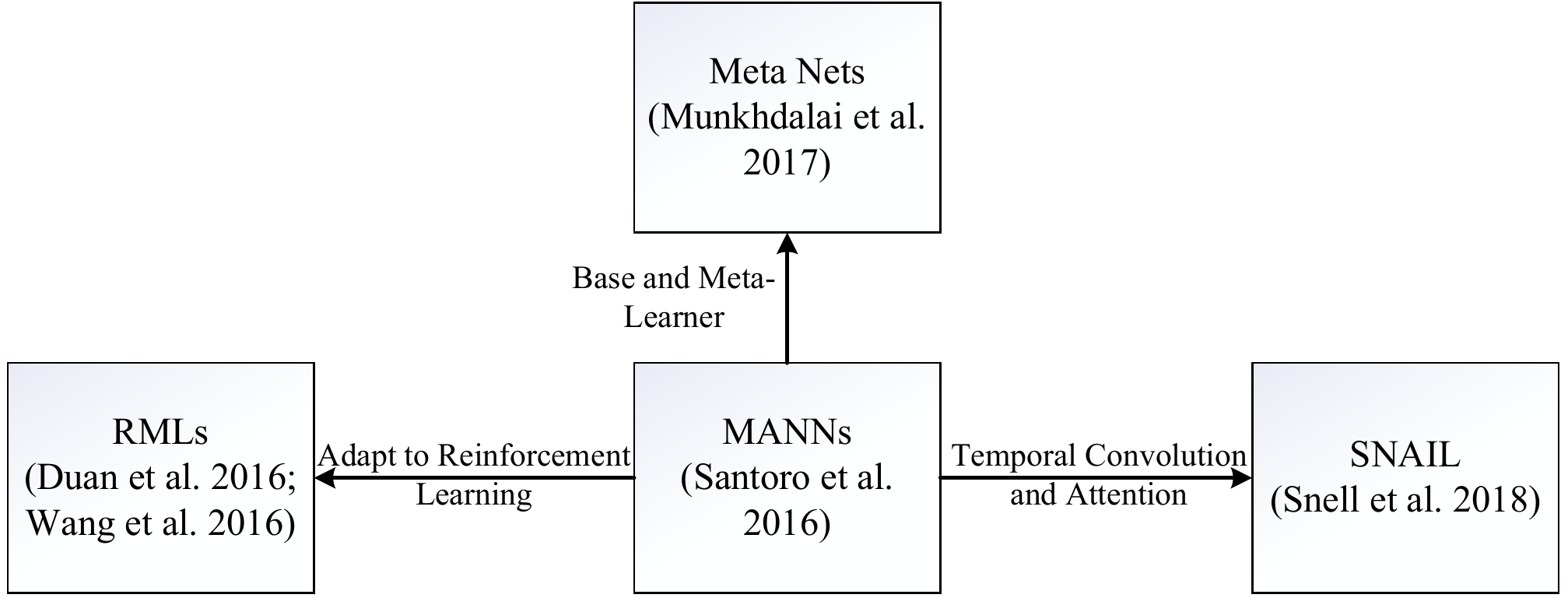}
    \caption{Development of model-based meta learning.}
    \label{basic_modules}
\end{figure}

\begin{figure}[!h]
    \centering
    \includegraphics[width=3.5 in]{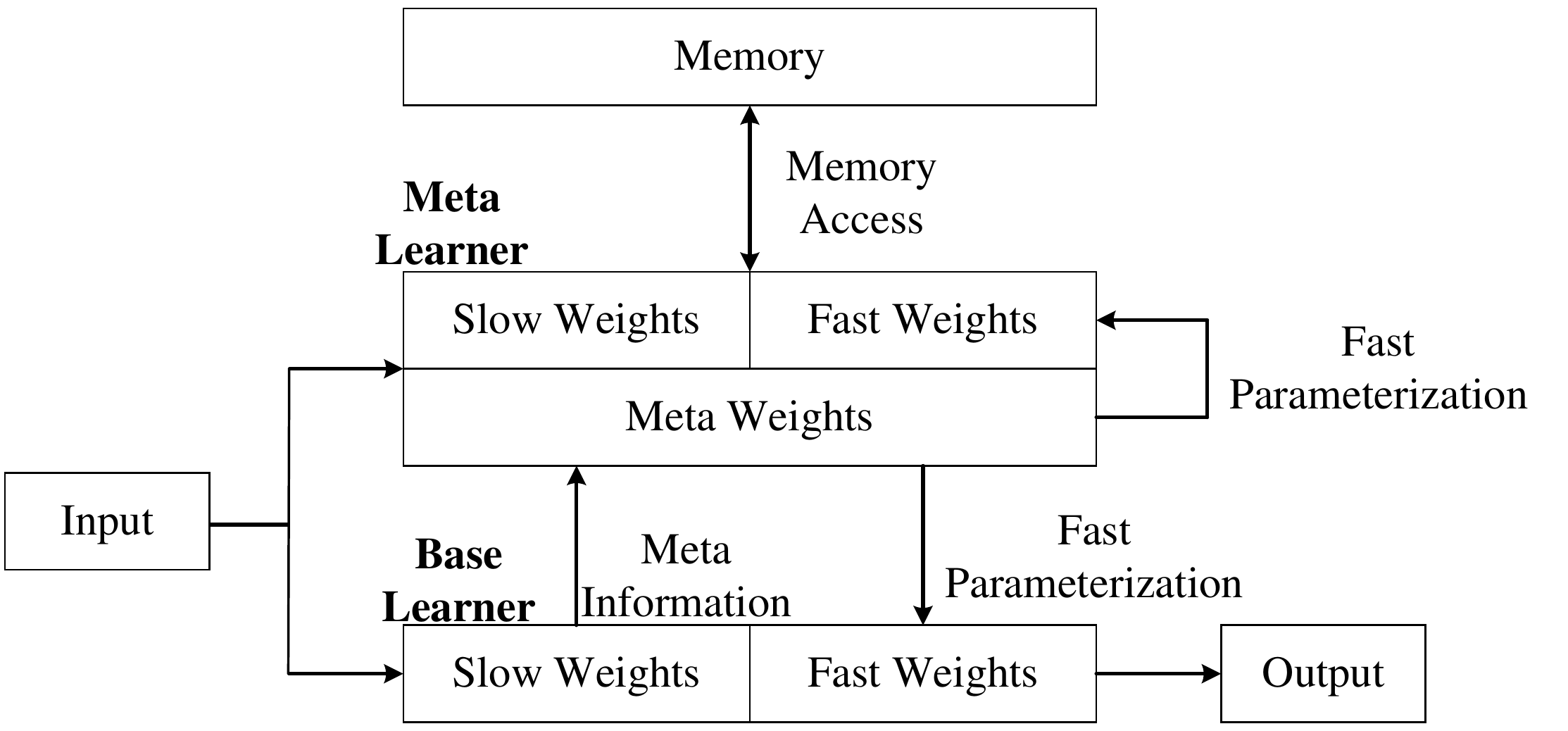}
    \caption{Architecture of meta network.}
    \label{basic_modules}
\end{figure}

\subsection{Model-based Meta Learning}
In contrast to the metric-based approaches deploying fixed neural networks at the testing phase, model-based meta-learning algorithms depend on the internal state of each task. Specifically, model-based approaches process the support set in a sequential fashion. In each time slot, the internal state captures relevant task-specific information with the given inputs, which can be used to make predictions for new inputs. Meanwhile, task information from previous inputs should be remembered, so that model-based methods have a memory component. In this subsection, we introduce four key model-based meta-learning methods, including memory-augmented neural networks (MANNs), meta networks (MetaNets), recurrent meta-learners (RMLs), and simple neural attentive meta-learner (SNAIL). The relationship of these four networks are shown in Fig. 13.

\subsubsection{Memory-augmented Neural Networks}
MANNs were proposed in \cite{Santoro} to allow for quick task-specific adaptation with the help of a neural turing machine (NTM) \cite{Graves} and an external memory. The learning procedure of MANNs is that the data of a task is processed as a sequence. First, the support set is input to MANN. Then, the query set is evaluated. The interaction between NTM and external memory is that NTM gradually accumulates meta knowledge across tasks, and the external memory helps to store the obtained knowledge. Given new inputs, NTM leverages the previously obtained meta knowledge stored in the external memory to make predictions. MANNs integrate the external memory and a neural network to achieve meta learning. Different from metric-based meta-learning, MANN can be used for both classification and regression problems. However, it has higher architectural complexity.

\subsubsection{Meta Networks}
Similar to MANN, MetaNets proposed in \cite{Munkhdalai} also leveraged an external memory to store the meta knowledge. However, different from MANN, MetaNets are divided into two distinct subsystems, which are base-learner and meta-learner, as shown in Fig. 14. The base-learner is used to perform tasks, and provide meta knowledge for the meta-learner. Then, the meta-learner calculates fast task-specific weights for itself and the base-learner. The training of MetaNet consists of three main procedures: (1) Acquisition of meta knowledge; (2) Generation of fast weights; (3) Optimization of slow weights. MetaNets depend on base-learner and meta-learner for each task. Although it can be used for both supervised and reinforcement learning scenarios, the learning architecture is quite complex and leads to a high burden on memory usage and computation time.

\subsubsection{Recurrent Meta-learner}
RMLs proposed in \cite{Abbeel1} and \cite{Jane1} were meta-learners based on recurrent neural networks (RNNs), and were specifically proposed for reinforcement learning scenarios. The internal learning architecture of the selected RNN allows for fast adaptation to new tasks. Similar to MANN, RML still uses memory to store the meta knowledge and the task data is sequentially input into the learning model. However, RMLs have simple learning architectures, mainly perform well on simple reinforcement learning tasks, and cannot be adapt to complex learning scenarios.

\subsubsection{Simple Neural Attentive Meta-learner}
Similar to MANN, SNAIL proposed in \cite{Mishra1} still processes task data in sequence. However, rather than using external memory, SNAIL deploys a special model architecture to serve as memory. The special model consists of 1D temporal convolutions \cite{Oord1} and a soft attention mechanism \cite{Ashish}. The 1D convolutions are used for memory access, and the attention mechanism allows SNAIL to pinpoint specific experiences. Furthermore, SNAIL contains three building blocks, which are DenseBlock, TCBlock, and AttentionBlock. The DenseBlock deploys a single 1D convolution to the input and connects to the result, the TCBlock consists of a series of DenseBlocks, and the AttentionBlock learns the important parts of prior experience. A key advantage of SNAIL is that it can be used for both supervised and reinforcement learning scenarios, and it achieves better learning accuracy performance than that of the other three model-based meta-learning algorithms.

\begin{figure}[!h]
    \centering
    \includegraphics[width=3.5 in]{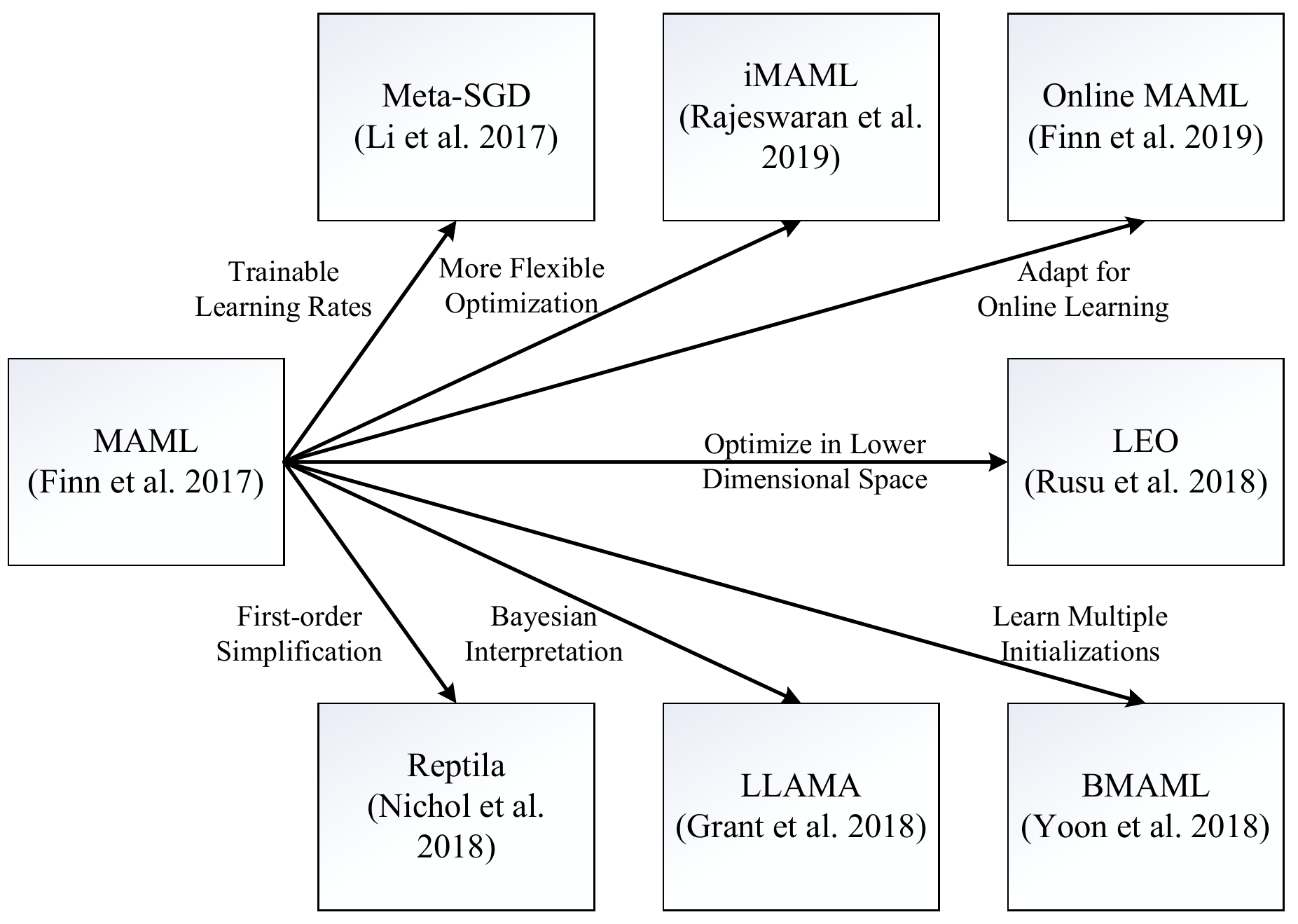}
    \caption{Development of MAML.}
    \label{basic_modules}
\end{figure}

\begin{figure}[!h]
    \centering
    \includegraphics[width=3.5 in]{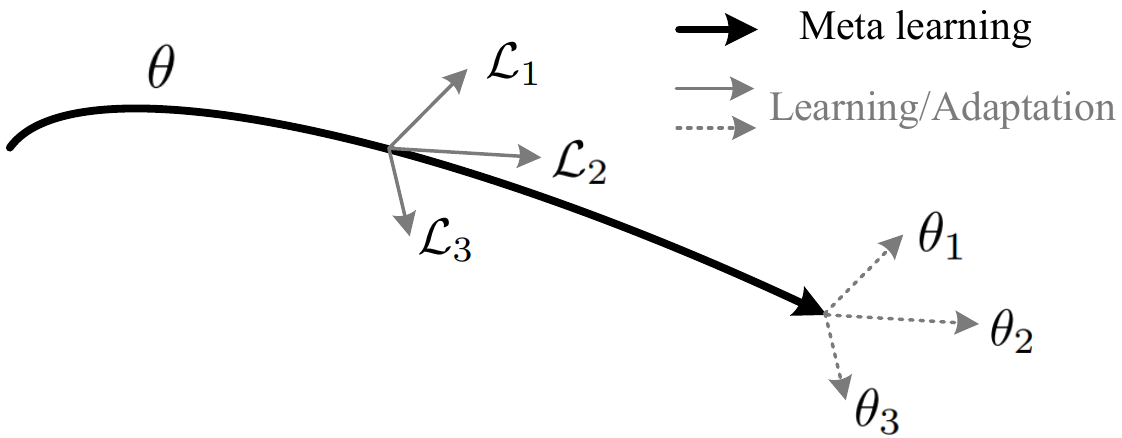}
    \caption{Diagram of MAML.}
    \label{basic_modules}
\end{figure}

\subsection{Gradient-based Meta Learning}
Different from the metric-based and model-based meta learning approaches, gradient-based meta-learning is mainly trained with an interleaved training procedure, including inner loop of task-specific adaptation and outer loop of meta initialization training \cite{Zintgraf,Kaiyi}. The traditional gradient-based meta learning is model-agnostic meta learning (MAML).  Based on MAML, several other advanced meta learning models have been proposed, which are meta-SGD, reptile, Bayesian MAML (BMAML), Laplace approximation for meta adaptation (LLAMA), latent embedding optimization (LEO), MAML with Implicit Gradients (iMAML), and online MAML, as shown in Fig. 15.

\subsubsection{MAML} MAML is a model and task-agnostic algorithm for meta-learning that trains model weights with a small number of gradient steps and leads to fast learning on a new task \cite{MAML}. Thus, MAML has two advantages: 1) it can be fine-tuned, which means that it quickly adapts to new tasks, and 2) it requires fewer training samples \cite{Chua}. The learning trend of MAML is shown in Fig. 16. From the learned initialization weights $\bm{\theta}$, MAML can quickly move to the optimal set of weights $\bm{\theta}_{i}^{*}$ for the task $\mathcal{T}_i$ $(i=1, 2, 3)$.

The meta learning model is represented by a function $f_{\bm{\theta}}$ with weights $\bm{\theta}$. When it adapts to a new task $\mathcal{T}_i$, the model weights $\bm{\theta}$ become $\bm{\theta}_{i}^{'}$. The updated weights $\bm{\theta}_{i}^{'}$ is computed by using gradient descent on task $\mathcal{T}_i$, which is denoted as
\begin{equation}
    \bm{\theta}_{i}^{'} = \bm{\theta} - \alpha\nabla_{\bm{\theta}}\mathcal{L}_{\mathcal{T}_i}(f_{\bm{\theta}}).
\end{equation}
The step size $\alpha\in(0,1)$ can be fixed as a hyperparameter or dynamically meta-learned. The learning model weights are trained by optimizing the performance of $f_{\bm{\theta}_{i}^{'}}$ with respect to $\bm{\theta}$ across the tasks sampled from $p(\mathcal{T})$. More concretely, the meta-objective is to minimize the loss function of all tasks with the learned initialization weights $\bm{\theta}$, which is presented as
\begin{equation}
    \min_{\bm{\theta}}\sum_{\mathcal{T}_i\sim p(\mathcal{T})}\mathcal{L}_{\mathcal{T}_i}(f_{\bm{\theta}_{i}^{'}}) = \sum_{\mathcal{T}_i\sim p(\mathcal{T})}\mathcal{L}_{\mathcal{T}_i}(f_{\bm{\theta}-\alpha\nabla_{\bm{\theta}}\mathcal{L}_{\mathcal{T}_i}(f_{\bm{\theta}})}),
\end{equation}
where $\mathcal{L}_{\mathcal{T}_i}(f_{\bm{\theta}_{i}^{'}})$ is the loss function of the $i$th task $\mathcal{T}_i$ with its model weights $\bm{\theta}_{i}^{'}$, and $\bm{\theta}_{i}^{'}$ is updated by (104). According to (104) and (105), the optimization of meta learning is performed over the initialized model weights $\bm{\theta}$, and the objective is achieved by the updated model weights $\bm{\theta}^{'}$. The initialized model weights $\bm{\theta}$ are updated as
\begin{equation}
    \bm{\theta} = \bm{\theta} - \beta\nabla_{\bm{\theta}}\sum_{\mathcal{T}_i\sim p(\mathcal{T})}\mathcal{L}_{\mathcal{T}_i}(f_{\bm{\theta}_{i}^{'}}),
\end{equation}
where $\beta$ is the meta step size, and $\nabla_{\bm{\theta}}\sum_{\mathcal{T}_i\sim p(\mathcal{T})}\mathcal{L}_{\mathcal{T}_i}(f_{\bm{\theta}_{i}^{'}})$ is calculated as
\begin{equation}
    \nabla_{\bm{\theta}}\!\!\!\!\sum_{\mathcal{T}_i\sim p(\mathcal{T})}\!\!\!\!\mathcal{L}_{\mathcal{T}_i}(f_{\bm{\theta}_{i}^{'}}) = \!\!\!\!\sum_{\mathcal{T}_i\sim p(\mathcal{T})}\!\!\!\!(\textbf{I} - \alpha\nabla^{2}f_{i}(\bm{\theta}))\nabla f_{i}(\bm{\theta} - \alpha\nabla f_{i}(\bm{\theta})).
\end{equation}
In (107), $\textbf{I}$ is an identity matrix. The detailed procedures of MAML are introduced in $\textbf{Algorithm}$ 2. Based on the same assumptions of FL methodologies in Section III, the upper bound of $\mathbb{E}\|\nabla\mathcal{L}( f_{\bm{\theta}})\|$ is derived as
\begin{equation}
    \mathbb{E}\|\nabla\mathcal{L}( f_{\bm{\theta}})\|\leq \mathcal{O}\left(\sqrt{\frac{\hat{\sigma}^2}{B} + \frac{\hat{\sigma}^2}{BD_{\text{o}}} +\frac{\hat{\sigma}^2}{D_{\text{in}}}}\right) + \epsilon, 
\end{equation}
where $B$ is the number of tasks, $D_{\text{o}}$ is the size of datasets of outer loop, and $D_{\text{in}}$ is the size of datasets of inner loop, $\hat{\sigma}$ is defined in assumption 4, and $0 < \epsilon < 1$. (108) means that if the batch sizes $B$, $D_{\text{o}}$, and $D_{\text{in}}$ are selected properly, for any $\epsilon > 0$, MAML is able to converge after limited number of iterations.

\begin{algorithm}[t]
\begin{algorithmic}[1]
\caption{Model-Agnostic Meta-Learning}
\STATE $p(\mathcal{T}$):distribution over tasks.
\STATE Initialize step size hyperparameters $\alpha$ and $\beta$, randomly initialize learning weights $\bm{\theta}$.
\WHILE{not done}
    \STATE Sample batch of tasks $\mathcal{T}_i\sim P(\mathcal{T})$.
    \FOR{all $\mathcal{T}_i$}
        \STATE Evaluate $\nabla_{\bm{\theta}}\mathcal{L}_{\mathcal{T}_i}(f_{\bm{\theta}})$ with respect to $K$ data samples.
        \STATE Calculate adapted weights with gradient descent: $\bm{\theta}_{i}^{'} = \bm{\theta} - \alpha\nabla_{\bm{\theta}}\mathcal{L}_{\mathcal{T}_i}(f_{\bm{\theta}})$.
    \ENDFOR
    \STATE Update $\bm{\theta} = \bm{\theta} - \beta\nabla_{\bm{\theta}}\sum_{\mathcal{T}_i\sim p(\mathcal{T})}\mathcal{L}_{\mathcal{T}_i}(f_{\bm{\theta}_{i}^{'}})$.
\ENDWHILE
\end{algorithmic}
\end{algorithm}

However, calculating Hessian vector $\nabla^{2}f_{i}(\bm{\theta})$ in (107) increases computation complexity, which can result in high computation latency \cite{yifanhu}. To address this issue, first-order MAML (FO-MAML) was proposed in \cite{FOnichol}, where the authors directly ignored $\nabla^{2}f_{i}(\bm{\theta}))$. Surprisingly, the convergence performance of this method is nearly the same as that with full second derivatives, suggesting that most of the performance improvement in MAML comes from the gradients of the objective at the first-order update, rather than the second updates from differentiating through the gradient update. Previous research works observed that ReLU neural networks were almost locally linear, which suggested that the second derivatives may be close to zero in most cases, partially explaining the performance improvement of the first-order approximation \cite{Goodfellow1}. In addition, the first-order approximation can achieve $33\%$ speed-up in terms of network computation.

To further reduce the computation time, authors in \cite{Fallah1} introduced Hessian-free MAML (HF-MAML), which did not require computation of the Hessian vectors, and its computation complexity was the same as that of FO-MAML, but it achieved better convergence rate than FO-MAML. It is because Hessian-free is a method to avoid the vanishing gradient problem while using backpropagation in DNNs  \cite{Martens,Martens1,Kiro,Martens3}. The idea behind HF-MAML is that for any function $\phi$, the product of its Hessian $\nabla^{2}\phi(\bm{\theta})$ and any vector $\textbf{v}$ can be approximated as
\begin{equation}
    \nabla^{2}\phi(\bm{\theta})\textbf{v} \approx \left[\frac{\nabla\phi(\bm{\theta} + \delta\textbf{v}) - \nabla\phi(\bm{\theta} - \delta\textbf{v})}{2\delta}\right],
\end{equation}
with an error of at most $\rho\delta\|\textbf{v}\|^2$, where $\rho$ is the parameter for Lipschitz continuity of the Hessian of $\phi(\bm{\theta})$. By integrating (109) into (107), we can obtain that
\begin{equation}
    \phi(\bm{\theta}) = f_{i}(\bm{\theta}), \textbf{v} = \nabla f_{i}(\bm{\theta} - \alpha\nabla f_{i}(\bm{\theta})).
\end{equation}
Thus, $\nabla^{2}f_{i}(\bm{\theta})\nabla f_{i}(\bm{\theta} - \alpha\nabla f_{i}(\bm{\theta}))$ in (107) is calculated by (110), which decreases the computation complexity.

\subsubsection{Meta-SGD} Meta-SGD was proposed in \cite{zhenguo}, which is an easily trainable meta-learner that could initialize and adapt any learner in just one step, for both supervised learning and reinforcement learning. Different from MAML, Meta-SGD can achieve a much higher accuracy not only by the learner initialization, but also by the learner update direction and learning rate, all in a single meta learning process. The main difference is the way it updates $\bm{\theta}_{i}^{'}$, different from MAML that updates $\bm{\theta}_{i}^{'}$ based on (104), Meta-SGD updates $\bm{\theta}_{i}^{'}$ using
\begin{equation}
    \bm{\theta}_{i}^{'} = \bm{\theta} - \bm{\alpha}\circ\nabla_{\bm{\theta}}\mathcal{L}_{\mathcal{T}_i}(f_{\bm{\theta}}),
\end{equation}
where $\bm{\alpha}$ is a vector of the same size as $\bm{\theta}$ that determines both the update direction and learning rate, and $\circ$ denotes the element-wise product. The adaptation term $\bm{\alpha}\circ\nabla_{\bm{\theta}}\mathcal{L}_{\mathcal{T}_i}(f_{\bm{\theta}})$ is a vector whose direction denotes the update direction. Note that learning rate $\bm{\alpha}$ in (111) is a vector rather than a scalar in (104), and (111) allows for a higher flexibility in the sense that each weight has its own learning rate. Compared with MAML, Meta-SGD is easier to implement and can learn more efficiently due to that both of its update direction and learning rate can be optimized.

\begin{figure}[!h]
    \centering
    \includegraphics[width=3.5 in]{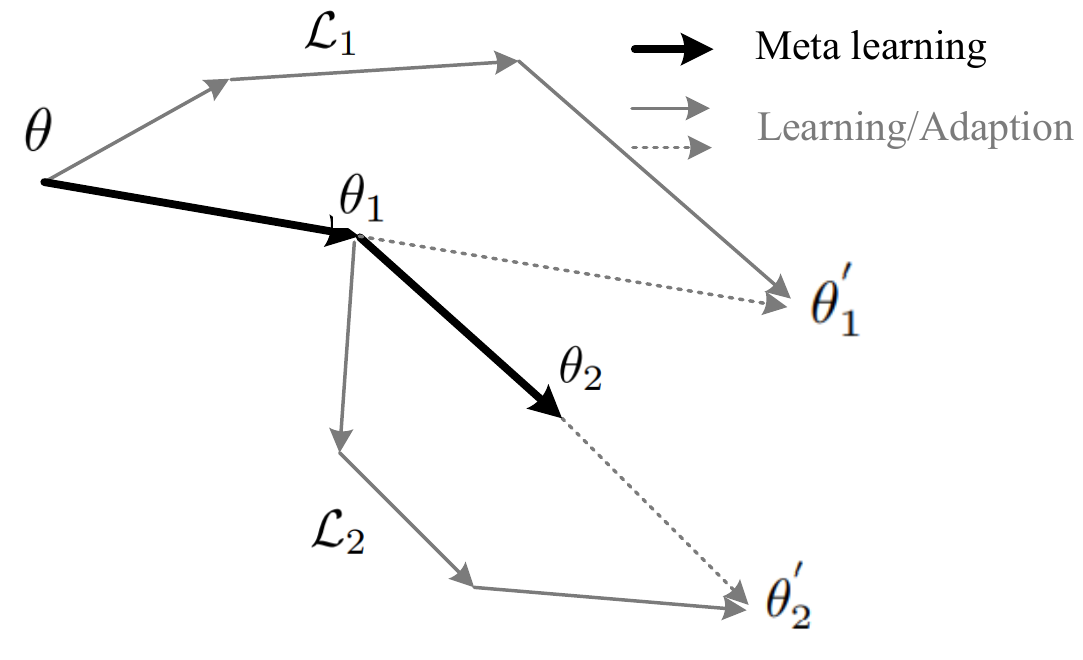}
    \caption{Diagram of Reptile.}
    \label{basic_modules}
\end{figure}

\subsubsection{Reptile} Like MAML, Reptile learns a weight initialization that can be fine-tuned quickly on a new task. However, the way in which Reptile tries to obtain the optimal weights is quite different from MAML. Given the initialized model weight $\bm{\theta}$, it works by repeatedly sampling only one task in each time slot, training on it, and moving the initialized weights towards the trained weights on that task \cite{reptile}. As shown in Fig. 17, the initialized weights $\bm{\theta}$ are moving towards the optimal weights for tasks 1 or 2. Because in each time slot, only one task is selected to train the learning weights, the initialized weights $\bm{\theta}$ oscillate between tasks 1 and 2. For example, if selecting task 1 to train $\bm{\theta}$, $\bm{\theta}$ is moving towards $\bm{\theta}_{1}^{'}$ and is updated as $\bm{\theta}_{1}$ after several iterations. Then, if stop using task 1 and selecting task 2 to train $\bm{\theta}_{1}$, $\bm{\theta}_{1}$ is moving towards $\bm{\theta}_{2}^{'}$ and is updated as $\bm{\theta}_{2}$ after several iterations. Therefore, Reptile is a very simple meta-learning algorithm, and does not require updating model weights through the optimization process like MAML, making it more suitable for optimization problems where a limited number of update steps are required, and saving time and memory costs. In Reptile, $\bm{\theta}_{i}^{'}$ for the $i$th task is updated using (111). However, for the initialization weights $\bm{\theta}$, it moves toward to the trained weights, which is updated as
\begin{equation}
    \bm{\theta} = \bm{\theta} - \beta(\bm{\theta}_{i}^{'} - \bm{\theta} ).
\end{equation}
In (112), $\bm{\theta}_{i}^{'} - \bm{\theta}$ is the distance between initialization weights $\bm{\theta}$ and learning weights for the $i$th task $\bm{\theta}_{i}^{'}$. The Reptile algorithm is shown in $\textbf{Algorithm}$ 3. Although the Reptile is an extremely simple meta learning technique, the convergence and accuracy performance may be a bit worse than that of MAML because of its simple learning procedure.

\begin{algorithm}[t]
\begin{algorithmic}[1]
\caption{Reptile Meta-Learning}
\STATE $p(\mathcal{T}$):distribution over tasks
\STATE Initialize step size hyperparameters $\beta$, randomly initialize learning weights $\bm{\theta}$
\FOR{$i$ = 1,2,...}
    \STATE Sample batch of tasks $\mathcal{T}_i\sim P(\mathcal{T})$.
    \STATE Calculate $\bm{\theta}_{i}^{'}$ via (111).
    \STATE Calculate initialization weights $\bm{\theta}$ via (112). 
\ENDFOR
\end{algorithmic}
\end{algorithm}

\subsubsection{BMAML} Unlike MAML that learns a distribution over potential solutions, Bayesian MAML (BMAML) in \cite{Yoon} learns $M$ possible weights $\bm{\Theta} = \{\bm{\theta}\}_{i=1}^{M}$ and jointly optimizes them in parallel. To update these weights, authors in \cite{Yoon} deployed Stein Variational Gradient Descent (SVGD) \cite{qiangliu}. SVGD is a non-parametric variational inference method, which leverages the advantages of Markov chain Monte Carlo (MCMC) \cite{Geyer} and variational inference. Also, it converges faster than MCMC because its update rule is deterministic. Specifically, SVGD maintains $M$ instances of model weights, called particles. In the $t$th time slot, each model weight vector $\bm{\theta}_t\in\bm{\Theta}$ is updated using
\begin{equation}
    \bm{\theta}_{t+1} = \bm{\theta}_t + \alpha\phi(\bm{\theta}_t),
\end{equation}
where $\alpha$ is learning rate, and $\phi(\bm{\theta}_t)$ is given as
\begin{equation}
    \phi(\bm{\theta}_t)\!\! =\!\! \frac{1}{M}\!\!\sum_{m=1}^{M}\!\!\left[k(\bm{\theta}_t^{m}, \bm{\theta}_t)\nabla_{\bm{\theta}_t^{m}}\log p(\bm{\theta}_t^{m}) \!+\! \nabla_{_{\bm{\theta}_t^{m}}}k(\bm{\theta}_t^{m}, \bm{\theta}_t)\right].
\end{equation}
Here, $k(\bm{\theta}_t^{m}, \bm{\theta}_t)$ in (114) is a similarity kernel between $\bm{\theta}_t^{m}$ and $\bm{\theta}_t$. In (114), the update of one particle relies on the other gradients of particles, $k(\bm{\theta}_t^{m}, \bm{\theta}_t)\nabla_{\bm{\theta}_t^{m}}\log p(\bm{\theta}_t^{m})$ moves the particle in the direction of gradients of other particles based on particle similarity, and $\nabla_{_{\bm{\theta}_t^{m}}}k(\bm{\theta}_t^{m}, \bm{\theta}_t)$ enforces repulsive force between particles so that they do not collide to a same point. Then, these particles are used to approximate the probability distribution of labels in testing datasets, which is denoted as
\begin{equation}
    p(y_{j}^{\text{test}}|\bm{\theta}_{j}^{'}) = \frac{1}{M}\sum_{m=1}^{M}p(y_{j}^{\text{test}}|\bm{\theta}_{\mathcal{T}_j}^{m}),
\end{equation}
where $\bm{\theta}_{\mathcal{T}_j}^{m}$ is the $m$th particle calculated by training the support dataset (training dataset) $\mathcal{D}_{\mathcal{T}_j}^{S}$ of the task $\mathcal{T}_j$, and $p(y_{j}^{\text{test}}|\bm{\theta}_{\mathcal{T}_j}^{m})$ is the data likelihood of the task $\mathcal{T}_j$.

To train BMAML, authors in \cite{Yoon} proposed a novel meta loss, called Chaser Loss. This loss aims to minimize the distance between the approximated parameter distribution achieved from the support set $p_{\mathcal{T}_j}^{n}(\bm{\theta}_{\mathcal{T}_j}|\mathcal{D}^{S}, \bm{\Theta}_{0})$ and true distribution $p_{\mathcal{T}_j}^{n+s}(\bm{\theta}_{\mathcal{T}_j}|\mathcal{D}^{S}\cup\mathcal{D}^{Q})$. Here, $n$ is the number of SVGD steps, and $\bm{\Theta}_{0}$ is the set of initial particles. Because the true distribution is unknown, we need to approximate it by running SVGD for $s$ additional steps to obtain $\bm{\Theta}_{\mathcal{T}_j}^{n+s}$, where $s$ additional steps are executed on both the support and query sets. The proposed meta-loss is written as
\begin{equation}
    \mathcal{L}_{\text{BMAML}}(\bm{\Theta}_0) = \sum_{\mathcal{T}_j\in B}\sum_{m=1}^{M}\|\bm{\theta}_{\mathcal{T}_j}^{n,m} - \bm{\theta}_{\mathcal{T}_j}^{n+s,m}\|_{2}^{2},
\end{equation}
where $B$ is the number of sampled tasks. The BMAML algorithm is shown in $\textbf{Algorithm}$ $\textbf{4}$, where $d(\bm{\Theta}_{\mathcal{T}_j}^{n}(\bm{\Theta}_0), \bm{\Theta}_{\mathcal{T}_j}^{n+s}(\bm{\Theta}_0))$ is the dissimilarity between two distributions $\bm{\Theta}_{\mathcal{T}_j}^{n}(\bm{\Theta}_0)$ and $\bm{\Theta}_{\mathcal{T}_j}^{n+s}(\bm{\Theta}_0)$.

BMAML is a robust optimization-based meta learning that can generate $M$ potential solutions for a task. However, it has to store $M$ parameter sets in memory over time, which results in substantial memory costs.

\begin{algorithm}[t]
\begin{algorithmic}[1]
\caption{Bayesian MAML}
\STATE Initialize $\bm{\Theta}_0$
\FOR{t = 1,... until convergence}
    \STATE Sample a batch of tasks $B$ from $p(\mathcal{T})$.
    \FOR{task $\mathcal{T}_j\in B$}
    \STATE Calculate $\bm{\Theta}_{\mathcal{T}_j}^{n}(\bm{\Theta}_0) = \text{SVGD}_{n}(\Theta_{0}; \mathcal{D}_{\mathcal{T}_j}^{S},\alpha)$.
    \STATE Calculate $\bm{\Theta}_{\mathcal{T}_j}^{n+s}(\bm{\Theta}_0) = \text{SVGD}_{s}(\bm{\Theta}_{\mathcal{T}_j}^{n+s}(\bm{\Theta}_0); \mathcal{D}_{\mathcal{T}_j}^{S}\cup\mathcal{D}_{\mathcal{T}_j}^{Q},\alpha)$.
    \ENDFOR
    \STATE $\bm{\Theta}_0 = \bm{\Theta}_0 - \beta\nabla_{\bm{\Theta}_0}\sum_{\mathcal{T}_j\in B}d(\bm{\Theta}_{\mathcal{T}_j}^{n}(\bm{\Theta}_0), \bm{\Theta}_{\mathcal{T}_j}^{n+s}(\bm{\Theta}_0))$.
\ENDFOR
\end{algorithmic}
\end{algorithm}

\subsubsection{LLAMA} Authors in \cite{Grant} reformulated MAML as a method for probabilistic inference in a hierarchical Bayesian model. Through integrating MAML into a probabilistic framework, a probability distribution over task-specific weights $\bm{\theta}_j^{'}$ is learned, and multiple potential solutions can be obtained for a task. This extended MAML is called Laplace approximation for meta adaptation (LLAMA). To minimize the error on the query set $\mathcal{D}_{\mathcal{T}_j}^{Q}$, the model must output large probability scores for true classes. The log-likelihood loss function is denoted as
\begin{equation}
    \mathcal{L}_{\mathcal{D}_{\mathcal{T}_j}^{Q}}(\bm{\theta}_{j}^{'}) = - \sum_{(\textbf{x}_i, y_i)\in\mathcal{D}_{\mathcal{T}_j}^{Q}}\log P(y_i|\textbf{x}_i,\bm{\theta}_j^{'}).
\end{equation}
To predict the correct label $y_i$, authors in \cite{Grant} deployed ML-Laplace to compute task-specific weights $\bm{\theta}_{j}^{'}$ updated from the initialization weights $\bm{\theta}$, and estimated the negative log-likelihood. The detailed LLAMA algorithm is shown in \textbf{Algorithms 5}  and \textbf{6}.

LLMAMA extends MAML in a probabilistic style, which means that there are multiple potential solutions for a single task. However, it can only be deployed for supervised learning with high computational costs, and the Laplace approximation in ML-LAPLACE may be inaccurate, which further decreases the accuracy.

\begin{algorithm}[t]
\begin{algorithmic}[1]
\caption{LLAML}
\STATE Randomly initialize $\bm{\Theta}$.
\WHILE{not converge}
    \STATE Sample a batch of tasks $B$ from $p(\mathcal{T})$.
    \STATE Estimate $\mathbb{E}_{(\textbf{x}_i, y_i)\sim p_{\mathcal{T}_j}}[-\log P(y_i|\textbf{x}_i, \bm{\theta})]$ using ML-LAPLACE in Algorithm 6.
    \STATE $\bm{\theta} = \bm{\theta} - \beta\nabla_{\bm{\theta}}\sum_{j}\mathbb{E}_{(\textbf{x}_i, y_i)\sim p_{\mathcal{T}_j}}[-\log P(y_i|\textbf{x}_i, \bm{\theta})]$.
\ENDWHILE
\end{algorithmic}
\end{algorithm}

\begin{algorithm}[t]
\begin{algorithmic}[1]
\caption{ML-LAPLACE}
\STATE $\bm{\theta}_{j}^{'} = \bm{\theta}$.
\FOR{k = 1,...,K}
    \STATE $\bm{\theta}_j^{'} = \bm{\theta}_j^{'} + \alpha\nabla_{\bm{\theta}_j^{'}}\log P(y_i\in\mathcal{D}_{\mathcal{T}_j}^{S}|\bm{\theta}_j^{'}, \textbf{x}_i\in\mathcal{D}_{\mathcal{T}_j}^{S})$.
\ENDFOR
\STATE According to \cite{Grosse}, calculate curvature matrix $\hat{\textbf{H}} = \nabla_{\bm{\theta}_j^{'}}^2[-\log P(y_i\in\mathcal{D}_{\mathcal{T}_j}^{Q}|\bm{\theta}_{j}^{'}, \textbf{x}_i\in\mathcal{D}_{\mathcal{T}_j}^{Q})] + \nabla_{\bm{\theta}_j^{'}}^2[-\log P(\bm{\theta}_j^{'}|\bm{\theta})]$.
\RETURN $-\log P(y_i\in\mathcal{D}_{\mathcal{T}_j}^{Q}|\bm{\theta}_{j}^{'}, \textbf{x}_i\in\mathcal{D}_{\mathcal{T}_j}^{Q}) + \eta\log(\det(\hat{\textbf{H}}))$.
\end{algorithmic}
\end{algorithm}

\begin{figure}[!h]
    \centering
    \includegraphics[width=3.5 in]{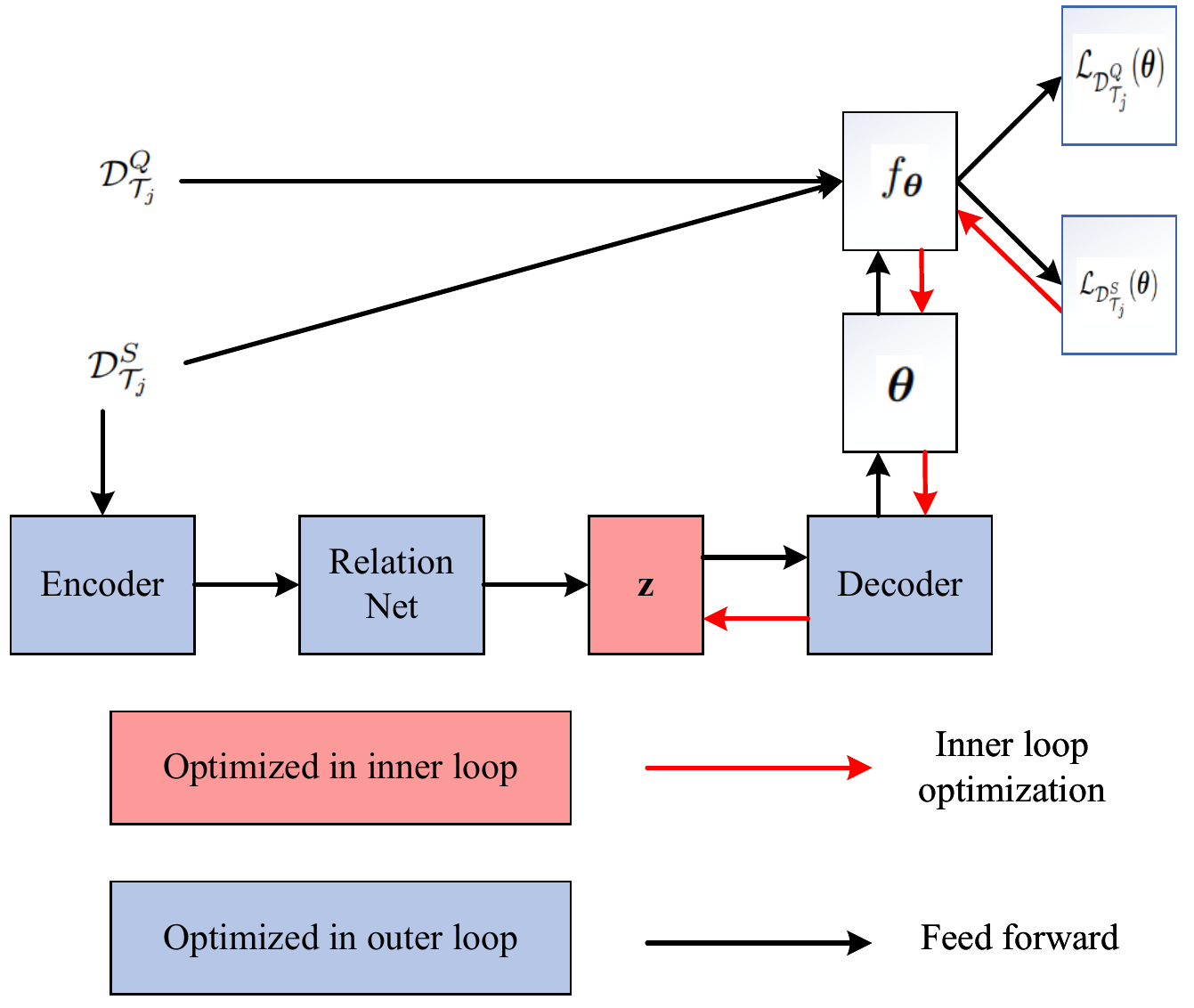}
    \caption{Diagram of LEO.}
    \label{basic_modules}
\end{figure}

\subsubsection{LEO} MAML operates in a high-dimensional parameter space using gradient information from only a few data samples from the support set, which can result in poor generalization. To deal with this issue, authors in \cite{Rusu} proposed a latent embedding optimization (LEO) to learn a lower dimensional latent embedding space, which indirectly updates a set of initialized weights $\bm{\theta}$. The detailed procedures of LEO are shown in Fig. 18. Given a task $\mathcal{T}_i$, the data samples from the support set pass through a stochastic encoder to produce
$(Nk)^{2}$ pairs of hidden codes, where $N$ is the number of classes in the support set, and $k$ is the number of data samples per class. Then, these paired codes are input into a relation network \cite{Sung}. The outputs are grouped by class, and parameterized by a probability distribution over latent codes $\textbf{z}_n$ for class $n$ in a low dimensional space. The decoder further generates a task-specifc model weights $\bm{\theta}_n$ for class $n$. The loss from the generated weights is propagated backward to update the model weights. In practice, generating a such high-dimensional set of parameters from a low-dimensional space is quite problematic. Thus, LEO uses pre-trained models, and only generates weights for the final layer, which limits the dimension of the model.

The key advantage of LEO is that it optimizes model weights in a lower dimensional latent embedding space, which improves generalization performance. However, it is more complex than that of MAML, because it needs to encode and decode the data samples.

\subsubsection{iMAML} Because of the higher-order derivatives, MAML with Implicit Gradients (iMAML) was considered in \cite{Rajeswaran1} to deal with the issue of long optimization path in MAML caused by gradient degradation problems, such as vanishing and exploding gradients \cite{Antoniou,Flennerhag}. Authors in \cite{Rajeswaran1} integrated regularization into the objective of MAML to guarantee appropriate learning while avoiding over-fitting. The objective of  iMAML is formulated as
\begin{equation}
    \min_{\bm{\theta}_{i}^{'}}\mathcal{L}(\bm{\theta}_{i}^{'},\mathcal{D}_k) + \frac{\lambda}{2}\|\bm{\theta}_{i}^{'} - \bm{\theta}\|^2,
\end{equation}
where $\lambda$ is a scalar hyperparameter that controls the regularization strength. In (118), the regularization term $\frac{\lambda}{2}\|\bm{\theta}_{i}^{'} - \bm{\theta}\|^2$ encourages $\bm{\theta}_{i}^{'}$ to remain close to $\bm{\theta}$. The regularization strength $\lambda$ plays an important role similar to the learning rate $\alpha$ in MAML, controlling the strength of prior model weights $\bm{\theta}$ relative to the dataset $\mathcal{D}_{\mathcal{T}}$. Ideally, the objective in (118) is solved by iteratively performing gradient descent to obtain the optimal $\bm{\theta}_{i}^{'}$. However, authors in \cite{Rajeswaran1} considered an implicit Jacobian to obtain $\bm{\theta}_{i}^{'}$ as
\begin{equation}
\frac{\partial\bm{\theta}_{i}^{'}}{\partial\bm{\theta}} = \left(\textbf{I} + \frac{1}{\lambda}\nabla_{\bm{\theta}}^{2}\mathcal{L}_i(\bm{\theta}_i)\right)^{-1}.
\end{equation}
According to \cite{Flake} and \cite{Griewank}, Jacobian only depends on the final result of the algorithm, and not the path taken by the algorithm, thus, it effectively decouples the meta-gradient computation from the choice of inner loop optimizer.

iMAML significantly decreases memory costs because it does not need to store Hessian matrices like MAML, allowing for a higher flexibility in the selection of the inner loop optimizer. However, the computational costs are the same as that of MAML.

\subsubsection{Online MAML} MAML assumes that a large set of tasks are available for meta training. However, in a practical system, tasks are likely available sequentially, which means that tasks may reveal one after the other. To deal with this issue, online meta learning was proposed in \cite{Rajeswaran}. The objective of online meta learning is to minimize the regret, where the regret is defined as the difference between the loss of the meta-learner and the best performance achievable from online learning with non-convex loss functions \cite{xianggao}. This objective is captured by the regret over the entire sequence, and is denoted as
\begin{equation}
    \text{Regret}_{T} = \sum_{t=1}^{T}\mathcal{L}_{\mathcal{T}_t}(\bm{\theta}_{t}^{'}) - \min_{\bm{\theta}}\sum_{t=1}^{T}\mathcal{L}_{\mathcal{T}_t}(\bm{\theta}_{t}),
\end{equation}
where $\sum_{t=1}^{T}\mathcal{L}_{\mathcal{T}_t}(\bm{\theta}_{t}^{'})$ reflects the accumulative loss calculated by the updated weights, and $\min\limits_{\bm{\theta}}\sum_{t=1}^{T}\mathcal{L}_{\mathcal{T}_t}(\bm{\theta}_{t})$ presents the minimum obtainable loss from a fixed set of initial model weights. The goal for the meta learner in  (120) is to sequentially obtain model weights $\bm{\theta}_{t}^{'}$ that perform well on the loss sequence. To update $\bm{\theta}_{t+1}^{'}$, one of the simplest algorithms is following the leader (FTL) \cite{Hannan,Kalai}, which updates the weights using
\begin{equation}
    \bm{\theta}_{t+1}^{'} = \arg\min_{\bm{\theta}}\sum_{k=1}^{t}\mathcal{L}_{\mathcal{T}_k}(\bm{\theta}_{k}).
\end{equation}
The gradient descent to perform meta update is given by
\begin{equation}
    \bm{\theta}_{t+1} = \bm{\theta}_{t} - \beta\nabla_{\bm{\theta}}\mathbb{E}_{\mathcal{T}_k\thicksim p_t(\mathcal{T})}\mathcal{L}_{\mathcal{T}_k}(\bm{\theta}_{k}),
\end{equation}
where $p_t({\mathcal{T}})$ is a uniform distribution over tasks in the $t$th time slot, and $\beta$ is the meta learning rate. In online meta learning, memory usage keeps increasing over time. This is because in each time slot, the incoming tasks and their corresponding datasets are stored in memory, which is used to obtain the model weights. The summary of gradient-based meta learning is summarized in Table VI.

\begin{table*}
\centering
\caption{Summary of Gradient-based Meta Learning Algorithms}
\begin{tabular}[c]{c|c|c|c}
\hline
\hline Gradient-based Meta Learning & Advantage & Disadvantage & Condition \\
\hline MAML \cite{MAML} & \makecell{Fine-tuned quickly \\Require fewer training samples} & \makecell{High computation complexity \\High computation latency\\High memory costs} & \makecell{Classification, regression\\ and reinforcement learning}\\
\hline Meta-SGD \cite{zhenguo} & \makecell{Trainable learning rate\\Higher accuracy} & Large number of model weights & \makecell{Supervised learning\\ Reinforcement learning} \\
\hline Reptile \cite{reptile} & \makecell{First-order simplification\\Low time and memory costs} & \makecell{Low convergence rate\\ Low learning accuracy} & \makecell{Supervised learning\\ Reinforcement learning}  \\ 
\hline BMAML \cite{Yoon} & \makecell{Learn multiple initializations} & \makecell{High memory costs} & \makecell{Supervised learning\\ Reinforcement learning} \\
\hline LLAMA \cite{Grant} & \makecell{Bayesian interpretation\\Multiple solutions for a single task} & \makecell{High computaion costs} & Supervised learning\\
\hline LEO \cite{Rusu} & Optimize in lower dimensional space & \makecell{High computation complexity \\ Limited applicability to few-shot learning} & \makecell{Supervised learning\\ Reinforcement learning}\\
\hline iMAML \cite{Rajeswaran1} & \makecell{ Low memory costs} & \makecell{High computation costs} & \makecell{Supervised learning\\ Reinforcement learning}\\
\hline Online MAML \cite{Rajeswaran} & Adapt to online learning & \makecell{High computation complexity \\High computation latency\\High memory costs} & \makecell{Classification, regression\\ and reinforcement learning}\\
\hline
\hline
\end{tabular}
\end{table*}

\subsection{Summary and Lessons Learned}
In this section, we have reviewed three types of meta learning methodologies. We summarize the approaches along with references. From this review, we gather the following lessons learned:

\begin{itemize}
    \item Metric-based meta learning learn a feature space that is deployed to classification tasks based on input similarity scores. The advantages of metric-based meta learning are that (1) the approach of similarity-based classification is simple and (2) the computation latency at the testing phase is low when tasks are small, it is because the learning model does not need to make task-specific adjustments. However, when the tasks at the testing phase are distant from the tasks used in the training phase, the learning accuracy decreases. In addition, metric-based meta learning is usually used in supervised learning scenarios.
    \item In model-based meta learning, tasks are processed and represented in the state of the model-based system, which is then used to make classifications. Advantages of model-based meta learning are the flexibility of the internal dynamics of systems, and their broader applicability compared to metric-based meta learning. Unfortunately, the learning performance of model-based meta learning is worse than that of metric-based meta learning. Also, model-based meta learning is usually used in supervised learning scenarios.
    \item Gradient-based meta learning aims to learn new tasks quickly. The key advantage of gradient-based meta learning is that it can achieve much better learning performance on more task distributions than metric-based and model-based meta learning. However, gradient-based meta learning optimizes a meta-learner for each task which results in high computation and memory costs. In addition, gradient-based meta learning is usually used in supervised learning and reinforcement learning scenarios.
    \item Except for the theoretical convergence analysis of MAML, the convergence theory of other metric-based, model-based, and gradient-based meta learning algorithms is not derived. Therefore, detailed convergence analysis of meta learning is still need to be investigated.
\end{itemize}

\begin{figure}[!h]
    \centering
    \includegraphics[width=3.0 in]{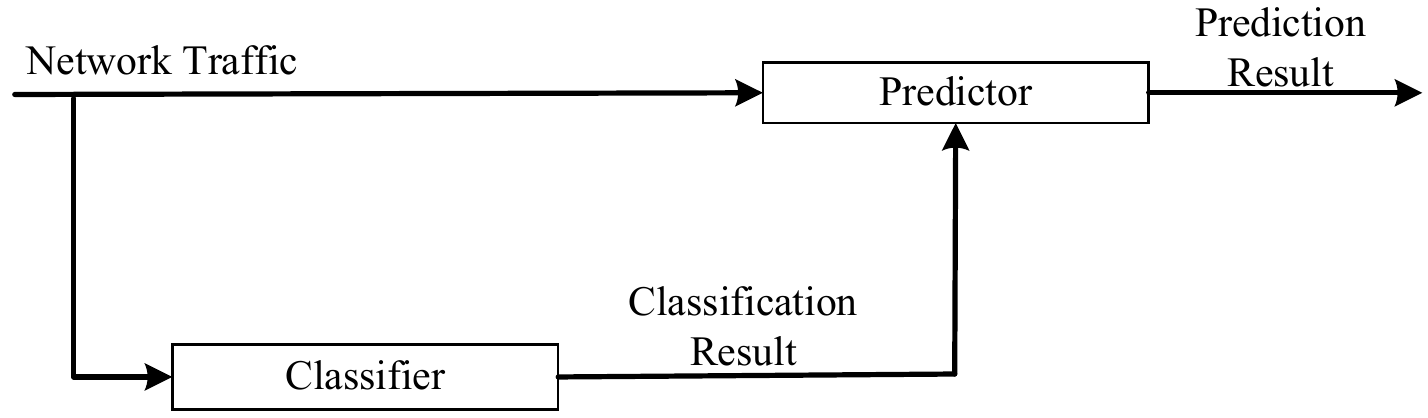}
    \caption{Feed forward structure for traffic prediction.}
    \label{basic_modules}
\end{figure}

\begin{figure}[!h]
    \centering
    \includegraphics[width=2.8 in]{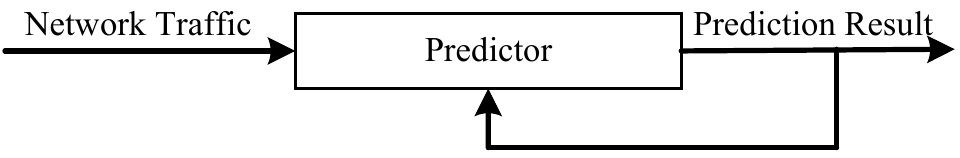}
    \caption{Feed back structure for traffic prediction.}
    \label{basic_modules}
\end{figure}

\section{Meta Learning in Wireless Communications}
Due to the advantages of gradient-based meta learning, such as good generalization performance on new tasks and the model being easy to fine-tune, it has been widely used for optimization problems in wireless networks, such as traffic prediction \cite{Moayyedi}, transmission rate maximization \cite{Jingyuan,yi_yuan}, and multiple-input and multiple-output (MIMO) detectors \cite{Jingzhang}. 

\subsection{Traffic Prediction}
One of the traditional methods for network traffic prediction is a feed-forward predictor, which consists of a traffic classifier trained to recognize specific types of traffic, such as videos, web traffic, file downloading, and a predictor that takes the network traffic and classification results as inputs, as shown in Fig. 19. However, it requires a large amount of labeled datasets to train each traffic classifier, which leads to high computation complexity. To address this issue, authors in \cite{Moayyedi} proposed a feedback traffic prediction architecture based on a meta-learning scheme. The feed-back architecture is presented in Fig. 20, where the predictor is selected based on observed prediction accuracy by DRL, rather than the traffic class. The reason for using meta learning is that, it is recently employed in robust adversarial learning, which can exploit models by taking advantage of obtainable model information and using it to create malicious attacks \cite{X_lee,A_Havens,K_Frans}. According to Fig. 21, the meta-learning scheme used in the predictor consists of a master policy and a set of sub-policies. The master policy is responsible for selecting which sub-policy is used for prediction during the next prediction interval. The meta-learning scheme in \cite{Moayyedi} allows for the updating of sub-predictors in real-time, so that the sub-predictors have the ability to adapt to variations in traffic patterns over time. The prediction accuracy of meta learning scheme significantly outperforms that of any single predictor. However, when a new sub-predictor is added to sub-predictors, the master policy needs to be retrained, which leads to high computational costs.

\begin{figure}[!h]
    \centering
    \includegraphics[width=3.5 in]{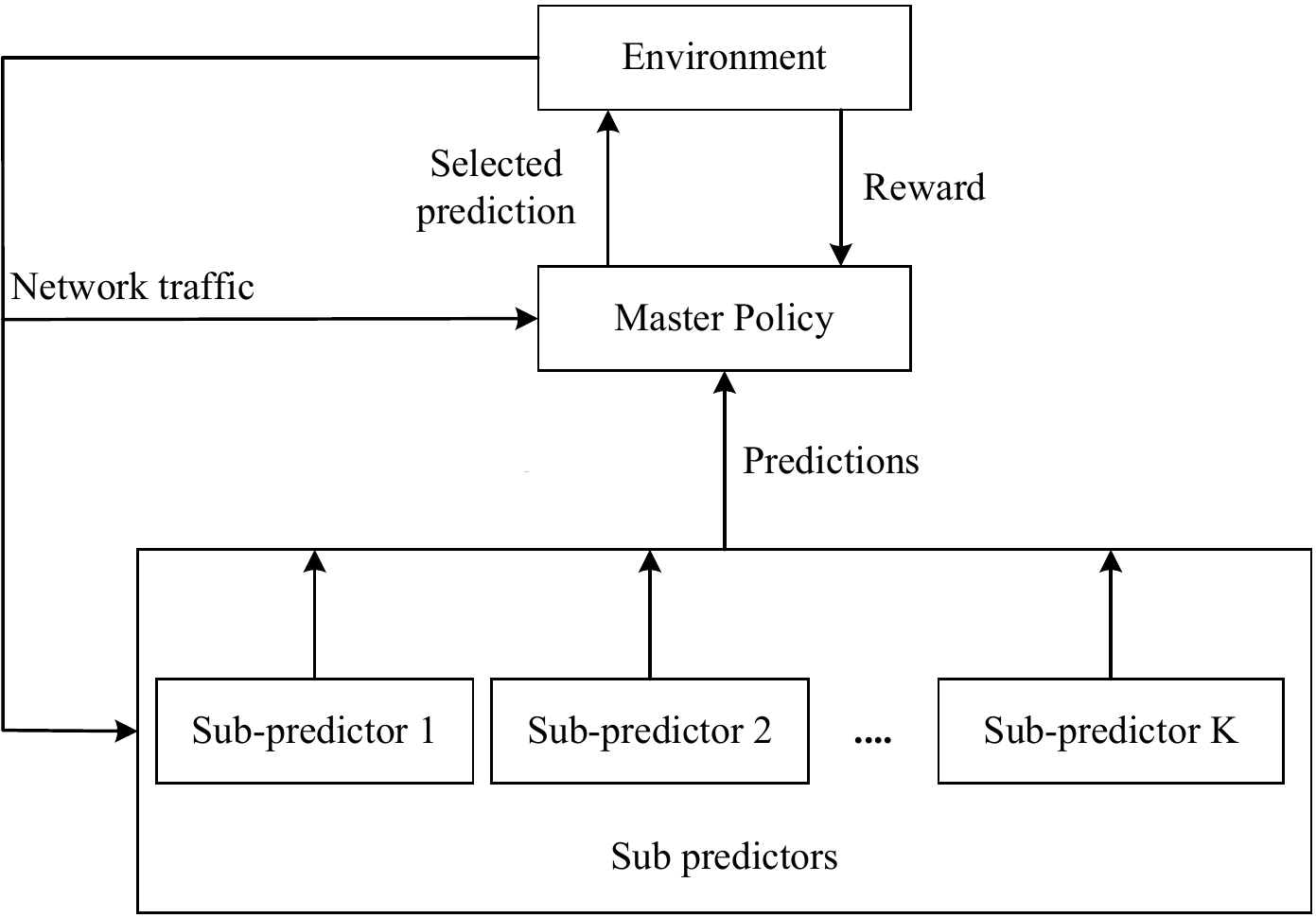}
    \caption{Meta learning and DRL for traffic prediction.}
    \label{basic_modules}
\end{figure}

\subsection{Transmission Rate Maximization}
The design of beamforming vectors that maximize the weighted sum rate (WSR) is an NP-hard problem and the iterative weighted minimum mean square error (WMMSE) is the most widely used technique to achieve optimal beamforming vectors. Although authors in \cite{H_sun,W_Xia,Pellaco,Chowdhury} considered deep learning or graph neural networks (GNNs) to estimate the beamforming vectors, the sum rate performance of these methods was not higher than that of the WMMSE algorithm. An alternative method is to deploy meta-learning algorithms to solve the problem. Authors in \cite{Jingyuan} proposed a meta-learning-aided beamformer (MLBM) algorithm to solve the WSR maximization problem. The objective of MLBM is to minimize the global loss function $F(\textbf{u}, \bm{w}, \textbf{V})$, where $\textbf{u}$, $\bm{w}$, and $\textbf{V}$ are receiver gain vector, positive user weight vector, and transmit beamforming vector of all devices, respectively. Particularly, $F(\textbf{u}, \bm{w}, \textbf{V})$ is divided into three sub-problems $f(\textbf{u})$, $f(\bm{w})$, and $f(\textbf{v})$, and authors refer to the minimization of $F(\textbf{u}, \bm{w}, \textbf{V})$ as minimization of these three sub-problems $f(\textbf{u})$, $f(\bm{w})$, and $f(\textbf{v})$. A meta-learner neural network is deployed to treat these three sub-problems as three tasks and sequentially update $\textbf{u}$, $\bm{w}$, and $\textbf{V}$ of these three sub-problems, which is similar to the inner loop of task-specific adaptation of gradient-based meta-learning. Then, the updated $\textbf{u}$, $\bm{w}$, and $\textbf{V}$ are used to calculate global loss function $F(\textbf{u}, \bm{w}, \textbf{V})$, and update the learning weights of the meta learner, which is similar to the outer loop of meta initialization training of gradient-based meta-learning. Through iteratively updating $\textbf{u}$, $\bm{w}$, $\textbf{V}$, and meta-learner weights, MLBM can achieve a higher transmission rate than that of the WMMSE algorithm particularly in the high SNR regime, and achieves a similar performance when SNR is small. In addition, authors in \cite{yi_yuan} compared meta learning and transfer learning in beamforming design, and verified that meta learning was able to provide a higher transmission rate compared to that of transfer learning.

\subsection{MIMO Detectors}
Deep neural networks (DNNs) have the potential for efficiently balancing the bit-error rate minimization and computation complexity of MIMO detectors \cite{Ito,HHe,Jzhang,Sahin}. Unfortunately, the existing DNN-based MIMO detectors are difficult to be deployed in practical systems due to their slow convergence speed and low robustness in new environments. To deal with this issue, authors in \cite{Jingzhang} proposed meta-learning-based MIMO detectors, which could be used in channel-sensitive environments. In particular, an expectation propagation (EP) for signal detection is unfolded as EPNet, and damping factors are set as trainable parameters to adapt to new channels \cite{Minka,Seeger,Cespedes,Cespedes1}. Damping factors are relevant to channel statistics. To train the damping factors, a large amount of labeled channel state information (CSI) is required. However, in a dynamic real-time wireless system, it is impractical to obtain enough CSI to train the damping factors. Thus, meta learning is deployed to update the damping factors efficiently by using a small training set so that they can quickly adapt to new environments.

\subsection{Summary and Lessons Learned}
In this section, we have reviewed three research directions of meta learning over wireless networks. We summarize the approaches along with references. From this review, we gather the following lessons learned:

\begin{itemize}
    \item Meta learning is effective for wireless networks with multiple tasks or sub-problems, and achieves better learning and network transmission performance, such as higher learning accuracy and transmission rate, compared with conventional optimization methods.
    \item For the research works we have discussed in this section, the proposed scheme in each work can quickly adapt to new environments with the help of meta learning. However, the computation and memory costs of meta learning are much higher than that of traditional optimization approaches.
    \item Although meta learning has been widely applied in the design of signal processing and network management  \cite{Moayyedi,Jingyuan,yi_yuan,Jingzhang,9290055,9593131,9457160}, its applications in wireless networks still face several challenges. First, multiple meta learning models can be generated at the base stations. Sophisticated model selection schemes are still unknown to adapt to different learning tasks. Second, the meta learning models are transmitted to devices by using extra radio resources, which has a heavy burden on wireless networks \cite{9814634}. Therefore, it is a dilemma to balance the learning performance and the communication costs of meta learning.
    \item To the best of the authors' knowledge, no research works focus on how wireless factors affect meta learning. Therefore, we directly introduce how meta learning is used to solve the proposed optimization problems in wireless networks. Also, the convergence analysis of meta learning over wireless networks is not investigated.
\end{itemize}

\section{Federated Meta Learning}
Gradient-based meta learning algorithms, such as MAML, are well known for their rapid adaptation and good generalization to multiple learning tasks, which makes them particularly suitable for federated settings where the decentralized training data is non-IID and highly personalized \cite{Finn,Taesup,Vuorio,Fallah1}. In this section, federated meta learning methodologies and their applications over wireless networks are discussed.

\subsection{FedMeta Methodologies}
In this subsection, we introduce three main research directions of FedMeta methodologies, including MAML/Meta-SGD-based FedMeta, Collaborative FedMeta, and ADMM-FedMeta. The introduced FedMeta methodologies are fundamental FedMeta algorithms from the CS community and the transmission between servers and devices is error-free without considering any wireless factors.

\begin{figure}[!h]
    \centering
    \includegraphics[width=3.5 in]{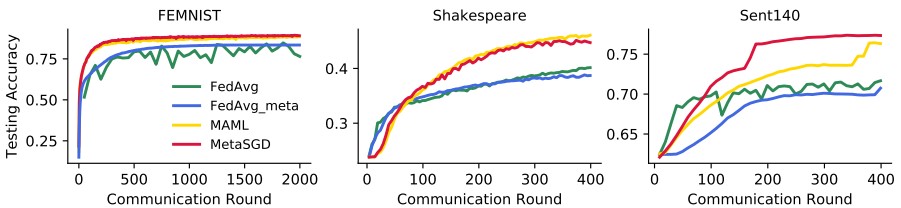}
    \caption{Comparison of the convergence speed of FedMeta and FedAvg from \cite{feichen}.}
    \label{basic_modules}
\end{figure}

\subsubsection{MAML/Meta-SGD-based FedMeta} The federated meta learning framework was first proposed in \cite{feichen}, where authors integrated MAML and Meta-SGD \cite{zhenguo} into FL. The objective of the algorithm is to collaboratively meta-train an algorithm using datasets from distributed devices. For model aggregation in the server, as shown in \textbf{Algorithm 7}, the server maintains the initialization parameters $\bm{\theta}$ and $\bm{\alpha}$, updates them through the testing loss from the selected devices, and transmits them to the selected devices. For the local model training and testing in devices, as shown in $\textbf{Algorithm}$ 8, first, the $k$th device trains the learning weights $\bm{\theta}$ obtained from the server using its support dataset $D_{S}^{k}$. Second, the $k$th device tests the trained learning model, calculates the testing loss $\mathcal{L}_{D_{Q}^{u}}(\bm{\theta})$ based on its query set $D_{Q}^{u}$, and transmits $\mathcal{L}_{D_{Q}^{u}}(\bm{\theta})$ to the server. For Meta-SGD, the vector $\bm{\alpha}$ is also delivered to the server as part of the algorithm parameters which are used for parameter updating. The detailed FedMeta with MAML and Meta-SGD at the server, and model training of MAML or Meta-SGD at the device are introduced in \textbf{Algorithm 7} and \textbf{Algorithm 8}, respectively. According to \cite{feichen}, the comparison of the convergence speed of FedMeta and FedAvg is shown in Fig. 22. It is observed that FedMeta provides a faster convergence speed and higher learning accuracy.

\subsubsection{Collaborative FedMeta} Authors in \cite{Senlin} proposed a platform-based collaborative learning framework, where a model was first trained in a set of edge nodes by FedMeta, and then it was rapidly adapted to learn a new task at the target edge node with a few data samples. This can deal with the constrained computing resources and limited local data issues of each edge node. The FedMeta algorithm used in \cite{Senlin} is the same as that used in \cite{feichen}. However, according to Fig. 23, the main differences are that: 1) Authors in \cite{Senlin} deployed a set of source edge nodes only with a support dataset to train the local models, rather than authors in \cite{feichen} assuming that each device had both support and query datasets to train and test the local model; 2) Authors in \cite{Senlin} assumed that each source edge node focused on only one task to train the local model, rather than authors in \cite{feichen} assuming that each device had multiple tasks; 3) Authors in \cite{Senlin} used a platform to aggregate local models from all source edge nodes and transmitted the aggregated model to the target edge node for new task adaptation, rather than authors in \cite{feichen} assuming that each device was able to adapt to new tasks. With model training at multiple edge nodes, FedMeta can adapt to multiple tasks in parallel. However, for cases with a large number of tasks, delivering multiple learning models simultaneously can result in high transmission latency.

\begin{algorithm}[t]
\begin{algorithmic}[1]
\caption{FedMeta with MAML and Meta-SGD at the server}
\STATE Initialize step size hyperparameters $\alpha$ and $\beta$, randomly initialize learning weights $\bm{\theta}$.
\FOR{each time slot t=1,2,...}
    \STATE Sample $K$ devices, and distribute $\bm{\theta}$ for MAML or ($\bm{\theta}, \bm{\alpha}$) for Meta-SGD to these $K$ devices.
    \FOR{the $k$th device in $K$ devices}
        \STATE Obtain testing loss $\nabla_{\bm{\theta}}\mathcal{L}_{D_{Q}^{k}}(\bm{\theta}_{k})$ from the model training of the MAML.
        \STATE Obtain testing loss $\nabla_{(\bm{\theta},\bm{\alpha})}\mathcal{L}_{D_{Q}^{k}}(\bm{\theta}_{k})$ from the model training of the Meta-SGD.
    \ENDFOR
    \STATE Update $\bm{\theta} = \bm{\theta} - \frac{\beta}{K}\sum_{k=1}^{K}\nabla_{\bm{\theta}}\mathcal{L}_{D_{Q}^{k}}(\bm{\theta}_{k})$ for the MAML.
    \STATE Update $(\bm{\theta}, \bm{\alpha}) = (\bm{\theta}, \bm{\alpha}) - \frac{\beta}{K}\sum_{k=1}^{K}\nabla_{(\bm{\theta},\bm{\alpha})}\mathcal{L}_{D_{Q}^{k}}(\bm{\theta}_{k})$ for the MetaSGD.
\ENDFOR
\end{algorithmic}
\end{algorithm}

\begin{algorithm}[t]
\begin{algorithmic}[1]
\caption{Model Training for MAML and MetaSGD}
\STATE Sample support set $D_S^k$ and query set $D_Q^k$ of the $k$th device.
\STATE $\mathcal{L}_{D_{S}^{k}}(\bm{\theta}) = \frac{1}{|D_{S}^{k}|}\sum_{(x,y)\in D_{S}^{k}}l(f_{\bm{\theta}}(x),y)$.
\STATE $\bm{\theta}_k = \bm{\theta} - \alpha\nabla\mathcal{L}_{D_S^k}(\bm{\theta})$ for the MAML and $\bm{\theta}_k = \bm{\theta} - \bm{\alpha} \circ \nabla\mathcal{L}_{D_S^k}(\bm{\theta})$ for the MetaSGD.
\STATE $\mathcal{L}_{D_{Q}^{k}}(\bm{\theta}_k) = \frac{1}{|D_{Q}^{k}|}\sum_{(x^{'},y^{'})\in D_{Q}^{k}}l(f_{\bm{\theta}_k}(x^{'}),y^{'})$.
\STATE Transmit $\nabla_{\bm{\theta}}\mathcal{L}_{D_{Q}^{k}}(\bm{\theta}_{k})$ and $\nabla_{(\bm{\theta},\bm{\alpha})}\mathcal{L}_{D_{Q}^{k}}(\bm{\theta}_{k})$ to the server.
\end{algorithmic}
\end{algorithm}

\begin{figure}[!h]
    \centering
    \includegraphics[width=3.5 in]{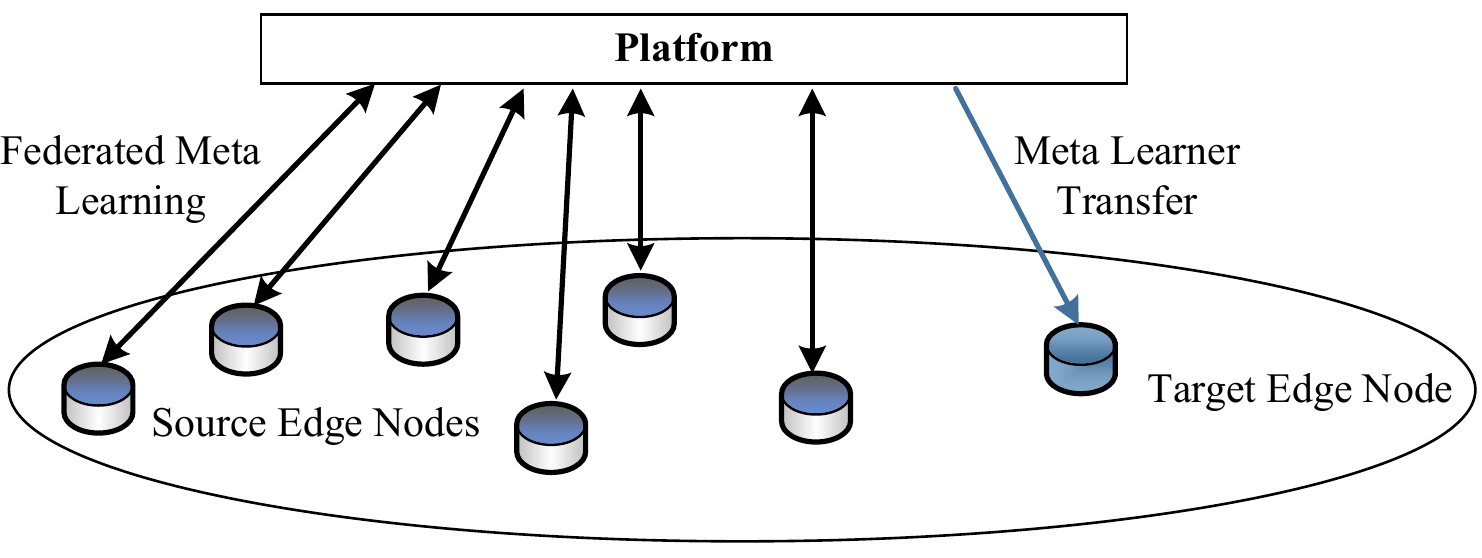}
    \caption{Platform-based collaborative federated meta learning framework.}
    \label{basic_modules}
\end{figure}

\subsubsection{ADMM-FedMeta} Based on \cite{Senlin}, authors in \cite{shengyue1} proposed an ADMM-based algorithm, called ADMM-FedMeta, to decompose the initial optimization problem into several sub-problems which can be solved in parallel across edge nodes and the platform. First, authors still applied the platform-based FedMeta architecture in \cite{Senlin} to enable edge nodes to collaboratively learn a meta-model with the knowledge transfer of previous tasks. Then, the FedMeta problem is defined as a regularized optimization problem, where the previous knowledge is extracted as regularization, and the optimization problem is denoted as
\begin{align}
\min_{\bm{\theta}_i, \bm{\theta}} &\sum_{k\in \mathcal{I}}\frac{D_k}{\sum_{k\in\mathcal{I}}D_k}\mathcal{L}_k(\phi_k(\bm{\theta}_k), \mathcal{D}_k^q) + \lambda D_{h}(\bm{\theta}, \bm{\theta}_{p}),\\
\text{s.t.} &~~~~ \bm{\theta}_k - \bm{\theta} = \bm{0}, k\in\mathcal{I}, \nonumber
\end{align}
where $\phi_k(\bm{\theta}_k)$ is written as (104), $D_{h}(\bm{\theta}, \bm{\theta}_{p})$ is a regularization parameter that can extract the valuable knowledge from the prior model to facilitate fast edge training and alleviate catastrophic forgetting \cite{Parisi}. In (123), $\bm{\theta}_{p}$ is the prior model weights, $\lambda$ is a penalty parameter that is used to balance the trade-off between the loss and regularization, $\mathcal{I}$ is the set of edge nodes, $\mathcal{D}_k$ is the dataset of the $k$th ($k\in\mathcal{I}$) edge node, $D_k$ is the number of data samples in $\mathcal{D}_k$, and $\mathcal{D}_k$ is divided into two disjoint datasets, i.e., the support set $\mathcal{D}_k^{s}$ and query set $\mathcal{D}_k^{q}$. Through penalizing variations in the model via regularization, the learned model from (123) is close to the prior model for enabling collaborative edge learning without forgetting prior knowledge, so that the learned meta model can adapt to different tasks. To solve the optimization problem in (123), the augmented Lagrangian function is deployed, which is written as
\begin{align}
    &\mathcal{L}(\{\bm{\theta}_k, \bm{w}_k\}, \bm{\theta}) = \sum_{k\in \mathcal{I}}(\frac{D_k}{\sum_{k\in\mathcal{I}}D_k}\mathcal{L}_k(\phi_k(\bm{\theta}_k), \mathcal{D}_k^q) \nonumber\\&+ <\bm{w}_k, \bm{\theta}_k - \bm{\theta}> + \frac{\rho_k}{2}\|\bm{\theta}_k - \bm{\theta}\|^2) + \lambda D_h(\bm{\theta}_k, \bm{\theta}),
\end{align}
where $\bm{w}_k$ is a dual variable and $\rho_k > 0$ is a penalty parameter. To optimize $\bm{\theta}_k$ $\bm{\theta}$, and $\bm{w}_k$, ADMM method is applied \cite{Boyd,Mhong,Magnusson,fwang,ywang,fwang1,Barber,bjiang,Lanza,Mukkamala}. The traditional ADMM decomposes the optimization in (110) into a set of sub-problems that can be solved in parallel, which means that calculating $D_h(\bm{\theta}_k, \bm{\theta})$ and $\mathcal{L}_k(\phi_k(\bm{\theta}_k), \mathcal{D}_k^q)$ separately. Thus, the vector $\bm{\theta}_k$, $\bm{\theta}$, and $\bm{w}_k$ are updated alternatively as follows
\begin{equation}
    \bm{\theta}^{t+1} = \arg\min_{\bm{\theta}}\mathcal{L}(\{\bm{\theta}_k^t, \bm{w}_k^t\}, \bm{\theta}),
\end{equation}
\begin{equation}
    \bm{\theta}_k^{t+1} = \arg\min_{\bm{\theta}_k}\mathcal{L}_k(\bm{\theta}_k, \bm{w}_k^t, \bm{\theta}^{t+1}),
\end{equation}
and
\begin{equation}
    \bm{w}_k^{t+1} = \bm{w}_k^{t} + \rho_k(\bm{\theta}_k^{t+1} - \bm{\theta}^{t+1}).
\end{equation}
In (125), (126), and (127), the platform and each device select the weights that can achieve the minimum loss, and update the dual variable by the difference of local and global weights. Based on (125), (126), and (127), the updating strategy is 1) updating $\bm{\theta}$ at the platform and 2) updating $\{\bm{\theta}_k, \bm{w}_k\}$. The advantage of using ADMM-FedMeta is that the decoupled sub-problems can be allocated to multiple edge nodes to solve simultaneously, which helps to alleviate the local computational costs and improve the computation efficiency.  According to \cite{shengyue1}, the comparison of the convergence speed of ADMM-FedMeta and FedAvg is shown in Fig. 24. It is observed that ADMM-FedMeta achieves a much higher convergence speed than that of FedAvg.

\begin{figure}[!h]
    \centering
    \includegraphics[width=3.5 in]{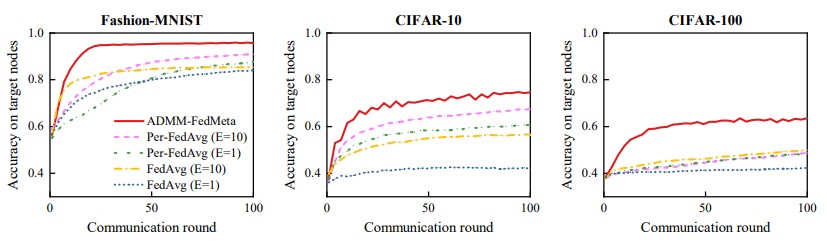}
    \caption{Comparison of the convergence speed of ADMM-FedMeta and FedAvg from \cite{shengyue1}.}
    \label{basic_modules}
\end{figure}

\subsection{FedMeta in Wireless Networks}
FedMeta integrates the advantages of FL and meta learning, which enables local model sharing without privacy issues and fast adaptation to new tasks. Through learning an initial shared model, devices can quickly adapt the learned model to their local datasets via one or a few gradient descent steps. Despite its advantages, FedMeta still has several challenges: First, the number of participating devices can be enormous. When devices are randomly selected, it can lead to a low convergence speed \cite{shengyue1}. Second, the convergence performance of FedMeta in a wireless network is highly related to its latency, which includes computation latency, determined by the size of local datasets and CPU types of devices, and transmission latency, determined by channel gains, interference, and transmission power \cite{Dinh}. If these factors are not optimized, high latency can result in unexpected training delay and communication inefficiency \cite{Kairouz}. In this subsection, we introduce device selection and energy efficiency of FedMeta over wireless networks, based on the fundamental FedMeta methodologies.

\subsubsection{Device Selection} To deal with high training delay and communication inefficiency, authors in \cite{shengyue} developed a FedMeta with a non-uniform device selection scheme to accelerate the convergence, and rigorously analyzed the contribution of each device to the global loss reduction in each time slot. Then, a resource allocation problem integrating FedMeta into multi-access wireless systems is proposed to jointly improve the convergence rate and minimize the latency along with energy cost. The learning structure of FedMeta proposed in \cite{shengyue} is the same as that in \cite{feichen}. The device selection problem in the $t$th time slot is formulated as
\begin{align}
\max_{z_k}& \sum_{k\in \mathcal{N}}z_ku_{k}^t\\
\text{s.t.}&\sum_{k\in \mathcal{N}}z_k = n_k \nonumber\\
&z_k\in\{0,1\}.\nonumber
\end{align}
In (128), $z_k$ is a binary variable, if $z_k = 1$, the $k$th device is selected, otherwise, not, and $\mathcal{N}$ is the set of devices. $u_{k}^t$ is the contribution of the $k$th device to the convergence in the $t$th time slot, defined as
\begin{equation}
    u_{k}^t = \sum_{i = 0}^{\tau - 1}\|\nabla F_{k}(\bm{\theta}_{k}^{i,t})\|^{2} - 2(\lambda_1 + \frac{\lambda_2}{\sqrt{D_k}})\|\nabla F_{i}(\bm{\theta}_{k}^{i,t})\|,
\end{equation}
where $F_{k}(\bm{\theta}_{k}^{i,t})$ is a meta function of the $k$th device, $\tau$ is the number of iterations of gradient descent, $D_k$ is the number of data samples in the $k$th device, and $\lambda_1$ and $\lambda_2$ are position constants. The detailed proof of $u_{k}^t$ is presented in Appendix J of the technical report \cite{Syue}. To apply FedMeta over wireless networks, the authors propose a resource allocation problem, capturing the trade-offs among the convergence, computation and communication latency, and energy consumption. Then, the optimization problem is decomposed into two sub-problems. The first sub-problem aims at controlling the CPU-cycle frequencies for devices to minimize energy consumption and computation latency. The second sub-problem controls transmission power and resource block allocation to maximize the convergence speed while minimizing the transmission cost and latency. Both of the sub-problems are solved by KKT conditions. 

\subsubsection{Energy Efficiency} In FedMeta, each task is owned by a device, and each updating iteration needs to communicate with the server. In each iteration, each device updates its local model and transmits it to the server, where the meta model is updated in a global step and then the updated meta model is feedback to all devices. Since these procedures involve local computation and communication energy consumption, it is important to minimize the computation costs and save energy for communication, especially for a device with limited computation capability and energy. To minimize energy and computation costs when performing FedMeta, authors in \cite{Elgabli} considered an energy-efficient FedMeta framework, where a meta-backward algorithm was proposed, to learn a meta model with low computation and communication energy consumption. In the backward manner, in the $k$th step, $\bm{\theta}_{i}^{k}$ is computed as
\begin{equation}
    \bm{\theta}_{i}^{k} = \bm{\theta}_{i}^{k+1} + \alpha\nabla_{\bm{\theta}_{i}^{k}}\mathcal{L}(\bm{\theta}_{i}, \mathcal{D}_i).
\end{equation}
However, when computing in a backward manner, the term $\nabla_{\bm{\theta}_{i}^{k}}\mathcal{L}(\bm{\theta}_{i}, \mathcal{D}_i)$ cannot be calculated. To address this issue, $\nabla_{\bm{\theta}_{i}^{k}}\mathcal{L}(\bm{\theta}_{i}, \mathcal{D}_i)$ can be replaced by $\nabla_{\bm{\theta}_{i}^{k+1}}\mathcal{L}(\bm{\theta}_{i}, \mathcal{D}_i)$, because they are close to each other under smoothness assumption.
In the $k$th backward step, to find the optimal $\bm{\theta}_{i}^{k}$, the optimization problem is fomulated as
\begin{align}
    &\min_{\{\bm{\theta}_{i}\}_{i=1}^{N}}\sum_{i=1}^{N}\left(\mathcal{L}(\bm{\theta}_{i}) - \mathcal{L}(\bm{\theta}_{i}^{k,0})\right)^2, \\
    &\text{s.t.}~~\|\bm{\theta}_{i} - \bm{\psi}^{k+1}\|^2 \leq \delta_k, \nonumber
\end{align}
where $\bm{\psi}^{k+1} = \frac{1}{N}\sum_{i=1}^{N}\bm{\theta}_{i}^{k+1}$, $N$ is the total number of tasks, $\delta_k\rightarrow 0$, and
\begin{equation}
    \bm{\theta}_{i}^{k,0} = \bm{\theta}_{i}^{k+1} + \alpha\nabla_{\bm{\theta}_{i}^{k+1}}\mathcal{L}(\bm{\theta}_{i}^{k+1}, \mathcal{D}_i).
\end{equation}
Obviously, based on the constraint in (131), we need to find the optimal $\bm{\theta}_{i}^{k}$ which is close to the average weight $\bm{\psi}$ among all tasks. To do so, a projection gradient descent (PGD) is introduced to solve the problem in (131). Note that PGD is a standard way to solve constrained optimization problems \cite{Wainwright,Bahmani}. The projection problem for the $i$th task is written as
\begin{align}
    &\min_{\bm{\theta}_{i}^{k}}~~\frac{1}{2}\|\bm{\theta}_{i}^{k} - \bm{\theta}_{i}^{k,0}\|^2, \\
    &\text{s.t.}~~\|\bm{\theta}_{i}^{k} - \bm{\psi}^{k+1}\|^2 \leq \delta_k. \nonumber
\end{align}
The Lagrangian function of (133) is denoted as
\begin{equation}
L(\bm{\theta}_i, \bm{\psi}, \mu) = \frac{1}{2}\|\bm{\theta}_{i}^{k} - \bm{\theta}_{i}^{k,0}\|^2 + \mu(\|\bm{\theta}_{i}^{k} - \bm{\psi}^{k+1}\|^2- \delta_k).
\end{equation}
Using KKT conditions, we can solve the problem in (133). The proposed meta-backward algorithm is computationally efficient, as it has a closed-form solution calculated by KKT conditions in each iteration. 

\subsection{Convergence Analysis of FedMeta}
For FedMeta methodologies, authors usually derive the upper bound of $\{\mathcal{L}(\bm{\theta}^{T}) - \mathcal{L}(\bm{\theta}^{*})\}$ and $\|\nabla f(\bm{\theta}^{*})\|$ of FedMeta and ADMM-FedMeta for convergence analysis, respectively. Except from the basic assumptions 1 - 4 introduced in Section III, another assumption is introduced in the following.

$\textbf{Assumption 6.}$ Each loss functions $\mathcal{L}_{n}(\bm{\theta})$ is $\mu$-strongly convex for any $\bm{\theta}$ and $\bm{\theta}^{'}$:
\begin{equation}
    \langle\nabla \mathcal{L}_{n}(\bm{\theta}) - \nabla \mathcal{L}_{n}(\bm{\theta}^{'}), \bm{\theta} - \bm{\theta}^{'}\rangle \geq \mu\|\bm{\theta} - \bm{\theta}^{'}\|^2,
\end{equation}
where $\mu$ is a positive constant.
The upper bound of $\{\mathcal{L}(\bm{\theta}^{T}) - \mathcal{L}(\bm{\theta}^{*})\}$ of FedMeta in \cite{Senlin} is derived as
\begin{equation}
    \mathcal{L}(\bm{\theta}^{T}) - \mathcal{L}(\bm{\theta}^{*}) \leq \xi^{T}[\mathcal{L}(\bm{\theta}) - \mathcal{L}(\bm{\theta}^{*})] + \frac{G(1 - \alpha\mu)}{1- \xi^{T_0}}h(T_0),
\end{equation}
where $\alpha$ is learning rate, $G$ is defined in assumption 3, $\xi$ is a constant determined by learning rate, $\mu$, and $L$ defined in assumption 2, $T_0$ is the duration of one local update step, and $T = NT_0$ is a fixed duration given the number of local update steps $N$. The term $\frac{G(1 - \alpha\mu)}{1- \xi^{T_0}}h(T_0)$ captures the error introduced by both task dissimilarity and multiple local updates through the function $h(T_0)$. In (136), $h(T_0)$ indicates how the task similarity and $T_0$ impact the convergence performance, namely, given a fixed duration $T$, the convergence error decreases with the task similarity while
increasing with the number of local update steps when $T_0$ is large. Therefore, the platform-based FedMeta is able to balance between the platform-edge communication cost and the local computation cost by controlling the number of local update steps per communication round, depending on the task similarity across the edge nodes. Furthermore, for the convergence analysis of ADMM-FedMeta in \cite{shengyue1}, it is proved that $\bm{\theta}^{t}$ has at least one limit point and each limit point $\bm{\theta}^{*}$ has a stationary solution, namely, $\|\nabla f(\bm{\theta}^{*})\| = 0$.

Note that there are no research works obtained the theoretical convergence analysis of FedMeta with device selection, resource allocation, and energy consumption over wireless networks.

\subsection{Summary and Lessons Learned}
In this section, we have reviewed three research directions of FedMeta methodologies and two research directions of FedMeta over wireless networks. We summarize the approaches along with references. From this review, we gather the following lessons learned:

\begin{itemize}
    \item FedMeta is the integration of FL with meta learning, which not only guarantees user privacy but also quickly adapts to multiple tasks. However, for scenarios with plenty of tasks, optimizing a meta-learner for each task and frequently delivering local and global models between servers and devices lead to high computation and communication latency. Although we mainly focus on FedMeta with MAML and Meta-SGD in this section, other meta learning algorithms, such as Reptile, BMAML, LLAMA, LEO, iMAML, Online MAML, can still be used in FedMeta.
    \item Given that FedMeta is a specialization of FL, FedMeta has all the challenges that exist in FL, plus additional challenges. The key challenge is how to effectively train models across multiple devices. Architecture challenge includes whether it is possible to use a peer-to-peer architecture. Another challenge would arise if the system should not only focus on the globally best algorithm for a task (an entire dataset) but if per-instance algorithm selection should be learned. This would make the whole system even more complex.  
    \item For the research works we have discussed in this section, a large number of devices participating in FedMeta results in lower convergence rate, higher communication latency, higher energy costs, and system heterogeneity than that of FL and meta learning. Device selection and energy-efficient FedMeta cannot achieve significant effectiveness. Therefore,  the trade-off between the local model
    updating and global model aggregation in FedMeta to minimize the convergence time and energy cost from a long-term perspective needs to be investigated. Also,  how to characterize the convergence properties and communication complexity of FedMeta over wireless networks requires further research.  
\end{itemize}

\section{Implementation Platforms}
In this section, implementation platforms of FL, meta learning, and FedMeta are introduced.

\subsection{FL Platforms}
Based on the introduced research works of FL methodologies and their applications over wireless networks, several platforms for FL are constructed, which are PySyft, TensorFlow Federated (TFF), Federated AI Technology Enabler (FATE), Tensor/IO, Functional FL in Erlang (FFL-ERL), CrypTen, and LEAF.

\begin{itemize}
    \item \textbf{PySyft:} PySyft is an open-source multi-language library enabling secure and private machine learning by wrapping and extending popular deep learning frameworks such as PyTorch in a transparent, lightweight, and user-friendly manner. Its aim is to both help popularize privacy-preserving techniques in machine learning by making them as accessible as possible by Python bindings and common tools familiar to researchers and data scientists, as well as to be extensible such that new FL, Multi-Party Computation, or Differential Privacy methods can be flexibly and simply implemented and integrated \cite{Ziller2021PySyftAL}. PySyft decouples private data from model training, using FL within PyTorch.
    \item \textbf{TFF:} TensorFlow Federated (TFF) is an open-source framework for machine learning and other computations on decentralized data. TFF has been developed to facilitate open research and experimentation with FL. TFF enables developers to use the included federated learning algorithms with their models and data, as well as to experiment with novel algorithms. The building blocks provided by TFF can also be used to implement non-learning computations, such as aggregated analytics over decentralized data \cite{TFF}.
    \item \textbf{FATE:} FATE is an open-source project initiated by Webank’s AI Department to provide a secure computing framework to support the federated AI ecosystem. It implements multiple secure computation protocols to enable big data collaboration with data protection regulation compliance \cite{FATE}. Furthermore, FATE is able to support various FL architectures and ML algorithms.
    \item \textbf{Tensor/IO:} Tensor/IO is a platform that brings TFF to mobile devices such as iOS and Android \cite{tensor}. Although this platform cannot implement any ML algorithms, the platform cooperates with TFF to ease the implementation and deployment of ML algorithms on mobile phones.
    \item \textbf{FEL-ERL:}  The functional programming language Erlang is well-suited
    for concurrent and distributed applications, which is suitable for establishing real-time systems. FEL-ERL was proposed by Gregor Ulm,  Emil Gustavsson, and Mats Jirstran in \cite{fel-erl}. They evaluated FEL-ERL in two scenarios: one in which the entire system has been written in Erlang, and another in which Erlang is relegated to coordinating device processes that rely on performing numerical computations in the programming language C. The authors found that Erlang incurs a performance penalty, but for certain use cases this may not be detrimental, considering the trade-off between speed of development (Erlang) versus performance (C).
    \item \textbf{CrypTen: } CrypTen is a new framework built on PyTorch to facilitate research in FL \cite{knott2021crypten}. There are a few benefits to using this platform. One benefit is that CrypTen enables machine learning researchers, who may not be cryptography experts, to easily experiment with FL. Another benefit is that the platform is made with real-world challenges in mind. Therefore, it has the potential to be applicable to large number of real-world applications.
    \item \textbf{LEAF: } LEAF is a modular framework for FL, multi-task learning, meta learning, and FedMeta. LEAF was proposed in \cite{leaf} which allowed researchers to reason about new proposed solutions under more realistic assumptions than previous benchmarks. LEAF is kept updating with new datasets, metrics, and open-source solutions to foster informed and grounded progress in ML field.

\end{itemize}

\subsection{Meta Learning and FedMeta Platforms}
Based on the introduced research works of meta learning/FedMeta methodologies and their applications over wireless networks, LEAF platform has been used for these two learning algorithms' implementation, which has already been introduced in the previous section. Recently, a novel meta learning platform called Awesome-META+ is created.

\begin{itemize}
    \item $\textbf{Awesome-META+: }$ Awesome-META+ provides a comprehensive and reliable meta learning framework code that can adapt to multiple domains and
    improve academic research efficiency. Furthermore, it provides a convenient and simple model deployment solution to lower the threshold and promote the development of meta-learning and its transfer fields. In addition, it provides a comprehensive and complete information summary and learning platform for the meta learning field to stimulate the vitality of the meta-learning community \cite{awesome}.
\end{itemize}

\section{Real-World Applications}
In this section, real-world applications for FL, meta learning, and FedMeta in real scenarios are introduced.

\subsection{FL}
FL is used in Google keyboard, intelligent medical diagnosis system, and autonomous driving vehicles.

\subsubsection{Google Keyboard} Google started a project in 2016 to implement FL among mobile devices to improve the learning accuracy of keyboard input prediction, while guaranteeing the privacy of users simultaneously \cite{keyboard}. Through developing language models, the recommendation system is also promoted. Also, it can be extended to other recommendation applications by integrating with FL. When mobile devices send requests, the corresponding suggestions are quickly provided by learning models.

\subsubsection{Intelligent Medical Diagnosis System} Different hospitals own the electronic health records of different patient populations and these records are difficult to share across hospitals because of the protection of patient privacy. This creates a big barrier to developing effective analytical approaches that are generalizable. Fortunately, FL is able to use interconnected medical systems to transform medical data into diagnostic evidence to assist doctors in forward-looking diagnosis of patients \cite{Sheller}.

\subsubsection{Autonomous Driving Vehicles}
The autonomous driving system is a complicated system with a large amount of data. With multiple task layers from sensing, object detection, and tracking, to external object movement intention estimation, driving decision, and actuation, designing an autonomous management system requires real-time dynamics capturing and a huge amount of data from devices other than itself. Traditional centralized ML in autonomous driving aggregates large-scale data from all devices to a central server, which results in high computation and communication latency. Fortunately, FL sufficiently utilizes the computing capabilities of multiple learning agents to improve learning efficiency while providing better privacy for vehicles. Also, FL empowers adaption to environment dynamics, providing feature learning in different geographical locations, weather conditions, and pedestrians behavior dynamics \cite{Hongyi_Zhang}.

\subsection{Meta Learning}
Meta learning is used in highly automated AI,  natural language processing, and robotics.

\subsubsection{Automated AI}  In recent years, with the increasing amount of data, data scientists cannot address all challenging tasks in ML due to a lack of expertise and experience in the respective domain. Fortunately, automated AI is a data mining-based formalism that aims to reduce human effort and speed up the development cycle through automation. Based on the context of automated AI, one main advantage of meta learning techniques is that they allow hand-engineered algorithms to be replaced with novel automated methods which are designed in a data-driven way \cite{Dyrmishi}.

\subsubsection{Natural Language Processing}
Deep learning is one of the mainstream techniques in the NLP area and creates significant performance in many NLP problems. However, deep learning models are data-hungry, which limits learning models' application to different NLP tasks because collecting in-genre data for model training is costly. To address this issue, meta learning is deployed to the NLP area to learn more general NLP models, including better parameter initialization, optimization strategy, network architecture, distance metrics, and beyond \cite{Hung}.

\subsubsection{Robotics}
In the real world, a robot may encounter any situation from motor failures to finding itself in a rocky terrain where the dynamics of the robot are significantly different from one to another. As a result, the ability to adapt rapidly to unforeseen situations is one of the main open challenges for robotics. To solve the problem, meta learning is considered, where the robot enable to adapt
to the current situation with only a few gradient steps by using a single set of meta-trained parameters as initial parameters \cite{Rituraj}.

\subsection{FedMeta}
FedMeta integrates the advantages of FL and meta learning, and it can be used in autonomous driving vehicles, automated AI, NLP, and robotics, which have been introduced in previous sections.

\section{Open Problems and Future Directions}
Given the general research areas and challenges in integrating FL, meta learning, and FedMeta methodologies and their applications over wireless networks, we discuss the remaining research problems and future directions. The ultimate goal is to seamlessly integrate FL/meta-learning/FedMeta with wireless networks, resulting in a wide range of new designs ranging from techniques for computation offloading to network architectures.

\subsection{Federated Learning}
\subsubsection{FL Methodologies}
\begin{itemize}
    \item \textbf{Unreliable Model Upload:} In FL, devices may deliver low-quality local models trained by malicious data samples, which can adversely affect the learning accuracy. To address this issue, a reputation metric can be deployed to measure the reliability of each device, and select reliable devices for model aggregation.
    \item \textbf{Systematic and Model Heterogeneity:} A large number of devices and hardware specifications bring systematic heterogeneity to FL in practical systems. Also, FL has coupled with many different learning paradigms, which brings model heterogeneity. To deal with systematic heterogeneity, multi-center FL can be considered, where the devices with similar heterogeneity are clustered into a group for model aggregation. Meanwhile, to deal with model heterogeneity, one possible solution is to learn a personalized model for each device, so-called on-device personalization \cite{Mansour}. Its goal is to train a model for each device, based on the dataset of each device.
    \item \textbf{Imbalanced Data:} FL mainly focuses on IID and non-IID data. However, the data of real-world applications, such as computer vision and biomedicine, follows an imbalanced distribution \cite{Oksuz,Chongsheng_Zhang}. Imbalanced data typically refers to a problem with classification problems where the classes are not represented equally \cite{H_He}. To deal with this issue, a monitor scheme is integrated into FL that can infer the composition of training data in each FL round and detect the existence of possible global
    imbalance \cite{L_wang}.
    \item \textbf{Unsupervised FL:} Supervised FL enables multiple devices to share the trained model without sharing their labeled data. However, in real-world applications, the observed data may be unlabeled, which could limit the applicability of FL. To address this issue, a federation of unsupervised learning (FedUL) was proposed in \cite{FedUL}, where the unlabeled data were transformed into surrogate labeled data for each device, a modified model was trained by supervised FL, and the unsupervised FL learning model was recovered from the modified model. However, FedUL is not suitable for non-IID data. Thus, FedUL can be further improved by integrating with advanced FL aggregation or optimization schemes for non-IID settings \cite{jianyuwang}.
    \item \textbf{Federated Reinforcement Learning:} In FRL, rather than just uploading and downloading models, the agents need to exchange intermediate results and observations between themselves or with a central server. However, because of limited communication resources, the overhead is high, especially with an increasing number of agents. Meanwhile, some deep reinforcement learning (DRL) algorithms, such as deep Q network (DQN) \cite{Mnih} and Deep Deterministic Policy Gradient (DDPG) \cite{Lillicrap}, have multiple layers or networks, which contain millions of parameters, resulting in extremely high overhead. To solve these issues, several research directions should be considered. First, dynamic global model methods need to be designed to optimize the number of model exchanges. Second, devices or agents need to exchange the important parts of models or observations.
    \item \textbf{Robust to Poisoned Data:} FL models are usually trained by non-poisoned data. However, in practical scenarios, malicious servers or devices have negative effects on model training. Although FL by itself has a certain level of resilience against attacks, the frequent connections between servers and devices may spread the risk over networks and reduce the learning performance. Therefore, how to design an FL algorithm that is robust to poisoned data and design resilient networks for FL to avoid spreading attacks needs to be investigated.
\end{itemize}

\subsubsection{FL over Wireless Networks}
\begin{itemize}
    \item \textbf{Learning Convergence Analysis:} One of the most important considerations of FL is the convergence performance. Most existing research works in \cite{Zhaohui,NHTran,Shiqiangwang,Amiri,mingzhe1,Howard_yang} deployed traditional optimization methods to optimize wireless factors, such as transmission scheduling, transmission error, and energy to analyze the convergence of FL, and assumed that the optimization problem was convex. However, FL over practical wireless networks may not satisfy these conditions, especially when the optimization problems are non-convex. Also, the performance of convergence can be affected by dynamic wireless channels and device mobility. To address these issues, one possible solution is to deploy an FL algorithm that can handle heterogeneous device datasets, and capture the trade-off between convergence and energy consumption of devices with heterogeneous computing and power resources \cite{Dinh}.
    \item \textbf{Device Dropout:} For the device selection schemes proposed in \cite{Nishio,Yoshida,Anh}, the authors assumed that the wireless connection of each device was always available. Nevertheless, in practical wireless systems, some devices may become inactive due to poor connectivity and energy constraints, namely, device dropout. Thus, they may leave the FL process and cannot participate in model aggregation, which can severely degrade the performance of FL, such as low learning accuracy and low convergence speed \cite{ttAnh}. To deal with this issue, new FL algorithms need to be designed to be robust for the network, where only a small number of dynamic users exist for model aggregation \cite{Kamp}. Also, through designing the communication protocol, devices can actively deliver local models to the edge server when they are in good connectivity conditions. 
    \item \textbf{Hierarchical FL:} Accuracy and latency are two main factors for FL over wireless networks. The dynamic wireless environment can severely affect the transmission performance, which can lead to a low transmission rate and high transmission errors, and further lead to high transmission latency and low learning accuracy, especially for hierarchical FL (HFL). HFL is an architecture that deploys FL in heterogeneous wireless networks with three levels, including devices, SBS, and MBS \cite{MSHAbad}. In each time slot, a set of devices are selected to train the allocated global model using FedSGD algorithms \cite{Fed_SGD}, and then the local models are transmitted to their corresponding SBSs to be aggregated. The MBS and SBSs communicate with each other periodically to maintain a central model. HFL combines the advantages of edge FL and cloud FL. The cloud server can access more learning models, and the edge server enjoys more efficient communications with devices, leveraging edge servers as intermediaries to perform partial model aggregation in proximity, and relieve core network transmission overhead \cite{sLUO,Letaief}. However, the disadvantages of HFL are that the cloud may have excessive communication overhead and high latency, especially when a large number of devices and edge servers exist in wireless networks. One possible solution is to consider the joint design of device clustering, asynchronous FL, and communication efficiency, and use DRL algorithms to select optimal edge servers and devices for model aggregation in different time slots \cite{Fed_HFL_DRL}.
    \item \textbf{Cooperative Edge Computing for FL:} When performing FL in a network with multiple edge servers, if there is only a small number of mobile devices exist, only one edge server can be selected for model aggregation, which can save computation resources and energy for other edge servers. If the mobile device is far away from the active edge server, the learning model downloading/uploading can be done in Device-to-Device (D2D) connections. However, when a large number of mobile devices exist, selecting only one edge server for model aggregation can lead to high overhead and workload. Thus, multiple edge servers should be active for model aggregation. In this case, mobile devices need to select the optimal edge servers for model aggregation, and active edge servers should further select one edge server for global model aggregation.
    \item \textbf{Fully Decentralized FL:} Fully decentralized FL is usually used in a scenario with no servers or there is a failure or an attack on the server. Model weights are transmitted by D2D communication. There are no research works considering how wireless factors, such as transmission power, wireless channel, and spectrum resource allocation, affect the convergence rate and learning accuracy of fully decentralized FL. Meanwhile, because of limited communication resources in D2D transmission, how to select proper devices for model sharing still needs to be investigated.
\end{itemize}

\subsection{Meta Learning}
\subsubsection{Meta Learning Methodologies}
\begin{itemize}
    \item \textbf{Non-stationary Data Distribution:} When a new task does not exist in the experience buffer, meta learning cannot guarantee learning convergence and accuracy \cite{Zinkevich}. Meta learning also requires a large number of task datasets. Usually, authors assume that tasks are independently and identically distributed \cite{Shwartz}, and do not consider non-stationary distribution. If the task datasets change dynamically and do not have the same distribution, meta learning cannot adapt to the variation efficiently. Thus, it is difficult for meta learning to address complex datasets. One potential way to solve this issue is to propose a meta learning algorithm that is robust to tasks with a non-stationary distribution.
    \item \textbf{Robustness for Meta Learning:} Most meta learning algorithms are trained and tested using a small number of benchmark datasets, which means that the characteristics of datasets used for training are close to the datasets for testing. Thus, to the best of our knowledge, there are no meta-learning frameworks that not only can quickly adapt to new tasks with the help of prior experience, but also are robust to bad data samples, e.g. mislabeled data or outliers.
    \item \textbf{High Computation Costs:} Meta learning has the ability to wide the applicability of deep learning algorithms to more real-world domains. Consequently, increasing the generalization ability of meta learning algorithms is quite important. Although meta learning can learn new tasks quickly, meta training can be quite computationally expensive. Therefore, how to decrease the required computation latency and memory costs of meta learning remains an open challenge.  
    \item \textbf{Theoretical Convergence Analysis:} Except for the theoretical convergence analysis of MAML, the convergence theory of other metric-based, model-based, and gradient-based meta learning algorithms is not derived. Therefore, detailed convergence analysis of meta learning is need still to be investigated.  
\end{itemize}

\subsubsection{Meta Learning over Wireless Networks}
\begin{itemize}
    \item \textbf{Theoretical Analysis of Wireless Factors in Meta Learning:} It is possible that meta learning involves training a large number of meta learners, which requires much more communication demand than traditional learning approaches, especially when the number of tasks or meta learners increases exponentially. Thus, factors of the dynamic wireless environment, such as channel state information, transmission error, transmission energy, and computation capability of each device, can affect the performance of meta learning, and how these factors affect the performance of meta learning should be studied. 
    
    \item \textbf{Efficient Task Training:} Plenty of tasks may exist in wireless networks corresponding to a large number of devices. Thus, how to effectively sample tasks to train meta learning to satisfy the requirements of all devices and adapt to new tasks from the dynamic environment are of utmost importance.
\end{itemize}

\subsection{Federated Meta Learning}
\subsubsection{FedMeta Methodologies}
\begin{itemize}
    \item \textbf{Privacy for FedMeta:} In current FL, the shared global model still includes all devices' privacy implicitly, while in FedMeta, a meta-learner is shared. Thus, whether FedMeta has additional advantages in protecting device privacy from the model attack perspective \cite{Shokri,CSong,Szegedy1,Tramer} still needs to be explored.
    \item \textbf{Efficient FedMeta:} Incorporating the experience replay and parameter isolation approaches into the proposed ADMM-FedMeta in \cite{shengyue1} may further mitigate the catastrophic forgetting. In addition, although ADMM-FedMeta can be directly applied to reinforcement learning, it may result in low sample efficiency. Thus, it is essential to develop efficient collaborative reinforcement learning for FedMeta.
\end{itemize}

\subsubsection{FedMeta over Wireless Networks}
\begin{itemize}
    \item \textbf{Multi-Model FedMeta:} When a large number of tasks and devices exist in wireless networks, the overhead can be extremely large, which severely affects the transmission quality, increases the transmission latency, and may decrease the learning accuracy. To deal with these issues, one potential solution is to consider multi-model FedMeta, where devices are clustered into multiple groups, aggregate meta models in advance, and transmit them to the server for further aggregation. Thus, the possibility of multi-model FedMeta needs to be further investigated. 
    
    \item \textbf{Theoretical Analysis of Wireless Factors in FedMeta:} Also, with a large number of tasks and devices existing in wireless networks, how to characterize the convergence properties and communication complexity of FedMeta considering factors of wireless networks, such as channel state information and transmission error, require further study.
\end{itemize}

\section{Conclusions}
In this paper, we presented a comprehensive tutorial on the research evolution on FL, meta-learning, and FedMeta methodologies and their applications over wireless networks. We introduced the design, optimization, and evolution of these three learning approaches, providing a detailed literature review and identifying future research opportunities. By examining the advancements and challenges in FL, meta-learning, and FedMeta, we aimed to provide valuable insights and guidelines for optimizing, designing, and operating these learning algorithms in future methodologies and wireless networks, particularly in the context of emerging 6G networks.

\ifCLASSOPTIONcaptionsoff
  \newpage
\fi

\bibliographystyle{IEEEtran}
\bibliography{IEEEabrv,Ref,ReferencesMP}

\end{document}